\documentclass{style/naturep}

\usepackage{authblk}
\usepackage{amssymb, amsmath, graphicx, algorithmicx, algorithm, algpseudocode}
\usepackage{bbm, lineno, ragged2e, setspace, longtable, hyperref}
\usepackage{cleveref, anyfontsize, float, booktabs, multirow, xcolor, tabularx, caption, subcaption, geometry}
\usepackage{enumitem}
\usepackage{array}
\usepackage{xspace}
\usepackage{makecell}

\title{\justifying\bfseries A multimodal and temporal foundation model for virtual patient representations at healthcare system scale}

\author[1,2,3,4,$^{\ddagger}$]{Andrew Zhang}
\author[1,2,3,5,$^{\ddagger}$]{Tong Ding}
\author[1,2,3,$^{\ddagger}$]{Sophia J. Wagner}
\author[1,2,3,6]{Caiwei Tian}
\author[1,7]{Ming Y. Lu}
\author[1]{Rowland Pettit}
\author[1]{Joshua E. Lewis}
\author[1,8]{Alexandre Misrahi}
\author[1]{Dandan Mo}
\author[1,4,$\dag$]{Long Phi Le}
\author[1,2,3,$\dag$]{Faisal Mahmood}

\affil[1]{Department of Pathology, Mass General Brigham, Harvard Medical School, Boston, MA}
\affil[2]{Cancer Program, Broad Institute of Harvard and MIT, Cambridge, MA}
\affil[3]{Data Science Program, Dana-Farber Cancer Institute, Boston, MA}
\affil[4]{Health Sciences and Technology, Harvard-MIT, Cambridge, MA}
\affil[5]{Harvard John A. Paulson School of Engineering and Applied Sciences, Harvard University, Cambridge, MA}
\affil[6]{Department of Biomedical Informatics, Harvard Medical School, Boston, MA}
\affil[7]{Electrical Engineering and Computer Science, Massachusetts Institute of Technology (MIT), Cambridge, MA}
\affil[8]{School of Computer and Communication Sciences, EPFL, Lausanne, Switzerland}
\affil[$^{\ddagger}$]{Contributed equally (Co-first)}
\affil[$\dag$]{Co-senior authors}

\newcommand\Heading[1]{\noindent\textbf{\Large{#1}}}
\newcommand\heading[1]{\noindent\textbf{\large{#1}}}
\newcommand{\ours}{\textsc{Apollo}\xspace}
\newcommand{\dataset}{\textsc{MGB-7M}\xspace}

\makeatletter
\let\saved@includegraphics\includegraphics
\AtBeginDocument{\let\includegraphics\saved@includegraphics}
\makeatother

\begin{document}

%%%% Title and Authors %%%%
\begin{spacing}{1.2}
\maketitle
\newline
\noindent Corresponding author: Faisal Mahmood (faisalmahmood@bwh.harvard.edu)

%%%% Abstract %%%%
\Heading{Abstract}

\noindent Modern medicine generates vast multimodal data across siloed systems, yet no existing model integrates the full breadth and temporal depth of the clinical record into a unified patient representation. In the present study, we describe a comprehensive effort to model human clinical trajectories. We introduce \ours, a multimodal temporal foundation model trained and evaluated on over three decades of longitudinal hospital records from a major US hospital system, composed of 25 billion records from 7.2 million patients, representing 28 distinct medical modalities and 12 major medical specialties.  
\ours learns a unified representation space integrating over 100 thousand unique medical events in our clinical vocabulary as well as images and clinical text.
This ``atlas of medical concepts'' forms a computational substrate for modeling entire patient care journeys comprised of sequences of structured and unstructured events, which are compressed by \ours into virtual patient representations.
To assess the potential of these whole-patient representations, we created 322 retrieval and prognosis tasks from a held-out test set of 1.4 million patients. We demonstrate the generalized clinical forecasting potential of \ours embeddings, including predicting new disease onset risk up to five years in advance (95 tasks), disease progression (78 tasks), treatment response (59 tasks), risk of treatment-related adverse events (17 tasks), and hospital operations endpoints (12 tasks). Linear probing with \ours embeddings exhibits strong performance across all task categories. On a 30-task cancer progression benchmark, \ours surpasses even task-specific fully-supervised transformer models. Using feature attribution techniques, we show that model predictions align with clinically-interpretable multimodal biomarkers.  We evaluate semantic similarity search on 61 retrieval tasks, and demonstrate the potential of \ours as a multimodal medical search engine using text and image queries. Together, these modeling capabilities establish the foundation for computable medicine, where the full context of patient care becomes accessible to computational reasoning.

% \newpage

\clearpage
\Heading{Introduction} \label{sec:intro}

Electronic Health Records (EHRs) function as the system of record for modern medicine, underpinning the tightly coupled domains of care, operations, and research by systematically capturing complex, longitudinal health trajectories. These rich repositories hold immense promise for precision medicine and the creation of a unified, continuously learning healthcare system~\cite{moor2023foundation,jensen2012mining}. Yet, healthcare faces a paradox: while generating nearly 30\% of global data volume at 50 petabytes annually, less than 3\% is effectively utilized for clinical insight~\cite{rbccm2023healthcare,arcadia2023report}. The imperative to close this gap is underscored by the devastating impact of fragmented care, which contributes to an estimated 795,000 diagnostic error-related deaths or permanent disabilities annually in the United States alone~\cite{newmantoker2023diagnostic}. Unlocking the full potential of EHRs is currently hindered not only by systemic fragmentation across modality-specific silos, but also inherent data complexities -- extreme dimensionality, irregularity, and sparsity~\cite{cheng2016risk,xiao2018opportunities}.

In recent years, foundation models have emerged as a paradigm shift across domains such as natural language processing~\cite{brown2020language,devlin2019bert}, computer vision~\cite{oquab2023dinov2,chen2020simple} and computational biology~\cite{lin2023evolutionary,nguyen2024sequence}, learning context-rich representations via large-scale self-supervision that transfer effectively to diverse downstream tasks~\cite{bommasani2021opportunities}. While their potential in healthcare is recognized~\cite{rajpurkar2022ai,tu2024towards}, existing models remain limited in scope. Most approaches analyze individual data modalities in isolation -- such as pathology images or clinical text -- which inherently limits their utility for tasks requiring a holistic view of the patient~\cite{alsentzer2019publicly,yang2022gatortron,chen2024towards,vorontsov2024foundation,perez2025exploring,tiu2022expert,liu2023multimodal,khader2023medical}. Efforts to build longitudinal EHR foundation models~\cite{wornow2023ehrshot,steinberg2024motor,li2020behrt,rasmy2021med,miotto2016deep,redekop2025zero,rencZeroShotHealth2024,zhangChronoFormerTimeAwareTransformer2025,kraljevic2024foresight,shmatko2025learning,li2023hibehrt} have largely been restricted to structured data (e.g., diagnostic codes or laboratory results)~\cite{kauffman2025embedding} due to the ease of leveraging existing ontologies (e.g., ICD-10, SNOMED, LOINC), data warehousing infrastructure, and discrete-token sequence modeling techniques inspired by large language models (LLMs). However, structured data provides only a limited snapshot of a patient; subtle nuances such as medical reasoning traces, disease progression, and novel biomarkers are often only available in free-text notes and imaging. With recent evidence suggesting that multimodal AI systems can outperform single-modality approaches by 6--33\% across diagnostic tasks~\cite{soenksen2022integrated,huang2023multimodal}, there is a critical need for unified models capable of synthesizing the full spectrum of patient data. This challenge mirrors recent efforts in creating virtual cells, where foundation models that integrate multimodal data -- for example, synthesizing transcriptomic, proteomic, and morphological signals to predict cellular responses~\cite{adduriPredictingCellularResponses2025,bunne2024virtualcell} -- have demonstrated how unified representations can capture complex system dynamics that single-modality approaches miss. Just as these models revealed emergent properties in cellular state transitions, virtual patients could unlock similar insights into health trajectories by synthesizing the full spectrum of longitudinal EHR data.

To realize this vision, we introduce \ours, a multimodal temporal foundation model that transforms the entirety of the longitudinal medical record into a unified computational representation. \ours first tokenizes individual medical events using modality-specific encoders and then fuses the resulting embeddings across timepoints using a transformer architecture (see \textbf{Results} and \textbf{Methods} for full architectural details). We develop \ours\ with over 25 billion medical events from 7.2 million patients spanning 33~years across a multi-institutional healthcare system, encompassing text-based, structured, and imaging modalities. Unlike curated research datasets limited to single departments or critical care cohorts~\cite{johnsonMIMICIVFreelyAccessible2023}, our corpus captures the full continuum of inpatient and outpatient care across 17~institutions and affiliated clinics, reflecting real-world heterogeneity in both practice and population. We evaluate \ours across 322 downstream tasks designed to assess clinical breadth, including patient retrieval (61 cohorts), predicting new disease onset (95 tasks), disease progression (78 tasks), treatment response (59 tasks), drug adverse events (17 tasks), and hospital operations endpoints (12 tasks). The whole-patient embeddings achieve strong performance on prognostic tasks, enable multimodal guided retrieval from a large-scale database, yield interpretable latent structures that mirror phenotypic clusters, and reveal multimodal biomarkers.
Together, these results demonstrate that unified, temporally grounded patient representations can convert the EHR from a static archive into an active computational substrate for care, research, and operations -- the foundation for next-generation computable medicine.

\Heading{Results} \label{sec:results}
\vspace{1em} 

\begin{figure}[ht!]
    \centering
    \includegraphics[width=\linewidth]{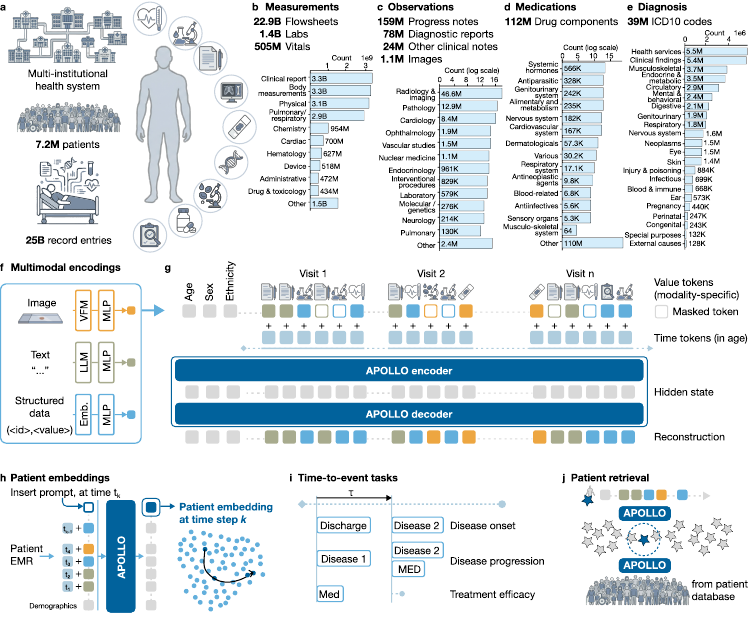}
    \caption{\footnotesize \textbf{Overview of \dataset and \ours.} (a) Overview of the pretraining dataset \dataset curated from 17 hospitals in one large-scale health care system consisting of 7.15 million patients. (b-e) Detailed distribution of \dataset including (b) LOINC code distribution of measurements, (c) distribution of diagnostic reports across medical domains, (d) medications grouped by ATC classification, and (e) ICD10 codes grouped by ICD10 chapters. (f) Modality-specific encodings of imaging data, text-based clinical notes and diagnostic reports, and structured measurements such as  laboratory tests, medications, and diagnosis. (g) Model architecture of \ours, an encoder-decoder architecture taking patient demographics (age, sex, ethnicity) and embeddings of all patient records in a temporal context as input. The pretraining objective is masked modeling, where modality-specific masked tokens are reconstructed. (h) At inference, all patient records are fed into the model together with one additional token that yields the patient embedding at the last time step. (i) Schematic of the curation of time-to-event tasks for risk prediction of disease onset, disease progression, and treatment outcome. (j) Schematic for patient retrieval for a query patient embedding (blue) from a large-scale database of candidate patients (gray).} 
    \label{fig:overview}
\end{figure}

\section*{Building a foundation model for clinical time series data}

\ours is a transformer-based model that integrates the entirety of a longitudinal, multimodal medical record, including clinical reports, structured data, and imaging data, in a temporal context. We curated \dataset, consisting of 7,155,044 patient electronic health records (EHR) with 25,296,943,893 distinct medical events. \dataset comes from a multi-institutional healthcare organization including specialty, ambulatory, and teaching hospitals (\textbf{\Cref{fig:overview}a}). The events span 33 years of which the majority of events are in the past 15 years. The dataset is roughly balanced between male (3,915,625) and female (3,236,747)  patients, with the majority of patients between 18 and 79 years old at the time of their last medical record (\textbf{Extended Data \Cref{tab:demo_basic}}).
\dataset covers the entire diagnostic spectrum including flowsheet measurements (22,936,351,260 events), laboratory tests (1,442,631,333 events), vital signs (505,193,723 events), and text-based observations including progress notes (158,683,290 notes) and diagnostic reports from various medical domains (77,978,881 reports), and imaging data (1,158,235 images) including hematology and histopathology imaging (\textbf{\Cref{fig:overview}b-e, Extended Data \Cref{tab:modality_distribution}}). 
\dataset covers a diverse spectrum of clinical entities with diagnostic reports from 12 major medical domains (\textbf{Extended Data \Cref{tab:modality_distribution}}), medications covering all 14 chapters of the Anatomical Therapeutic Chemical (ATC) Classification (\textbf{Extended Data \Cref{tab:meds-counts}}), and disease codes spanning all chapters of the International Classification of Diseases (ICD) ontology~\cite{icd10code}, with on average 1.8 million diagnoses per chapter (\textbf{Extended Data \Cref{tab:icd-counts}}). 

\ours integrates all medical events in the patient history into a unified, temporally aware mathematical representation. Inspired by large language models, we convert the medical record into a sequence of tokens: structured data are mapped to sequences of discrete event tokens, while unstructured data modalities are read as text tokens or image patch tokens. Each modality -- text, imaging, or structured -- is independently converted into continuous embeddings (value tokens) using modality-specific adapters. More specifically, we encode text-based modalities using a clinical large language model~\cite{yang2022gatortron}, we encode images using medical vision foundation models~\cite{ding2025multimodal,lu_visual-language_2024}, and we map structured tokens to learnable embeddings (\textbf{\Cref{fig:overview}f)}. Subsequently, all tokens are projected with modality-specific layers to a common representation space which reflects the semantic relationships between events. We prepend patient demographics (age at last event, sex, and ethnicity) to the input sequence and add the per-timestep patient age to each event as a positional (time) encoding (\textbf{\Cref{fig:overview}g}). By isolating the main computational unit (the temporal transformer) from the raw data, we reduce the risk of protected health information (PHI) leakage since no raw data enters the final model. \ours effectively scales training to 28 modalities from an active multi-institutional EHR system across all departments, satellite clinics, covering both inpatient and outpatient visits. This is in contrast to existing EHR modeling efforts~\cite{rencZeroShotHealth2024, pellegriniEHRsPatientPathways2025, zhangChronoFormerTimeAwareTransformer2025} utilizing curated, publicly accessible datasets which have undergone substantial curation and preprocessing, are restricted to specific clinical departments such as MIMIC-IV~\cite{johnsonMIMICIVFreelyAccessible2023}, or are limited to structured data modalities only~\cite{waxler2025comet}. 
We pretrain \ours employing masked token modeling~\cite{devlin2019bert} with modality-specific masks that prompt the output head for the correct modality during the forward pass (for more details see Section \textbf{Pretraining} in \textbf{Online Methods}, \textbf{Extended Data \Cref{tab:hparams_pretrain}}). To generate a patient embedding during inference, the last (most recent) time token is appended to the input sequence along with a modality-specific mask token as prompt, and the corresponding hidden state is used as the patient embedding (\textbf{\Cref{fig:overview}h}). In practice, any modality mask can be used as the prompt, and the resulting patient embedding is tuned towards that modality. For example, we later explore the applications of text- and image-prompted patient embeddings in multimodal semantic retrieval. For the majority of downstream tasks, however, we utilize the diagnosis mask as the prompt, due to the disease-oriented nature of most tasks.

We evaluate \ours on a wide set of 261 prognostic and 61 retrieval tasks designed to demonstrate the general-purpose utility of these patient-level embeddings.
To assess prognostic value, we train Cox proportional hazards models for time-to-event (TTE) prediction using the frozen \ours patient embedding as features (\textbf{\Cref{fig:overview}i}). Clinical endpoints include new disease onset, disease progression, treatment response endpoints, drug-related adverse events, and acute hospital operations endpoints. Beyond prognosis, \ours patient embeddings can be used as a search index for retrieving patients with similar disease history and medical phenotypes. Lastly, we extract insights into the clinical factors driving model risk predictions, marking a step towards biomedical discovery across the entirety of the medical record (\textbf{\Cref{fig:overview}j}).

\section*{\ours's embedding space as an atlas of medical concepts} 

\begin{figure}[ht!]
    \centering
    \includegraphics[width=\linewidth]{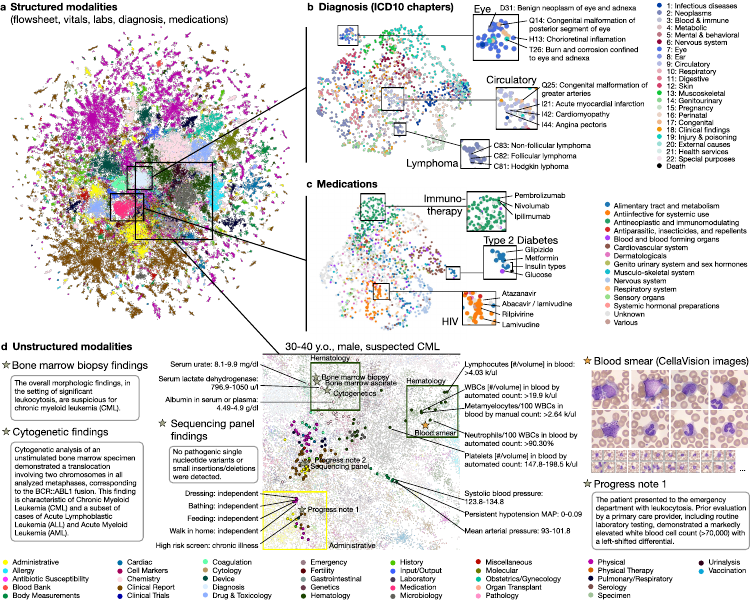}
    \caption{\footnotesize \textbf{\ours generates an atlas of medical concepts.} (a) Uniform manifold approximation and projection (UMAP) of the 103,940 discrete tokens that occur more than 100 times shows that \ours learns the underlying semantics of the discrete concepts. The emerging atlas of medical concepts exhibits meaningful spatial relationships both within modalities, as seen for (b) diagnosis codes and (c) for medications, as well as across modalities (d). (d) Structured and unstructured embeddings of the health record entries of a visit of one patient (30-40 y.o., male, suspected chronic myeloid leukemia), projected to the UMAP fit of (a).}
    \label{fig:concepts}
\end{figure}

For the first time, \ours brings siloed events across the entirety of the medical record into a unified embedding space, permitting the detailed study of intra- and cross-modal relationships. 
To understand how \ours unifies data representations across modalities, we examine the input  embedding space of \ours after pretraining on \dataset. The structured modalities, i.e., flowsheets, vitals, laboratory tests, medications, and diagnosis codes, generalize across medical events and, by analogy to language modeling, form \ours's vocabulary used to compose patient timelines. The low-dimensional representations of all 103,940 discrete tokens that occurred more than 100 times in the records show that \ours captures similar medical concepts by grouping flowsheets, vitals, and lab tests into distinct semantic clusters, such as allergy, coagulation, drug \& toxicology, or hematology, aligning with parent categories in the LOINC ontology (\textbf{\Cref{fig:concepts}a}).

Using modality-specific masks during pretraining, \ours learns a natural separation between modalities, such as diagnostic codes and medications. Upon closer examination of the diagnostic tokens, we observe that the disease tokens of related diseases are closely embedded (\textbf{\Cref{fig:concepts}b}). Across the ICD10 chapters, diseases of the eye cluster together with eye-related diseases from other chapters, such as benign neoplasms of eye and adnexa. Within diseases of the circulatory system, both acquired and congenital heart-related diseases are neighboring acute myocardial infarction. Similarly, different lymphomas cluster closely together. These results show that \ours learns multiscale semantic relationships between underlying conditions and associated anatomical organs that align with but are not limited by clinician-defined ontologies (e.g., ICD-10).

The embeddings of prescribed medications and drug components labeled by RxNorm ingredient code show similar semantic clustering (\textbf{\Cref{fig:concepts}c}). In particular, \ours closely embeds drugs such as pembrolizumab and nivolumab, both used in immunotherapies targeting the PD-1 protein in cancer cells, and antiretroviral drugs, such as abacavir, lamivudine, or rilpivirine, which are used in combination to treat HIV infections. Drugs related to type 2 diabetes, such as glipizide and metformin, are grouped together in close proximity to blood sugar indicators insulin and glucose. 

Beyond discrete tokens encoding structured information, \ours additionally encodes event-specific diagnostic reports, progress notes, or clinical images. Remarkably, these unstructured event embeddings map to locations in the atlas which correspond to semantically related structured tokens. To understand how text and images integrate with the discrete input embedding space, we analyze all embeddings of one patient (30-40 y.o., male) in one day of their medical record, consisting of 184 lab tests, 58 flow sheet entries, one medication, six clinical reports, three progress notes, and images from one blood smear (\textbf{\Cref{fig:concepts}d}). We observe that the clinical reports from the bone marrow aspirate and bone marrow biopsy are embedded close to the cytogenetic report, all three situated within the hematology cluster in green. All blood count-related lab tests are located in the same hematology cluster as are the blood smear images. The closest lab tests to the blood smear show elevated levels of neutrophils and their precursor form metamyelocytes, in line with the observation of leukocytosis mentioned in the biopsy note and progress note. The progress note describing the patient visit is located in the administrative events cluster together with routine flowsheet entries. In a completely self-supervised manner, \ours learns clinician-aligned representations of diverse medical events, both structured and unstructured, forming a multimodal atlas of medicine. These embeddings are the building blocks for modeling representations of the patient state.

\section*{\ours patient embeddings encode medical phenotypes}

\begin{figure}[ht!]
    \centering
    \includegraphics[width=0.93\linewidth]{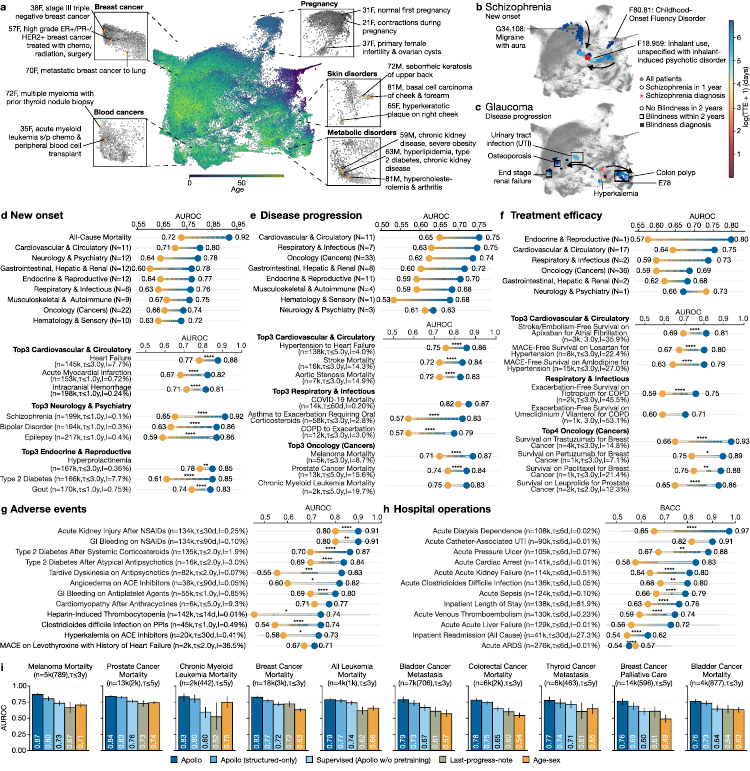}
    \caption{\footnotesize \textbf{Evaluation of \ours's patient embeddings.} 
    (a) Uniform Manifold Approximation and Projection (UMAP) visualization of 100 thousand randomly sampled patient embeddings from the data partition for downstream evaluation, labeled by age. Local neighborhoods reveal clustering of patients with similar clinical phenotypes.
    (b) Patient trajectories of 10 random patients before their diagnosis of Schizophrenia shows convergence to the same cluster in the patient embedding space.
    (c) Patient trajectories of 10 random patients with Glaucoma, five of which develop blindness within two years. The patient trajectories diverge from the disease cluster, where the patients with later diagnosed blindness cluster together. 
    (d-h) Quantitative evaluation on 261 curated downstream tasks. Performance of \ours (blue) and the age--sex reference (orange) is given in AUROC scores across all tasks.
    (d) New disease onset prediction (95 tasks),  
    (e) disease progression prediction (78 tasks), 
    (f) treatment response prediction (59 tasks),
    (g) adverse effect (17 tasks, where top 12 tasks are shown in the figure), and 
    (h) hospital operation tasks (12 tasks). 
    (i) Ablation study on architectural design choices of \ours, evaluated on disease progression of neoplastic diseases. Top 10 tasks are shown, more tasks can be found in \textbf{Extended Data Figure \ref{fig:edf_ablation}}. 
    Tasks in each plot are sorted by \ours's performance.
    Statistical significance was assessed by an unpaired, nonparametric bootstrap test of the baseline with respect to \ours. $^{*}p\leq0.05$, $^{**}p\leq0.01$, $^{***}p\leq0.001$, $^{****}p\leq0.0001$. N, number of tasks; n, number of patients; $\tau$, time-to-event; I, class imbalance in test set.}
    \label{fig:results}
\end{figure}

Through the time-aware transformer architecture, the individual medical events encoded by \ours are composed into patient-level representations which encode the entire clinical history of a patient. In a UMAP of 100 thousand randomly sampled patients, each at a random timepoint, we observe that their embeddings reflect underlying disease phenotypes, age gradients, and shared medical histories (\textbf{\Cref{fig:results}a}). We highlight clusters related to breast cancer, pregnancy, blood cancers, skin disorders, and metabolic disorders. Just like the individual event embeddings, these clusters are learned from scratch in a self-supervised manner. Because patient embeddings are dynamic and change over time, we can trace their trajectories as patients approach the onset of a disease or experience disease progression. For a cohort of patients who are eventually diagnosed with schizophrenia, their embeddings converge to the same region of the embedding space as they approach the diagnosis date (red stars; \textbf{\Cref{fig:results}b}). Similarly, patient trajectories of a cohort in the months following a diagnosis of glaucoma diverge into a group eventually developing blindness and a group showing different mostly unrelated conditions (\textbf{\Cref{fig:results}c}).

\section*{\ours predicts risk of new disease onset}
We evaluate the predictive capabilities of \ours on 95 downstream tasks for new disease onset. More specifically, we curate endpoints including 94 distinct disease codes spanning eight categories ranging from cardiovascular and circulatory diseases to cancers, as well as all-cause mortality (\textbf{\Cref{fig:results}d, Extended Data \Cref{tab:task_meta-new_onset}}). Each disease onset task includes at least 60 thousand patients in total, where the number of uncensored patients (those who experience the endpoint) ranges from 157 to 27,503 per task reflecting highly variable real-world disease incidences. All tasks are curated from the test set of \dataset spanning 1,431,009 patients in total with a varying number $n$ of patients considered per task depending on the inclusion criteria. For each disease, we predict time to diagnosis starting from a random hospital discharge (hereafter referred to as the \textit{snapshot time}; see \textbf{Evaluation Framework} in \textbf{Online Methods}) using a Cox-regression model on PCA-transformed \ours patient embeddings. As the primary performance metric, we report time-dependent cumulative/dynamic AUC evaluated at a fixed threshold duration between one and six years in the future depending on the physiological time course of the disease. As a statistical reference, we train separate Cox models using patient age and sex as features. Overall, \ours significantly outperforms the baseline on 74 out of 95 diseases ($p<0.05$).

Per disease category, \ours performs best on risk prediction of all-cause mortality within the next year (0.92 AUROC). Investigating the specific diseases in the best-performing categories, we find that \ours significantly improves risk prediction over age-sex features for cardiovascular and circulatory diseases: 3-year heart failure prediction improves from 0.77 to 0.88 AUROC, 1-year acute myocardial infarction improves from 0.68 to 0.82 AUROC, and 1-year intracranial hemorrhage improves from 0.71 to 0.81 (all $p<0.0001$). 
For conditions related to neurology and psychiatry, \ours predicts 1-year risks for schizophrenia with 0.92 AUROC compared to 0.65, bipolar disorder with 0.86 AUROC compared to 0.63, and epilepsy with 0.86 AUROC compared to 0.59 (all $p<0.0001$). 
For diseases of the endocrine and reproductive system, \ours significantly improves the 3-year risk prediction of hyperprolactinemia from 0.78 to 0.85 ($p<0.01$), for 3-year risk of type 2 diabetes from 0.61 to 0.85 ($p<0.0001$), and for 1-year risk of gout from 0.74 to 0.83 ($p<0.01$) AUROC, respectively (\textbf{\Cref{fig:results}d}). Results for all experiments can be found in \textbf{Extended Data \Cref{tab:perf-new_onset}}.

\section*{\ours patient representations predict disease progression}

To assess the performance of \ours in predicting disease progression in more targeted cohorts, we curate 78 tasks for predicting the risk of a patient to progress from a disease to a more severe form. Our tasks span eight disease categories, and we predict time between diagnosis of the original disease and the progressed form (\textbf{\Cref{fig:results}e, Extended Data \Cref{tab:task_meta-dp}}). The incidence for these tasks is higher than for the new onset tasks due to more restrictive inclusion criteria for the candidate set of censored patients (for more details see \textbf{Downstream tasks} in \textbf{Online Methods}). 
Similar to the new onset task formulation, we model the risk using a Cox-regression model with task performances computed over varying threshold durations, from 60 days (for acute diseases) up to five years (for more chronic diseases). Overall, \ours significantly outperforms the baseline on the large majority of tasks with 53 out of 78 diseases ($p<0.05$).

\ours shows the highest gains in risk prediction for disease progression for cardiovascular and circulatory diseases (on average from 0.65 to 0.75 AUROC), followed by the progression of diseases of the respiratory system and infectious diseases (on average from 0.63 to 0.75 AUROC), and the progression of cancers (on average from 0.62 to 0.74 AUROC). More specifically, \ours can predict the 5-year risk of hypertension to heart failure significantly better than the baseline (0.75 to 0.86), 3-year survival after stroke with 0.84 AUROC compared to 0.72 from the baseline, and 3-year survival after aortic stenosis with 0.83 compared to 0.72 for the baseline (all $p<0.0001$). 
Moreover, \ours significantly improves the risk prediction for asthma exacerbation within 3 years (from 0.57 to 0.83 AUROC, $p<0.0001$) and chronic obstructive pulmonary disease (COPD) exacerbation within 3 years (from 0.57 to 0.79, $p<0.001$).
Finally, survival prediction significantly improves for melanoma from 0.71 to 0.87 (3-year survival) and for prostate cancer from 0.74 to 0.84 (5-year survival) (both $p<0.0001$). It also improves for chronic myeloid leukemia from 0.75 to 0.83 (5-year survival), although this difference is not statistically significant due to relatively high variance across bootstraps. Results for all experiments can be found in \textbf{Extended Data \Cref{tab:perf-dp}}.

\section*{\ours predicts treatment response and risk of adverse events}

To evaluate how well \ours can model response to medications, we curate a set of 59 tasks for treatment response prediction spanning six disease categories, where the majority of treatments are neoplastic or immunomodulating agents (\textbf{\Cref{fig:results}f, Extended Data \Cref{tab:task_meta-te}}). Similar to the previous tasks, we use Cox regression to predict the risk for each endpoint starting from time of first treatment administration. Overall, \ours significantly outperforms the baseline on 30 out of 59 tasks ($p<0.05$).

Among the disease categories, the best performing groups are medications acting on the endocrine system, where \ours improves the risk prediction from 0.57 to 0.8 AUROC, and on the cardiovascular and circulatory system tract, where the performance increases from 0.64 to 0.75 AUROC using \ours embeddings (\textbf{\Cref{fig:results}f, Extended Data \Cref{tab:task_meta-te}}). In particular, \ours predicts stroke-free survival on apixaban therapy for atrial fibrillation with 0.81 AUROC significantly better than the baseline with 0.69 ($p<0.0001$) and major adverse cardiac events (MACE)-free survival on losartan for hypertension with 0.8 AUROC ($p<0.0001$) and on amlodipine for hypertension with 0.79 AUROC ($p<0.0001$). 
For treatments of the respiratory system, \ours predicts the survival on tiotropium for COPD with 0.75 AUROC compared to 0.59 ($p<0.001$) and exacerbation-free survival on umeclidinium/vilanterol for COPD with 0.71 AUROC compared to 0.6. 
Finally, for cancer treatments, \ours significantly improves survival prediction on trastuzumab and pertuzumab, both targeted therapies for HER2-positive breast cancers, from 0.66 to 0.93 ($p<0.0001$) and from 0.75 to 0.89 ($p<0.05$). Similarly, \ours significantly improves survival prediction for  chemotherapy with paclitaxel for breast cancer (from 0.75 to 0.88, $p<0.01$) and for hormone therapy with leuprolide for prostate cancer (from 0.65 to 0.86, $p<0.0001$). Results for all experiments can be found in \textbf{Extended Data \Cref{tab:perf-te}}.

To assess the ability of \ours to anticipate clinically relevant adverse events, we evaluate a curated set of 17 tasks covering cardiovascular, gastrointestinal, metabolic, and drug-induced complications (\textbf{\Cref{fig:results}g, Extended Data \Cref{tab:task_meta-ae}}). Across all adverse-event tasks, \ours significantly improves risk stratification on 12 out of 17 tasks over the baseline. For adverse events after nonsteroidal anti-inflammatory drug (NSAID) exposure, performance increases significantly for acute kidney injury (AUROC 0.80 to 0.91, $p<0.0001$) and GI bleeding (0.80 to 0.91, $p<0.01$). Similarly, the model captures metabolic risk with significant gains in predicting treatment-emergent type~2 diabetes following systemic corticosteroid use (0.70 to 0.87, $p<0.0001$) and atypical antipsychotic therapy (0.69 to 0.84, $p<0.0001$). Results for all experiments can be found in \textbf{Extended Data \Cref{tab:perf-ae}}.

It is important to distinguish between the task of comparing across treatments versus making predictions across patients in our scenario. Because each task conditions on a specific treatment, these predictions stratify within a treated cohort rather than across drugs. Therefore, the risk scores likely reflect a combination of baseline patient health and treatment-specific factors: patients at higher risk of adverse events from NSAIDs, for example, may have underlying conditions that predispose them to adverse events in general, rather than NSAID-specific risk. 

\section*{\ours for clinical operations management}  

We evaluate 12 operational tasks that reflect short-term in-hospital outcomes and resource demands (\textbf{\Cref{fig:results}h, Extended Data \Cref{tab:task_meta-ops}}). These tasks include predicting all-cause inpatient readmission within 30 days, inpatient length of stay over 7 days, and 10 acute care endpoints where we predict whether the endpoint will occur within 7 days of emergency department (ED) admission. For the length of stay and acute care endpoints, we provide the patient history up to 24 hours after ED admission as input into the model, thus the effective threshold duration is 6 days. Since the censorship is substantial for these tasks, with incidences as low as 0.01\%, we report performance using balanced accuracy (for more details see Section \textbf{Metrics} in\textbf{Online Methods}). Overall, \ours yields significant improvements in 9 out of 12 settings. In particular, balanced accuracy increases from 0.65 to 0.97 for in-hospital dialysis dependence, 0.64 to 0.80 for acute kidney failure, and 0.66 to 0.79 for sepsis (all $p<0.0001$). For broader operational metrics, the model improves prediction of prolonged hospital stay (0.63 to 0.76) and all-cause readmission within 30 days (0.54 to 0.62; both $p<0.0001$). Results for all experiments can be found in \textbf{Extended Data \Cref{tab:perf-ops}}. Taken together, these findings indicate that the learned temporal patient representations generalize across pharmacologic settings and time scales, potentially providing a basis for the optimization of hospital operations in the future.

\section*{\ours's risk scores stratify patient groups and are well-calibrated}

To assess whether the risk prediction with \ours can be used for patient stratification, we divide the patient cohort into low and high-risk groups based on the predicted risk scores. We select the top and bottom 25\% of risk scores for the high- and low-risk groups, respectively (\textbf{Extended Data \Cref{fig:edf_km_new_0,fig:edf_km_new_1,fig:edf_km_dp_0,fig:edf_km_dp_1,fig:edf_km_te_0,fig:edf_km_te_1,fig:edf_km_ae_0,fig:edf_km_op_0}}). Since the prevalence of many endpoints is very low in real-world scenarios, the stratification could be improved by using a fraction smaller than 25\%. Across all sets of tasks, we observe that the risk scores of \ours stratify the patient cohorts well, particularly for tasks with lower censorship, e.g., cancer mortality, heart failure mortality (\textbf{Extended Data \Cref{fig:edf_km_dp_0}}).
Calibration metrics further confirm that the risk scores of the model are well-calibrated across all groups of tasks, where we observe that tasks with larger sample sizes are better calibrated than tasks with fewer patients (\textbf{Extended Data  \Cref{fig:edf_calib_new_0,fig:edf_calib_new_1,fig:edf_calib_dp_0,fig:edf_calib_dp_1,fig:edf_calib_te_0,fig:edf_calib_te_1,fig:edf_calib_ae_0,fig:edf_calib_op_0}}). 

\section*{Ablation studies}

Semantic alignment across modalities is crucial for modeling diseases with inherently multimodal diagnostics such as neoplasms. When assessing long term risk predictions on durations up to five years, understanding the patient health state across modalities and modeling the temporal context of the patient history is important. To quantify the contribution of each component of our model, we perform a series of ablations on the neoplasm disease progression task set comprising 28 downstream tasks. We compare the full version of \ours to (i) a structured-data-only model trained with the same pretraining objectives, (ii) a supervised model with the same architecture trained end-to-end on each individual task, (iii) a progress-note baseline using the embedding of the most recent progress note, and (iv) the age-sex baseline (\textbf{\Cref{fig:results}i, Extended Data \Cref{fig:edf_ablation}}).

The strongest gains appear in hematologic malignancies: for chronic myeloid leukemia mortality, \ours improves over the progress note model by +0.31 AUROC and over the supervised model by +0.23. For solid tumors, \ours outperforms the supervised baseline by +0.16 on breast cancer palliative care, +0.13 on colorectal cancer mortality, and +0.12 on bladder cancer metastasis. Survival prediction for prostate cancer is already well captured by structured data alone (AUROC 0.83 vs.\ 0.84 with all modalities), whereas melanoma and breast cancer survival benefit more substantially from multimodal integration (+0.07 and +0.06 AUROC, respectively). These results show that the benefits of multimodal, temporally contextualized pretraining extend consistently across cancer types and clinical endpoints.

Across all tasks, \ours achieves a mean AUROC of 0.735 improving over the structured-only variant by +0.025 (0.71). This highlights the importance of our multimodal integration incorporating clinical notes and imaging for modeling long-term cancer progression. The supervised baseline reaches a mean AUROC of 0.626 (-0.109), indicating that task-specific fine-tuning alone is insufficient compared to our pretraining strategy. Similarly, the progress note baseline obtains 0.615 on average (-0.12) and the age-sex baseline achieves 0.619 (-0.116), demonstrating that modeling temporal context beyond the most recent encounter is essential. Overall, the ablations confirm that multimodal integration, temporal modeling, and large-scale pretraining are all necessary components of \ours, each contributing substantially to accurate long-term cancer progression risk assessment.

\begin{figure}[h!]
    \centering
    \includegraphics[width=\linewidth]{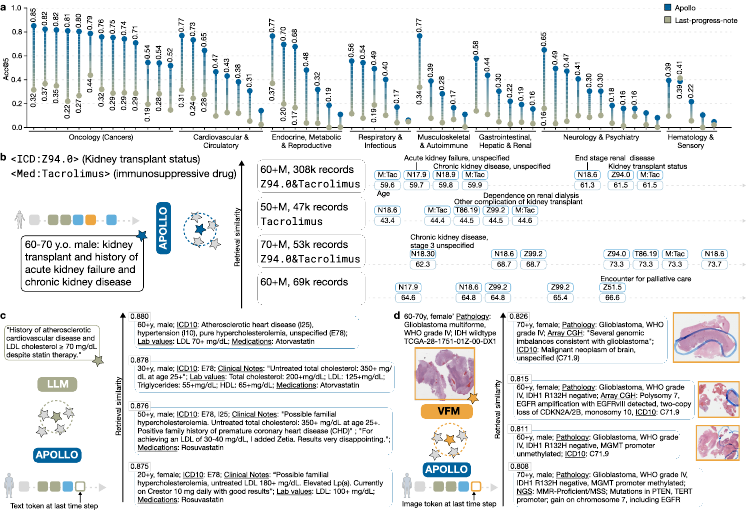}
    \caption{\footnotesize \textbf{\ours enhances patient retrieval.} (a) 61 patient retrieval tasks curated from combinations of ICD10 diagnosis codes and medications, assessed with accuracy among the five closest (Acc$@5$) embedded patients compared to retrieval of the latest progress note embedding, (b) qualitative evaluation of the closest embedded patient to a patient for kidney transplant maintenance. (c) Text-based retrieval on the example of a clinical trial inclusion criteria using \ours treatment assessment plan token as the last prompt. (d) Image-based retrieval using an external whole-slide image from TCGA using the corresponding TITAN embedding~\cite{ding2025multimodal} as last token. To preserve patient privacy, ages shown in the figure are randomly shifted by -2 to 2 years from their actual age, and lab test and vitals values are presented as ranges instead of exact numbers.}
    \label{fig:retrieval}
\end{figure}

\section*{Patient-level representations enable semantic search}

The prognostic evaluations above test whether \ours embeddings encode information predictive of future clinical events. A complementary question is whether the learned embedding space encodes similarities between patients based on past clinical events. Retrieval tasks probe this by querying the embedding space directly using nearest-neighbor search. We measure whether the returned patients have similar clinical profiles as the query. Strong retrieval performance would support the utility of \ours not only for prediction but also for cohort discovery, clinical trial matching, and medical reasoning.

By compressing entire medical histories into a single vector, we can quantify the similarity of any two patients via cosine distance and thus perform semantic retrieval. To mimic a clinical deployment scenario, we construct a search index composed of the entire evaluation set of 1.4 million patients, each embedded at the calendar time of January 1st, 2025.  We curated 61 patient cohorts defined by a particular diagnosis followed by administration of a first-line therapy. These inclusion criteria are extracted from the structured medical events via SQL. We divide each cohort into five equal folds, and set each member of one fold as the queries and retrieve the most similar patients from the remainder of the search index (for more details see Section \textbf{Retrieval} in \textbf{Online Methods}). We compute the agreement between the top-5 semantically retrieved patients and the cohort defined by the structured inclusion criteria. As a baseline, we construct a search index using the unimodal embedding of the most recent progress note prior to 1/1/2025 (\textbf{\Cref{fig:retrieval}a, Extended Data~\Cref{fig:edf_retrieval}}).

Compared to the baseline, embedding the entire patient history using \ours yields higher agreement with the SQL ground-truth cohorts across all tasks and achieves particularly high retrieval performance for diseases which more severely affect patient health. The top-performing cohorts include neoplasms (ovarian cancer on carboplatin, multiple myeloma on bortezomib, acute myeloid leukemia on cytarabine), circulatory diseases (chronic ischemic heart disease on aspirin, atrial fibrillation on apixaban, and hypertension on lisinopril), and metabolic and endocrine conditions (lipidemia on atorvastatin, hypothyroidism on levothyroxine, and T2DM on metformin). In contrast, lower retrieval performances are observed for less-serious conditions such as iron deficiency anemia or vitamin B12 deficiency anemia. These trends suggest that the patient state is primarily impacted by conditions or interventions which have an impact on future diagnoses or mortality. Indeed, this is expected as the patient state is obtained from the masked-diagnosis prompt (\textbf{\Cref{fig:overview}h}). A full list of cohorts and their retrieval performance can be found in \textbf{Extended Data \Cref{tab:task_meta-ret,tab:perf-ret}}.

We next take a deeper look at some qualitative retrieval results (\textbf{\Cref{fig:retrieval}b}). Our query patient is a 60-70 y.o. male with a history of chronic kidney disease and acute kidney failure, and was prescribed tacrolimus to prevent organ rejection following a kidney transplant (ICD10 code Z94.0). The top-4 closest patients (retrieved using \ours embeddings) have medical records ranging from 47 thousand to 308 thousand events. All four patients show histories of chronic kidney disease, acute kidney failure, and dependence on renal dialysis, which are very similar to the query patient. These results also show that semantic similarity in some cases can retrieve relevant patients which are missed by traditional SQL methods based on structured data. For example, the second patient is missing the ICD10 code for kidney transplant (Z94.0), despite having a history of kidney transplant complications (T86.19) and renal dialysis (Z99.2), followed by a prescription for tacrolimus. If we were to retrieve similar patients based only on the ICD10 code Z94.0, this patient would be missed.

Besides retrieval using a query patient, \ours also enables retrieval using arbitrary modalities, in particular text (\textbf{\Cref{fig:retrieval}c}) and images (\textbf{\Cref{fig:retrieval}d}). For example, we can write a set of inclusion criteria in natural language and embed it as the query (\textbf{\Cref{fig:retrieval}c}). To generate the search index, we use the corresponding modality as the prompt token (i.e., masked text prompt instead of masked diagnosis prompt) to generate embeddings for every patient in the evaluation set. The top four retrieved patients all exhibit medical histories which align with the desired inclusion criteria, including high LDL cholesterol despite statin therapy, and likely or confirmed atherosclerotic heart disease. Additionally, we can use a single pathology slide as the query (\textbf{\Cref{fig:retrieval}d}). In this case, we use a slide from TCGA rather than from our evaluation set, to test the external generalizability of \ours embeddings. Similarly to the text-retrieval experiment, we utilize the masked image prompt token to generate the patient embeddings for the search index. The top retrieved patients all exhibit the pathology indicated in the TCGA slide: Grade IV Glioblastoma. Furthermore, at least three of the top 4 retrieved patients specifically align with the IDH wildtype status of the TCGA slide. The second patient is confirmed IDH wildtype due to absence of the R132H mutation and the presence of the +7/-10 chromosomal copy number variant signature. The third and fourth patients are IDH1 R132H negative which implies likely wildtype status. In summary, \ours patient embeddings unlock scalable semantic similarity search using diverse query modalities, ranging from a single textual description or image to an entire patient history.

\section*{\ours is interpretable at the bedside and at the population level}

\begin{figure}[ht!]
    \centering
    \includegraphics[width=\linewidth]{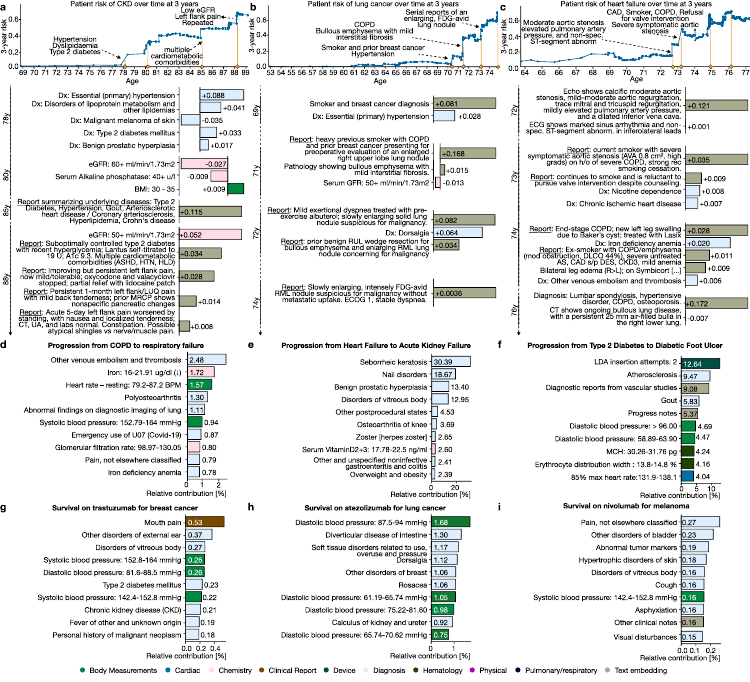}
    \caption{\footnotesize \textbf{\ours yields interpretable biomarkers at both the local and global level.}
(a–c) Local analysis. We plotted the model’s predicted 3-year risk for three example patients: (a) chronic kidney disease, (b) lung cancer, and (c) heart failure, as a function of age; markers indicate encounter times. At each prominent increase in risk, we performed a leave-one-token-out (LOTO) sensitivity analysis over events between the preceding visit $t_0$ and the current visit $t_1$: tokens are removed one at a time and the change in predicted risk at $t_1$ is recorded. Horizontal bars list the largest contributors in that interval, where positive values indicate increased risk and negative values decreased risk. 
(d–i) Global analysis. Population-level feature importance for six downstream tasks: (d) progression from COPD to respiratory failure, (e) heart failure to acute kidney failure, (f) type 2 diabetes to diabetic foot ulcer, (g) survival on trastuzumab for breast cancer, (h) atezolizumab for lung cancer, and (i) nivolumab for melanoma, obtained via Integrated Gradients (IG) as described in \textbf{Online Methods}. Bars show the top-10 tokens per task ranked by mean normalized IG among high-risk patients, and values are expressed as relative contribution (\%). To preserve patient privacy, ages shown in the figure are randomly shifted by -2 to 2 years from their actual age, and lab test and vitals values are presented as ranges instead of exact numbers.}
    \label{fig:interpretability}
\end{figure}

\ours produces dynamic risk trajectories that track an individual’s probability of an outcome as new evidence accrues (\textbf{\Cref{fig:interpretability}a–c}). For three index patients, we plot 3-year risk of chronic kidney disease (CKD), lung cancer, and heart failure, against age. The curves show long periods of stability punctuated by stepwise increases that coincide with clinically meaningful events. To explain the increases, we perform a leave-one-token-out (LOTO) sensitivity analysis over the interval that each jump happens. Specifically, we remove one event at a time from the interval and recompute the patient’s risk at the end of the interval. The largest positive changes identify tokens that most contributed to the increase, and negative values identify tokens whose removal would have increased risk, indicating protective or competing signals. For onset of CKD (\textbf{\Cref{fig:interpretability}a}), risk remained near zero through late 70s, then rose after documentation of essential hypertension, dyslipidaemia, and type 2 diabetes, and jumped again on a decline in eGFR to the low 50s and the findings of multiple cardiometabolic comorbidities, followed by repeated flank-pain evaluations. These tokens together reflect the emergence of chronic kidney damage on a background of long-standing vascular risk factors. For onset of lung cancer (\textbf{\Cref{fig:interpretability}b}), risk stayed low for decades, then increased with findings of smoking and prior breast cancer, and essential hypertension. Subsequent sharp rises came from COPD, emphysema and serial reports of an enlarging, FDG-avid lung nodule, mirroring the clinical work-up of incident lung cancer. For heart failure (\textbf{\Cref{fig:interpretability}c}), risk escalated following echocardiography evidence of severe aortic stenosis, ongoing smoking and COPD, and subsequent persistent refusal of valve intervention, progressive congestion and thrombosis. Additionally, the LOTO analysis also demonstrated protective events, such as borderline eGFR and normal serum alkaline phosphatase levels for CKD. It is worth noting that, besides canonical drivers, the top contributors also include medical events with clear associations with the diseases. While some of them are likely proxies such as markers of intensified surveillance or treatment, some reflect comorbidity clusters or shared risk environments, and a minority may represent hypothesis-generating signals for future study.

We next quantify feature importance at the population level for high risk patients that fall in the top hazard quartile using Integrated Gradients (IG) computed on the pre-Transformer inputs and aggregated by net sum across embedding dimensions, normalization and max-pooled per patient, and averaged across the population (for more details see Section \textbf{Interpretability} in \textbf{Online Methods}). Across six representative tasks (\textbf{\Cref{fig:interpretability}d-i}), \ours recovers clinically coherent risk structure. 
For progression from COPD to respiratory failure (\textbf{Figure~\ref{fig:interpretability}d}), high relative contributions are assigned to venous thromboembolism, tachycardia, abnormal lung imaging, severe hypertension and COVID-19, consistent with cardiopulmonary strain and intercurrent infection precipitating decompensation. Biomarkers including low serum iron and iron-deficiency anaemia are in line with recent works showing that iron deficiency in COPD is common and associated with worse exercise capacity, exacerbations, and prognosis~~\cite{amado2025non, dong2025association}.
For progression from heart failure to acute kidney failure (\textbf{Figure~\ref{fig:interpretability}e}), IG highlights obesity, post-procedural states and age proxies including seborrhoeic keratoses, nail disorders, osteoarthritis, vitreous body disease and zoster. While these likely partly reflect frailty and healthcare contact, several dermatologic and nail findings have been proposed as cutaneous markers of atherosclerotic and degenerative cardiovascular disease, including heart failure~~\cite{ghonemy2021clinical, katira2022dermatological}.
For progression from type 2 diabetes to diabetic foot ulcer (\textbf{Figure~\ref{fig:interpretability}f}), the dominant risk factors are atherosclerosis, vascular imaging, blood-pressure extremes and red-cell indices, reflecting macrovascular disease, haemodynamic load and systemic inflammation, all known to underlying diabetic foot pathology~~\cite{soyoye2021diabetes}. Gout also appears among the top contributors, which aligns with recent studies showing that gout in diabetes are linked to vascular complications, neuropathy and a high-risk diabetic foot, and that tophaceous gout can mimic or worsen foot ulcers~~\cite{rasmussen2025gout, rasmussen2025identifying}.
For survival on trastuzumab in breast cancer (\textbf{Figure~\ref{fig:interpretability}g}), IG primarily surfaces host factors including high blood pressures, type 2 diabetes, chronic kidney disease and prior malignancy, consistent with data that cardiometabolic comorbidities and comorbidity burden influence both treatment tolerance and overall survival in HER2-positive breast cancer~\cite{valiyaveettil2023cardiotoxicity, yalcciner2025impact, carreira2025use}. Symptoms such as mouth pain plausibly capture mucosal toxicity and treatment intensity over the course of therapy.
For survival on atezolizumab in lung cancer and nivolumab in melanoma (\textbf{Figure~\ref{fig:interpretability}h,i}), the dominant tokens include high blood pressures, diverticular disease, kidney stones, pain and musculoskeletal diagnoses, as well as inflammatory skin conditions such as rosacea and hypertrophic skin disorders. Similarly, these patterns resonate with the facts that baseline comorbidity and functional status shape immunotherapy outcomes~~\cite{poletto2023predictive, yalcciner2025impact}, and with a growing literature linking cutaneous immune-related adverse events, including rosacea like eruptions and other inflammatory dermatoses to favourable survival on immune checkpoint inhibitors~~\cite{du2023cutaneous, cho2022cutaneous}. To further investigate the generalization of these findings, we check additional local and global examples, including 3-year mortality in metastatic lung cancer and IG profiles for ovarian cancer metastasis, anthracycline cardiomyopathy, survival after aortic stenosis, and stroke/embolism-free survival with apixaban, all of which show similar concentration of attribution on clinically plausible drivers, multimorbidity clusters and care-process proxies (\textbf{Extended Data \Cref{fig:edf_interpretability}}).

Beyond these expected and recently reported associations, IG consistently surfaces tokens that are best interpreted as proxies for care processes such as post-procedural states and device-related codes, markers of multimorbidity and shared risk environments such as psychiatric, musculoskeletal or gastrointestinal diagnoses in cardiometabolic panels, and a small set of features with no obvious mechanistic link. We therefore treat these attributions as associative structure learned from the EHR rather than causal effects: they indicate which aspects of a patient’s record most drive predictions under \ours, highlight concordance with established risk factors and newer epidemiological observations, and nominate a handful of features such as iron deficiency in COPD or skin/nail aging markers in heart failure as testable, hypothesis-generating signals for future clinical studies.

Together, the local and global analyses show that \ours{’}s predictions are auditable: at the bedside, clinicians can inspect why a patient’s risk rose when it did, and at scale, investigators can recover population-level structure that aligns with clinical knowledge while also revealing testable, hypothesis-generating signals. These properties are essential if foundation models are to support clinical decision-making and discovery rather than operate as opaque black boxes.

 \newpage
\Heading{Discussion} \label{sec:discussion}

In this work, we introduce \ours, the first multimodal foundation model that integrates the complete longitudinal trajectory of patient data into a unified temporal patient representation. By training and evaluating on over 25 billion medical events spanning clinical notes, imaging, and structured records from more than 7 million patients, we demonstrate that it is possible to consolidate these fragmented data modalities from unimodal data silos into a single computationally meaningful representation space. This unified representation can form the basis of AI-enabled precision medicine, empowering a vast spectrum of clinical tasks from early disease detection to treatment response stratification. \ours's virtual patient representations can continuously be updated with new clinical events, shifting healthcare from reactive, episodic treatment to proactive, continuous risk management. Moreover, these general-purpose representations also open new frontiers for clinical research. For example, by accurately identifying patients with similar complex multimodal phenotypes, \ours can readily enhance clinical trial matching, helping to resolve persistent bottlenecks in recruitment. For instance, APOLLO's semantic retrieval identifies clinically similar patients even when structured coding is incomplete, as demonstrated by retrieval of a kidney transplant patient lacking the corresponding ICD-10 code (\textbf{Figure \ref{fig:retrieval}}b). In the longer term, this capability could lay the groundwork for next-generation trial designs, including \textit{in silico} trial simulation at patient cohort level and highly personalized treatment response prediction at the individual level, potentially reducing the cost and duration of bringing new therapies to patients.

Beyond predictive utility, \ours provides a glimpse of the potential for biomedical discovery from unified multimodal data. The model's emergent embedding spaces, both at the input (medical event) and output (patient) level, reflects semantic relationships among individual medical concepts and patient-level phenotypes. The interpretability analyses, both at the patient level (via LOTO) and population level (via Integrated Gradients), provide some insight into model decision-making and lay the foundation for trustworthy application of multimodal risk-prediction models as an evolution of established clinical risk scores that use a limited set of manually curated features. Moving beyond manually curated feature sets is desirable not only because of the potentially higher accuracy as more input data is taken into account, but also because it provides a mechanism for identifying novel biomarkers and utilizing them for potential therapeutic benefit.

Despite the potential of this technology, much work remains to be done. First, being trained on strictly observational data, \ours learns to make predictions that are associational, not causal. For example, treatment response prediction as evaluated here represents prognostic stratification across patients rather than efficacy estimation across different drugs. In other words, the model may primarily learn to predict which patients will respond to a particular drug, but not which drug will be the best choice for a given patient. This is due to the characteristics of our training data: the former capability is learned from patient timelines of multiple comparable patients for any given drug (which we have) while the latter requires outcome data of multiple comparable drugs for the same patient (which are far more difficult to obtain). While valuable for patient stratification, care must be taken not to interpret retrieved risk factors as direct targets for intervention without further causal validation. The counterfactual task of predicting differential outcomes across therapies for the same patient will be an important extension of this work.

On the data side, although the training dataset spans 17 hospitals and affiliated clinics, all data come from a single umbrella healthcare system (Mass General Brigham) operating on a shared Epic EHR platform. Accordingly, the present evaluation reflects within-system generalization. The patient population captured by MGB, while heterogeneous in many respects, is still primarily representative of the northeastern United States. Incorporating multi-institution data from additional health systems and care settings during training will be an important direction for broadening the geographic and operational scope of the learned representations.

In the realm of risk prediction, the current benchmarking framework is designed to enable high-throughput evaluation across a broad and diverse set of 261 prognostic tasks, with age and sex serving as a common statistical reference. For well-studied endpoints such as heart failure, stroke, and acute MI, validated clinical risk tools (e.g., ASCVD, CHA\textsubscript{2}DS\textsubscript{2}-VASc, HEART score, Framingham) are the natural comparators. We do not perform head-to-head comparisons here, as computing these scores at scale requires endpoint-specific inputs that are not consistently captured as structured variables across the full cohort. However, our interpretability analyses recover many of the same risk factors embedded in these established scores (\textbf{\Cref{fig:interpretability}}), providing early evidence of mechanistic alignment and motivating direct comparisons in future work.

On the modeling side, our design choices are necessarily constrained by computational feasibility when dealing with long sequences of comprehensive multimodal patient trajectory data. As a result, we use frozen, pre-trained, off-the-shelf unimodal encoders for computational efficiency. Future iterations that fine-tune or pre-train these encoders directly on in-domain EHR data would likely yield superior and more efficient representations. Similarly, we encode high-density modalities (such as long clinical notes) using basic aggregation techniques, such as averaging document chunk embeddings. Implementing more sophisticated, learnable aggregation rules could unlock more fine-grained signals. Finally, our study does not explicitly incorporate data streams from wearables, information about lifestyle, or patient-provider conversations. Future work should address these limitations via additional architectural innovations.

Ultimately, the training paradigm and broad evaluation pipeline presented here demonstrate that converting the static EHR archive into unified, longitudinal virtual patient representations is feasible at healthcare-system scale, establishing a foundation for AI-enabled precision medicine across the full spectrum of clinical care.

\clearpage
\Heading{Online Methods} \label{sec:methods}
\vspace{1em} % Adds a bit of space after the main title
\newline
\heading{Dataset description}

We collected a large-scale, multimodal electronic health record (EHR) dataset, \dataset, from the Mass General Brigham (MGB) healthcare system. The dataset contains 7,155,044 patients with 25,296,943,893 clinical events in total across diverse modalities from 1992 to 2025, with the longest patient record spanning 33 years. \dataset is comprised of 28 distinct modalities, including structured records such as diagnosis, medications, lab tests, vital signs and flowsheet, as well as unstructured medical notes and imaging data. A detailed distribution of patient demographics and modalities of \dataset can be found in \textbf{Extended Data \Cref{tab:demo_basic,tab:modality_distribution}}. The dataset was split into training, validation, and testing splits with a ratio of 75:5:20. 

The MGB institutional review board approved the retrospective analysis of the EHRs. The participants were not directly involved or recruited for the study. The requirement for informed consent to analyze the EHR data was thus waived. 

\heading{Data acquisition and preprocessing}

\textbf{Source systems and extraction.} We extracted EHR data from the institutional analytics warehouse (Snowflake) using SQL queries against the operational Epic system. Raw tables were harmonized, de-identified according to institutional policy, and cleaned for format inconsistencies, duplicated rows, and implausible timestamps, before downstream processing.

\textbf{Modalities.} The corpus comprised (i) demographic data including sex, ethnicity, and date of birth, (ii) temporally ordered structured data including diagnoses coded, medications coded, laboratory tests coded, vital signs, and flowsheet measurements, and (iii) temporally ordered unstructured data comprising clinical text and medical images, spanning multiple services as shown in \textbf{Figure \ref{fig:overview}b-e}. Imaging data included  anatomic pathology whole slide
images, gross images, electron-microscopy images, and hematology blood smear images, where we used both raw images and their diagnostic reports. For the remaining imaging services,  only diagnostic reports were used. 

\textbf{Preprocessing.} For demographic data, we converted each patient’s record timestamp to minutes since birth and tokenized sex (male, female, unknown) as discrete categorical tokens. Ethnicity information in the EHR was recorded in a multi-select format where one patient could identify as multiple ethnicities. To handle this complication, we pre-embedded each atomic ethnicity value (e.g., ``Ethiopian") using \texttt{bert-base-uncased}~\cite{devlin2019bert} using the template ``Ethnicity: \{ethnicity\}". These contextual text embeddings allow us to take advantage of the world knowledge in existing pretrained general-purpose language models; in support of this hypothesis, when plotted via UMAP, the atomic ethnicity embeddings were loosely clustered by their associated geographic region. For patients who identified as multiple ethnicities, we took the mean of their atomic embeddings.

For structured data, we anchored each domain to standard clinical terminologies where possible: diagnoses were represented as ICD10 codes, medications were mapped to RxNorm ingredient identifiers, and vitals and laboratory tests were referenced using LOINC. Measurements with numeric values (laboratory tests, vital signs, and flowsheet entries) were first reconciled to a common unit per test, then transformed by $sign(x) \times log(|x|+1)$ to stabilize heavy tails. To reduce the influence of outliers while retaining ordinal structure, we discretized each test into ten equal-width (post-transform) bins, where the central eight covered the empirical $2.5^{th}$--$97.5^{th}$ percentiles, and the remaining two captured extreme lower and upper values. Measurements with categorical values were canonicalized to a compact, test-specific answer set using an agentic workflow: one large language model proposed a canonical label inventory for each test and a second model mapped free-text responses (for example, “pos”, “positive”, “+”) to those labels. After normalization, we tokenized structured observations as follows: each ICD diagnosis and each RxNorm ingredient became a token; each numeric measurement was represented by a (code, quantile-bin) token; and each categorical measurement by a (code, canonical-category) token. This procedure yielded a structured vocabulary of 235,768 unique tokens with learnable embeddings similar as in large language models.

Unstructured data were compressed to fixed-dimension embeddings using modality-specific pretrained encoders. All clinical text—including progress notes, diagnostic reports, and other note types were embedded with a pretrained clinical language model (Gatortron-base~\cite{yang2022gatortron}). Notes exceeding the model’s context window were segmented into contiguous chunks and encoded independently, where chunk embeddings were then averaged to obtain a single note-level representation. For pathology and hematopathology, we retained both reports and images: whole-slide images were embedded with TITAN~\cite{ding2025multimodal}, hematology blood smear images with DinoBloom~\cite{koch2024dinobloom}; and electron microscopy and gross images with CONCHv1.5~\cite{lu_visual-language_2024}. For non-pathology imaging services, we used the accompanying diagnostic reports (text) rather than pixel data, providing a consistent representation across specialties.

\heading{Pretraining}

\paragraph{Architecture.}
We modeled each patient as a single temporally ordered sequence of heterogeneous events, starting with their demographics. We embedded the patient's sex (in \{Male, Female, Unknown\}) by a learned embedding layer $W^{\text{sex}} \in \mathbb{R}^{3 \times E}$, where $E$ denotes the model's latent embedding dimension. The preprocessed ethnicity embedding was projected to the model's latent space with a two-layer multilayer perceptron (MLP): $P_{eth}:\mathbb{R}^{B}\!\to\!\mathbb{R}^{E}$, where $B=768$ denotes the embedding dimension of \texttt{bert-base-uncased}. Another two-layer MLP, $P_{age}:\mathbb{R}^{+}\!\to\!\mathbb{R}^{E}$, was used for encoding patient's age at last visit. The three demographics tokens, together with a learnable CLS token, were prepended to each patient's timestamped records. For event \(t\) with timestamp \(\tau_t\) (normalized to fraction of 100 years), we formed an input embedding \(z_t \in \mathbb{R}^{E}\) by combining a content embedding with a learnable time encoding that replaces positional embeddings. Tokenized structured events were embedded by a learned embedding layer \(W^{emb}:\mathbb{N}_0\!\to\!\mathbb{R}^{E}\). Let \(x^{(k)}_t \in \mathbb{R}^{d_k}\) denote the pre-extracted embedding for unstructured event in modality $k$ at time $t$, where $d_k$ is the embedding dimension of the modality-specific encoder for modality $k$. A lightweight projector $P_k:\mathbb{R}^{d_k}\!\to\!\mathbb{R}^{E}$ maps each unstructured feature to the model's latent space. A learnable time MLP $\phi_{\text{time}}:\mathbb{R}\!\to\!\mathbb{R}^{E}$ encodes \(\tau_t\) and is added to the content embedding. Thus, 
\begin{align}
z_t \;=\;
\begin{cases}
W^{\text{emb}}_{t}\,[\,\text{token}_t\,] \;+\; \phi_{\text{time}}(\tau_t) & \text{for structured modalities,}\\[4pt]
P_{k_t}\!\left(x^{(k_t)}_t\right) \;+\; \phi_{\text{time}}(\tau_t) & \text{for unstructured modalities,}
\end{cases}
\end{align}
where $W^{\text{emb}}[\text{token}_t]$ denotes indexing into the embedding matrix for the discrete token at time $t$.

Each patient's time-ordered sequence of events was then input into a multimodal temporal integration network, which is a Transformer encoder $f_\theta$ ~~\cite{vaswani2017attention}. We use the base configuration, with \(12\) transformer blocks, hidden layer dimension \(E{=}768\), $12$ attention heads of dimension $64$, and a \(4E\)-width MLP.

\paragraph{Masking and reconstruction objectives.}
Pretraining followed a multimodal masked modeling objective. For each patient, we randomly select a set of indices \(\mathcal{M}\) i.i.d. with probability \(\rho=0.3\), and replaced their input with learned mask vectors.

For structured tokens, reconstruction was formulated as classification within semantically coherent vocabularies. Masked structured tokens at index \(t\in\mathcal{M}\) were replaced by an observation-type–specific mask vector \(m_{o_t}\in\mathbb{R}^{E}\), and decoding was performed within each modality. For laboratory, vital, and flowsheet measurements, we further constrained decoding by subdomains defined by LOINC \texttt{CLASS}. Laboratory measurements are assigned to classes using the official LOINC mapping; vital signs and flowsheet tests are assigned to analogous classes using a large language model, Qwen-32B~~\cite{qwen3technicalreport}, to harmonize local nomenclature. Decoder weights for discrete vocabularies were tied to the corresponding input embeddings. Let \(h_t\) be the encoder state for a masked structured token and \(\mathcal{V}_m\) the relevant (sub)modality vocabulary; then the pretraining objective is given by the cross-entropy loss
\begin{align}
\mathcal{L}_{\text{struct}}
=\frac{1}{|\mathcal{M}_{\text{struct}}|}\sum_{t\in\mathcal{M}_{\text{struct}}}
-\log p(y_t \mid h_t),\qquad
p(\cdot\mid h_t)\in\Delta^{|\mathcal{V}_{m_t}|-1}.
\end{align}

For unstructured events, reconstruction was treated as regression to the original encoder embedding. Masked unstructured tokens were replaced by an input-type–specific mask vector \(m^{(u)}_{k_t}\in\mathbb{R}^{E}\). The masked sequence \(\tilde{Z}=\{ \tilde{z}_t\}\) was encoded to contextual states \(H = f_\theta(\tilde{Z})\), with \(h_t \in \mathbb{R}^{E}\). A small, encoder-specific prediction head maps the contextual state \(h_t\) back to the raw embedding space, and training minimizes a combination of mean-squared error and cosine distance:
\begin{align}
\mathcal{L}_{\text{unstruct}}
=\frac{1}{|\mathcal{M}_{\text{unstruct}}|}
\sum_{t\in\mathcal{M}_{\text{unstruct}}}
\left[
\frac{1}{d_{k_t}}\big\|\hat{x}^{(k_t)}_t - x^{(k_t)}_t\big\|_2^2
\;+\;
\Big(1 - \cos\!\angle\!\big(\hat{x}^{(k_t)}_t, x^{(k_t)}_t\big)\Big)
\right].
\end{align}
The total objective is \(\mathcal{L}=\mathcal{L}_{\text{struct}}+\mathcal{L}_{\text{unstruct}}\), computed as a global mean over masked tokens of each family.

\paragraph{Implementation details.}
We used a maximum sequence length of 1,536 events during pretraining. Sequences shorter than this were right-padded with an attention mask, and sequences longer than this were handled by uniformly sampling a contiguous window of 1,536 events per step.
Optimization used AdamW with gradient clipping (max norm 1.0) and a cosine with warmup learning rate schedule. To stabilize optimization under multi-objective supervision and heterogeneous token availability, we employed two learning-rate groups: a lower rate for the shared encoder and embedding parameters, and a higher rate for the modality-specific projectors and prediction heads. This choice mitigates high-variance updates from rare modalities while allowing small heads to adapt rapidly. Weight decay was applied to matrix-shaped parameters, and biases and LayerNorm weights were excluded. The model was trained for a maximum iterations of 30,000, with validating on the validation set every 1,000 iterations. The final model was selected based on the best validation loss. Training was done under distributed data parallelism on eight NVIDIA A100-80GB GPUs. Detailed hyperparameters used for model pretraining can be found in \textbf{Extended Data Table~\ref{tab:hparams_pretrain}}.

\paragraph{Patient representation.}
Unless otherwise specified, we summarize a patient’s longitudinal record as a single vector by appending a diagnosis, i.e., ICD-10, mask token at inference time and assigning it the time encoding of the patient’s last visit. The augmented sequence is passed through the encoder, and the contextual state of this masked ICD token is taken as the patient embedding. Operationally, this “query” token asks the model to predict the patient’s current diagnostic profile, which is one of the objectives used during pretraining, thereby encouraging the embedding to aggregate those aspects of the history most predictive of present disease burden. This vector is then used as the patient representation for downstream analyses. For patients with extremely long histories, we capped the raw sequence to the most recent 100,000 events.

\heading{Evaluation framework}

We conducted time-to-event (TTE) evaluations across five primary categories: new disease onset, disease progression, treatment response, adverse event, and hospital operations. We evaluated \ours on 95 disease onset tasks (\textbf{Extended Data \Cref{tab:task_meta-new_onset}}), 78 disease progression tasks (\textbf{Extended Data \Cref{tab:task_meta-dp}}), 59 treatment response tasks (\textbf{Extended Data \Cref{tab:task_meta-te}}), 17 adverse event tasks (\textbf{Extended Data \Cref{tab:task_meta-ae}}), and 12 hospital operations tasks (\textbf{Extended Data \Cref{tab:task_meta-ops}}). The metadata for each task can be found in the corresponding tables.

In TTE tasks, the model (in our case, fixed \ours patient embeddings and a linear Cox head) receives as input all patient features up to a \textit{snapshot time}, and aims to predict the duration between the snapshot time and a predetermined \textit{endpoint}. For disease onset tasks, our objective was to predict the risk of a particular disease at the time of a hospital discharge (snapshot) prior to its diagnosis (endpoint). A higher risk corresponds to a shorter time to diagnosis, and a lower risk corresponds to a longer time to diagnosis. For disease progression tasks, we set the snapshot as the diagnosis of the disease of interest, and the endpoint is set to a proxy for disease progression. These proxies can include the diagnosis of a more severe disease variant, administration of a medication implying more severe disease, or death. For treatment response and adverse event tasks, the snapshot is taken at time of first administration for a particular medication, and the endpoint is set to a proxy for treatment response, such as death or occurrence of an adverse event. We choose a TTE formulation for these tasks because they are subject to right-censorship, which occurs when a patient does not reach the endpoint as of the last event in their record. Censorship is influenced by both endpoint risk (i.e., low risk of reaching the endpoint means higher chance of censorship) as well as unrelated confounding factors (e.g., a patient moved to a different hospital).

To evaluate model performance, we compute a variety of time-dependent discrimination and calibration metrics at a pre-specified threshold duration of $\tau \in \mathbb{R}^+$ days (\textbf{Extended Data Tables \ref{tab:perf-new_onset}-\ref{tab:perf-ret}}). The threshold duration is chosen individually for each task to reflect an interval in which an intervention applied at time $0$ could plausibly change the outcome at timepoint $\tau$ (in days). For example, more chronic diseases such as heart disease may require a longer $\tau$ since interventions (e.g., lifestyle changes) may take a long time to manifest as appreciable changes in disease risk. On the other hand, more acute endpoints (e.g., mortality from COVID) require smaller $\tau$ since biological timescales are shorter. Practically, the choice of $\tau$ is also informed by the data distribution, such that sufficient cases are accumulated by time $\tau$ in order for metric calculations to be valid. For a detailed discussion of each metric, see \textbf{Metrics \& statistical analysis}.

When constructing the evaluation splits on the validation set of \dataset for each task, all valid snapshot and endpoint event times were extracted from each patient's medical record. We then performed the following filtering steps to arrive at a single snapshot-endpoint pair for each patient, so that each patient is represented no more than once in each task split. For disease onset tasks, all hospital discharges after the first 5 visits are initially selected as valid snapshots and the first diagnosis of the disease was selected as the endpoint. To increase incidence of the endpoint and mimic how a real-world predictive model might be applied only to higher-risk patients, we kept only snapshots within two standard deviations of the mean age of diagnosis for the disease of interest. For disease progression tasks, the first diagnosis of the disease was selected as the snapshot and the first occurrence of the endpoint was selected. For treatment response tasks, the first administration of the treatment was selected as the snapshot. We kept the absolute first occurrence of the endpoint for all tasks unless otherwise specified below (see \textbf{Task Definitions}). We discarded the patient if no valid snapshot events existed prior to the endpoint event. We also discarded patients where the endpoint event occurred more than 100 years after the snapshot event. Furthermore, to filter out instances where snapshot and endpoint events occur close together due to potential administrative delays in documenting the snapshot event, we enforce a blackout period $b$ where patients who experience the endpoint event within $b$ days of the snapshot event are discarded. The duration of the blackout period depends on the time course of disease as defined by $\tau$:
\begin{equation}
    b = 
\begin{cases} 
   30 & \text{if } \tau > 90, \\
   7   & \text{if } 60 \leq \tau < 90, \\
   1   & \text{if } \tau < 60. \\
\end{cases}
\end{equation}
We further applied inclusion criteria for some disease onset tasks based on patient sex (see \textbf{Task Definitions}). After all filtering steps, if more than one valid snapshot-endpoint pair is present, a random one is selected for each patient.

We trained separate Cox proportional hazards models for each TTE task to estimate the risk of the endpoint event as a function of covariates. We extract patient-level features using \ours, where the input consists of all events in a patient's medical record up to the snapshot time. For the structured modalities and last progress notes baselines (see \textbf{Baselines}), some patients may have zero valid events prior to the snapshot time; these patients are dropped to prevent them from adversely affecting baseline performance. For computational efficiency, the features are dimensionality-reduced using principal components analysis (PCA) to the top-50 principal components, which are used as the covariates $x$. The PCA projection is fit on the training set only (validation and test sets are transformed without refitting). The Cox model estimates a hazard function
\begin{equation}
    h(t|x) = h_0(t) \exp(\beta^{\top} x),
\end{equation}
where $h_0(t)$ is a baseline hazard and $\beta$ is the regression coefficients estimated from the training set. The resulting linear predictor $\beta^{\top} x$ provides a continuous risk score used to rank patients by their relative risk of experiencing the endpoint. We use the \texttt{CoxPHFitter} implementation from \texttt{lifelines 0.30.0}, with the penalizer $\lambda$ set to $10^{-4}$ as default and increased if necessary for convergence. Exact values of $\lambda$ used for each experiment are specified in \textbf{Extended Data \Cref{tab:perf-new_onset,tab:perf-dp,tab:perf-te,tab:perf-ae,tab:perf-ops,tab:perf-ret}}.

Because many tasks have relatively rare endpoint events, we employed case-cohort sampling to improve computational efficiency and mitigate extreme censorship in the training data. Specifically, we constructed case-cohort splits with a maximum ratio of 4:1 censored to noncensored individuals in the training set. This resampling alters the censorship ratio in the training set; however, the validation and test sets are left unmodified and retain the original, natural censorship ratio to ensure unbiased model evaluation. Because the Cox partial likelihood used for coefficient estimation is invariant to the baseline hazard, the modified training censorship does not bias the learned risk scores $\beta^{\top} x$. However, absolute survival probability estimates do depend on the baseline hazard, so this function was re-estimated on the unmodified validation set using the Breslow estimator prior to evaluation on the test set. This two-step procedure ensures that the regression coefficients benefit from the enriched case representation during training while the survival probability estimates reflect the true event distribution under the natural censoring regime.

\heading{Baselines}

\paragraph{Age–sex baseline.}
To quantify the value of learned representations against a minimal set of demographics predictors, we constructed frozen patient features using only age and sex as covariates. Age was defined as the patient’s age (in days) at the time of the last visit included in the input sequence; sex was encoded as a 3-level categorical variable {Male, Female, Unknown} with dummy indicators. This baseline represents the information available in textbook risk scores that rely solely on demographics. During TTE evaluation, PCA was omitted for this baseline.

\paragraph{Structured modalities baseline.} To better understand the contribution of unstructured modalities (notes and images) to a patient representation, we performed an ablation study by training a variant of \ours\ using only structured tokens (diagnoses, medications, labs, vitals, flowsheets) with the identical pretraining objective, tokenizer, and encoder as the full model. At inference, patient representations were computed from structured inputs only (using same ICD-10 mask-token prompt as for \ours). The resulting features are 768-dimensional, same as \ours.

\paragraph{Last progress notes baseline.} To assess the benefit of temporal integration relative to a single text-only snapshot, we represented each patient by the embedding of their most recent progress note (by GatorTron-base~~\cite{yang2022gatortron}) prior to the prediction time. The resulting features are 1024-dimensional.

\paragraph{Supervised baseline.}
To evaluate the value of self-supervised pretraining, we trained the same transformer architecture end-to-end for each task without pretraining. Instead of the masked ICD query used by \ours to form a patient embedding, we used the CLS token and attached a two-layer MLP that outputs discrete hazards over time bins. Let $h_j \in (0,1)$ denote the hazard for bin $j$ (obtained by a sigmoid on the head logits), and $S_j=\prod_{k\le j}(1-h_k)$ the discrete survival through bin $j$. For an individual with event indicator $\delta\in\{0,1\}$ and observed bin $y$, the negative log-likelihood is
\begin{align}
\mathcal{L}_{\text{NLL}} = -\delta\big[\log S_{y-1} + \log h_y\big] - (1-\delta)\log S_y,
\end{align}
as in~\cite{zadeh2020bias}. 

For each task with target horizon $\tau$ days, we defined the bin edges $0=t_0<t_1<\cdots<t_{J}<t_{J+1}=\infty$ as:
\begin{align}
{t_j} =
\begin{cases}
\{0\}\cup\{365i:i=1,\ldots,10\}\cup\{\infty\} & \text{if} \tau>365,\\
\{0,7,15,30,90,180,365,\infty\} & \text{if} \tau\le 365.
\end{cases}
\end{align}

The head outputs one hazard per interior bin $[t_{j},t_{j+1})$.

We trained each model on a single GPU with batch size of 128, learning rate of 1e-5, optimized using AdamW optimizer. Each model was trained for a maximum 40 epochs, with early stopping on the validation set if no improvement on the validation loss was observed for 5 epochs.

\heading{Retrieval}

The retrieval tasks are set up as follows: a cohort is defined using a set of inclusion criteria and patients satisfying the inclusion criteria are extracted from the evaluation set of 1.4M patients using SQL queries. Embeddings are extracted for all patients at a single calendar time (January 1st, 2025) to mimic the state of all patients on that day. Patients who are still alive as of that date have their medical records truncated such that only medical events prior to that date are included in the embedding. Patients who died prior to that date have all events in their medical history included in the embedding. For each task, the cohort is divided into five equal folds, and during each round of evaluation one fold (20\% of the cohort) is removed from the search index to be used as the queries. Retrieval metrics are calculated on this fold, and we repeat the process with the remaining 4 folds to obtain mean and standard deviation performance.
For the multimodal retrieval examples for text and image (\textbf{\Cref{fig:retrieval}c,d} and \textbf{Extended Data~\ref{fig:edf_retrieval}}) retrieval results, we retrieve from the entire evaluation set.

The inclusion criteria for each cohort take the form of \texttt{diagnosis THEN medication}, where the medication is an approved first-line therapy for the diagnosis. The full list of cohorts and their descriptions is presented in \textbf{Extended Data \Cref{tab:task_meta-ret}}.

\heading{Interpretability}

To interrogate which inputs drive task-specific risk, we computed Integrated Gradients (IG)~\cite{pmlr-v70-sundararajan17a} from each patient’s pre-Transformer input sequence to the scalar output of the corresponding Cox head. Formally, letting $s(\cdot)$ denote the task head applied to the Transformer encoder $f_\theta(\cdot)$, and $Z=\{z_t\}_{t=1}^{T}\in\mathbb{R}^{T\times E}$ the input sequence immediately before the multimodal temporal modeling, we estimated token-level attributions
\begin{align}
    \mathrm{IG}(Z) \approx (Z - Z_0)\odot\int_{0}^{1} \nabla_{Z} s\big(f_\theta\big(Z_0 + \alpha,(Z-Z_0)\big)\big)\ d\alpha,
\end{align}
where $Z_0$ is a baseline sequence of the same shape. The baselines were modality-consistent: for structured tokens we used the empirical mean of the learned embedding table (mean over the 235,768 unique structured tokens), and for unstructured inputs we used the all-zeros vector in the corresponding latent space. In present work, we primarily focus on risk factors, thus we restricted attribution to patients in the top quartile of predicted log-hazard for each task. This concentrates signal on clinically meaningful high-risk profiles and avoids diluting importances with low-risk trajectories.

IG yields an $E$-dimensional attribution vector per token occurrence. We summarized each occurrence by the net attribution (sum over embedding dimensions),
$a_t = \sum_{e=1}^{E} \mathrm{IG}_{t,e}.$ To mitigate scale differences across patients, we normalized each patient’s token scores by their $\ell_1$ sum:
$\tilde{A}_{p,v} = \frac{A_{p,v}}{\sum_{u} A_{p,u} + \varepsilon},$
with $\varepsilon > 0$ a small constant to avoid division by zero. For tokens that repeat within a patient, we applied occurrence-wise max pooling to obtain a single per-patient score per token:
$A_{p,v}=\max_{t\in \mathcal{T}(p,v)} a_t,$
where $\mathcal{T}(p,v)$ indexes all occurrences of token $v$ for patient $p$.

Population-level importance for token $v$ was computed as the mean of per-patient normalized scores across patients who had $v$ at least once:
\begin{align}
\bar{A}_{v} = \frac{1}{|P_v|}\sum_{p\in P_v} \tilde{A}_{p,v}.
\end{align}
To suppress idiosyncratic or extremely rare features, we required $v$ to meet a minimum prevalence in the analysis set: it must appear in $\geq 2.5\%$ of either uncensored (event) or censored patients for that task. Tokens failing this criterion were excluded prior to ranking.

For each task, tokens were ranked by $\bar{A}_{v}$ in descending order to produce a population-level list of risk-increasing features. Structured measurements are reported at the resolution used during modeling, i.e. (test, bin) for numeric data and (test, canonical category) for categorical entries, while diagnoses and medications are reported by their ICD-10 and RxNorm ingredient identifiers, respectively. Unstructured events were handled identically in the pipeline.

\heading{Metrics \& statistical analysis}

 For each metric, we report mean and 95\% confidence interval computed across 100 non-parametric bootstraps of the test set. Below we define each of the metrics used.

\textbf{Cumulative/dynamic AUC.} The cumulative/dynamic AUC quantifies the probability that, for a randomly selected pair of individuals (one who experiences the event before time $\tau$ and one who remains event-free beyond $\tau$) the predicted risk score is higher for the individual who fails earlier. To accommodate right-censored observations, this metric employs inverse probability of censoring weighting (IPCW), where the censoring survival function $\hat{G}(t)$ is estimated by the Kaplan-Meier curve of censoring times on the test set. Formally, this metric is represented by
\begin{equation}
    \mathrm{AUC}(\tau) \;=\; \frac{\sum_{i}\sum_{j} w_{ij} \, \mathbf{1}\{\hat r_i > \hat r_j\}}{\sum_{i}\sum_{j} w_{ij}},
\end{equation}
where $w_{ij}$ are IPCW pair weights (estimated from $\hat G$). We use the implementation from scikit-survival \newline (\texttt{sksurv.metrics.cumulative\_dynamic\_auc}).

\textbf{Balanced accuracy.} This is defined as the average of sensitivity and specificity for classifying whether a patient has reached the endpoint by $\tau$. Patients censored prior to $\tau$ are removed. The binary risk threshold is chosen to maximize balanced accuracy on the validation set, then the metric is computed on the test set.

\textbf{Concordance index (c-index).} We compute the truncated Uno's c-index as implemented in \texttt{sksurv.metri\newline cs.concordance\_index\_ipcw}, where $\tau$ is the truncation time. This metric provides a global summary of concordance over all pairs of events prior to $\tau$. For two patients who both experience the event prior to $\tau$, the c-index corresponds to the probability that the predicted risk score is higher for the patient who fails earlier. This metric uses the same censoring survival function $\hat{G}(t)$ as cumulative/dynamic AUC.

\textbf{Brier score.} Brier score quantifies the mean squared error (MSE) between predicted risk probabilities $\hat p_i(\tau)$ and the observed event indicator at $\tau$. To account for censoring we use the time-dependent IPCW Brier score implemented in \texttt{sksurv.metrics.brier\_score}:
\begin{equation}
\mathrm{Brier}(\tau) \;=\; \frac{1}{n}\sum_{i=1}^{n} w_i(\tau)\big(Y_i(\tau)-\hat p_i(\tau)\big)^2,
\end{equation}
where $Y_i(\tau)=\mathbf{1}\{T_i\le\tau, \delta_i=1\}$ is the event indicator by $\tau$, and the IPCW weight is
\begin{equation}
w_i(\tau)=\frac{\mathbf{1}\{T_i>\tau\}}{\hat G(\tau)}+\frac{\mathbf{1}\{T_i\le\tau,\ \delta_i=1\}}{\hat G(T_i^-)}.
\end{equation}
Here $\hat G(T_i^-)$ is the Kaplan–Meier estimate of the censoring survival function just before $T_i$, and $T_i,\delta_i$ are the observed time and event indicator for patient $i$. The IPCW weights correct the contribution of censored observations so that the sample mean approximates the expected squared error at $\tau$. This metric uses the same censoring survival function $\hat{G}(t)$ as cumulative/dynamic AUC.

\textbf{Integrated Calibration Index (ICI).} This is the mean absolute calibration error over the range of predicted probabilities. First we group the patients in the test set by predicted risk into 10 decile bins. For each bin $k$, we compute the mean predicted risk probability $\hat p_{(k)}$ and the Kaplan–Meier estimate of observed risk  $\widehat{\mathrm{KM}}_{(k)}(\tau)$. The ICI is given by
\begin{equation}
\mathrm{ICI} = \frac{1}{K}\sum_{k=1}^{K}\big|\widehat{\mathrm{KM}}_{(k)}(\tau)-\hat p_{(k)}\big|.
\end{equation}

\textbf{Maximum Calibration Error (MCE).} Similar to ICI, we compute MCE over the 10 bins of predicted risk. The maximum absolute calibration deviation across the range of predicted probabilities is given by
\begin{equation}
\mathrm{MCE} = \max_{k} \big|\widehat{\mathrm{KM}}_{(k)}(\tau)-\hat p_{(k)}\big|.
\end{equation}

% =============================
\heading{Computing hardware and software}

We used Python (version 3.12.11) and PyTorch (version 2.6.0, CUDA 12.4) (https://pytorch.org/) for all experiments and analyses in the study, which can be replicated using open-source libraries as below. For model pretraining, we used 8 GPUs configured for multi-GPU training using distributed data-parallel (DDP). Our transformer additionally used flash-attn (version 2.7.4) and ninja (version 1.11.1.4).

We used implementations, specifically PCA, StandardScaler and Logistic Regression from Scikit-learn (version 1.6.1) in our downstream tasks. We used  the lifelines library (version 0.30.0) for its implementation of the Cox proportional hazards model in survival tasks. Matplotlib (version 3.10.3) and Seaborn (version 0.13.2) were used to create plots and figures. Use of other miscellaneous Python libraries is detailed in the Reporting Summary.

% =============================
\heading{Data availability}

Following institution policies, all requests for data collected or curated in-house will be evaluated on a case-by-case basis to determine whether the data requested is compliant with intellectual property and patient privacy obligations. Data can only be shared for academic research purposes and will require a material transfer agreement.

\heading{Code availability}

Code for the \ours pretraining and evaluation pipeline, including risk prediction and patient retrieval, will be made available for academic research purposes upon publication.

\heading{Author contributions}

A.Z., T.D., S.J.W., F.M. conceived the study and designed the experiments. A.Z., T.D., L.P.L., C.T., D.M., A.M., R.P. curated the dataset and performed data preprocessing. A.Z., T.D., M.Y.L. developed the \ours model architecture and pretraining pipeline. A.Z., S.J.W., C.T. performed the embedding space and patient-level UMAP analyses. A.Z., T.D., S.J.W., C.T., J.E.L., R.P. curated and evaluated the prognostic downstream tasks. T.D., M.Y.L. performed the interpretability analysis. T.D., S.J.W. designed and conducted the multimodal retrieval experiments. A.Z., T.D., S.J.W., F.M. prepared the manuscript. All authors contributed to the writing. L.P.L., F.M. supervised the research.

\heading{Acknowledgments}

This work was funded in part by the Brigham and Women’s Hospital (BWH) and Mass General Hospital (MGH) internal funds.
 
\end{spacing}

%%%% References %%%%

\newpage
\begin{nolinenumbers}
\heading{References} 
\vspace{2mm}

\begin{spacing}{0.9}
\bibliographystyle{naturemag}
\bibliography{main}
\end{spacing}
\end{nolinenumbers}

%%%% Extended data figures %%%%

\clearpage
\setcounter{figure}{0}
\renewcommand{\figurename}{\textbf{Extended Data Figure}}

\clearpage
\setcounter{table}{0}
\renewcommand{\tablename}{\textbf{Extended Data Table}}

\begin{nolinenumbers}
\newpage
\heading{Extended Data Figures}

\begin{figure}[ht!]
    \centering
    \includegraphics[width=0.85\linewidth]{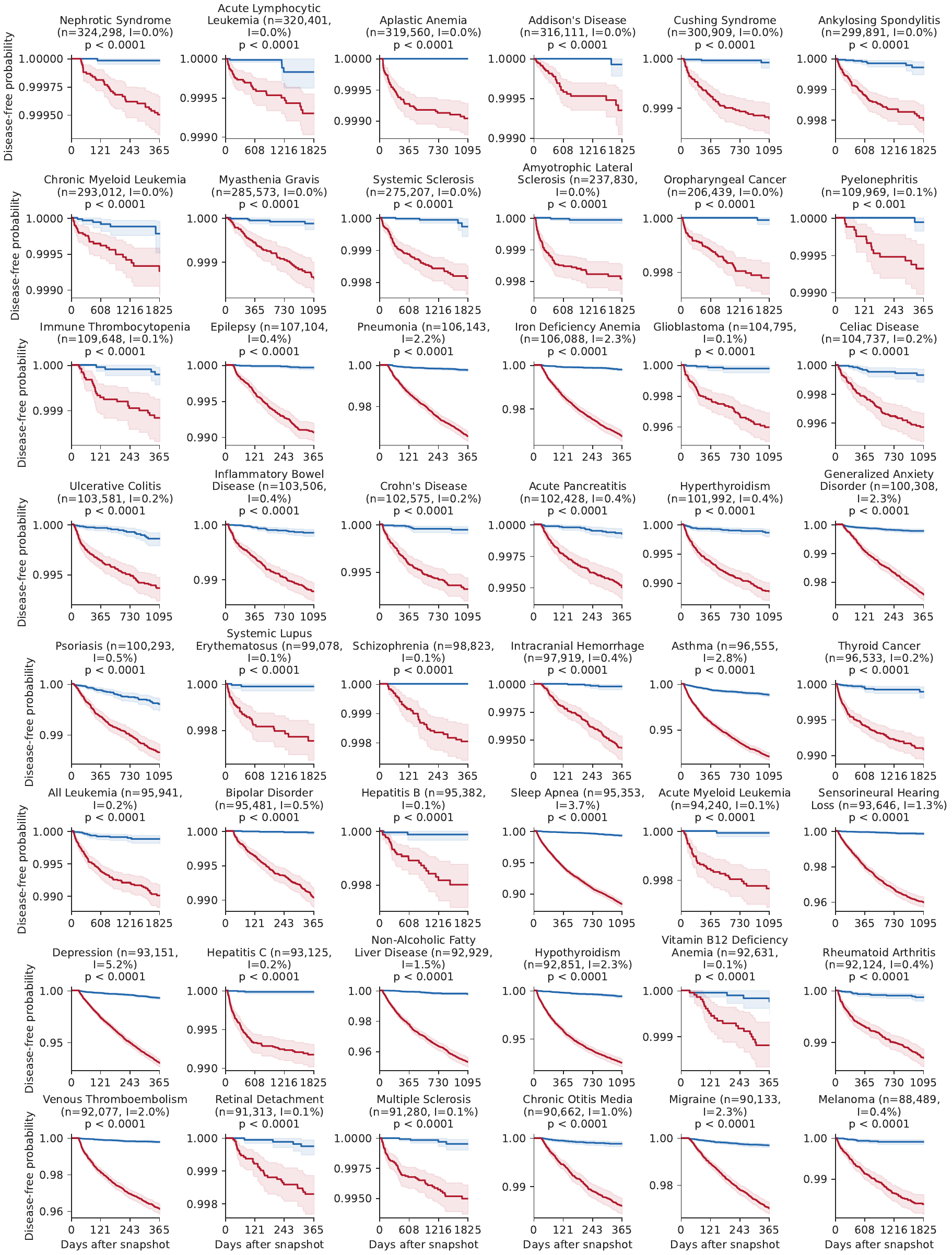}
    \caption{\textbf{Kaplan-Meier curves for new disease onset tasks} using the top 25$\%$ (red) and bottom 25$\%$ (blue) predicted risk scores. Tasks are sorted by number of events (n) in the test set per task. Statistical significance is given by the p-value for every task; n: number of events; I: label imbalance; $\tau$: task duration.}
    \label{fig:edf_km_new_0}
\end{figure}

\begin{figure}[ht!]
    \centering
    \includegraphics[width=\linewidth]{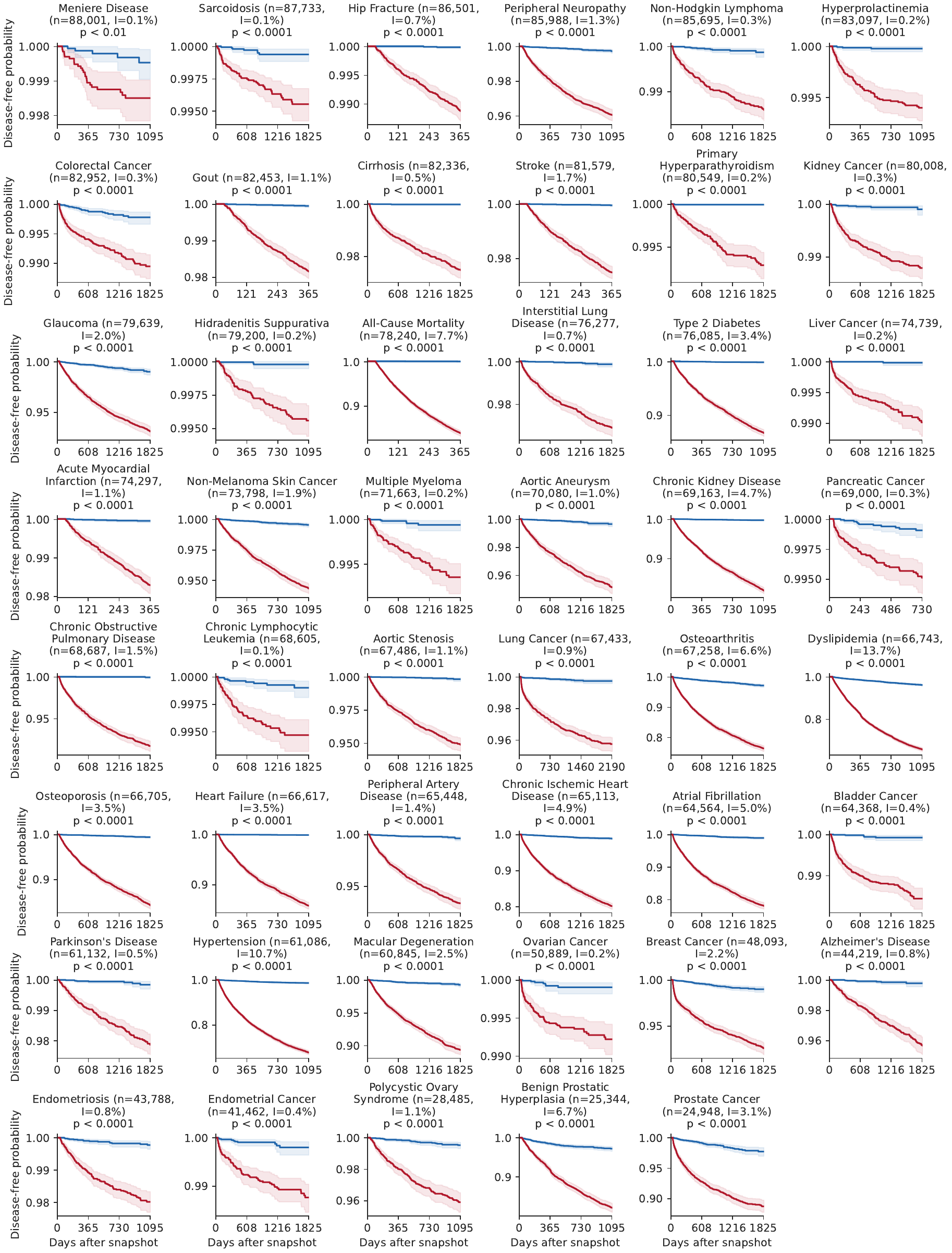}
    \caption{\textbf{Kaplan-Meier curves for new disease onset tasks.} Continued.}
    \label{fig:edf_km_new_1}
\end{figure}

\begin{figure}[ht!]
    \centering
    \includegraphics[width=0.96\linewidth]{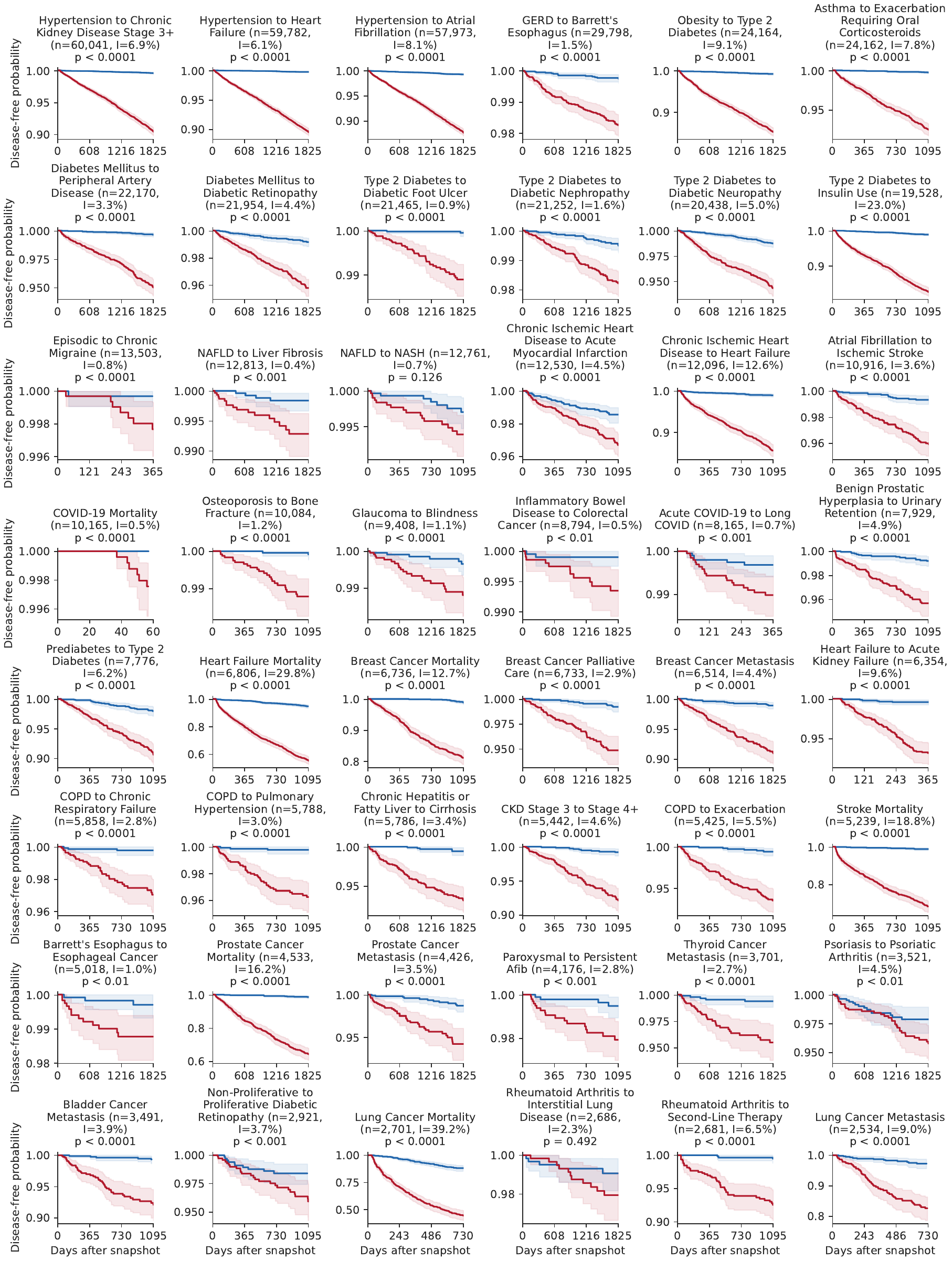}
    \caption{\textbf{Kaplan-Meier curves for disease progression tasks} using the top 25$\%$ (red) and bottom 25$\%$ (blue) predicted risk scores. Tasks are sorted by number of events (n) in the test set per task. Statistical significance is given by the p-value for every task; n: number of events; I: label imbalance; $\tau$: task duration.}
    \label{fig:edf_km_dp_0}
\end{figure}

\begin{figure}[ht!]
    \centering
    \includegraphics[width=\linewidth]{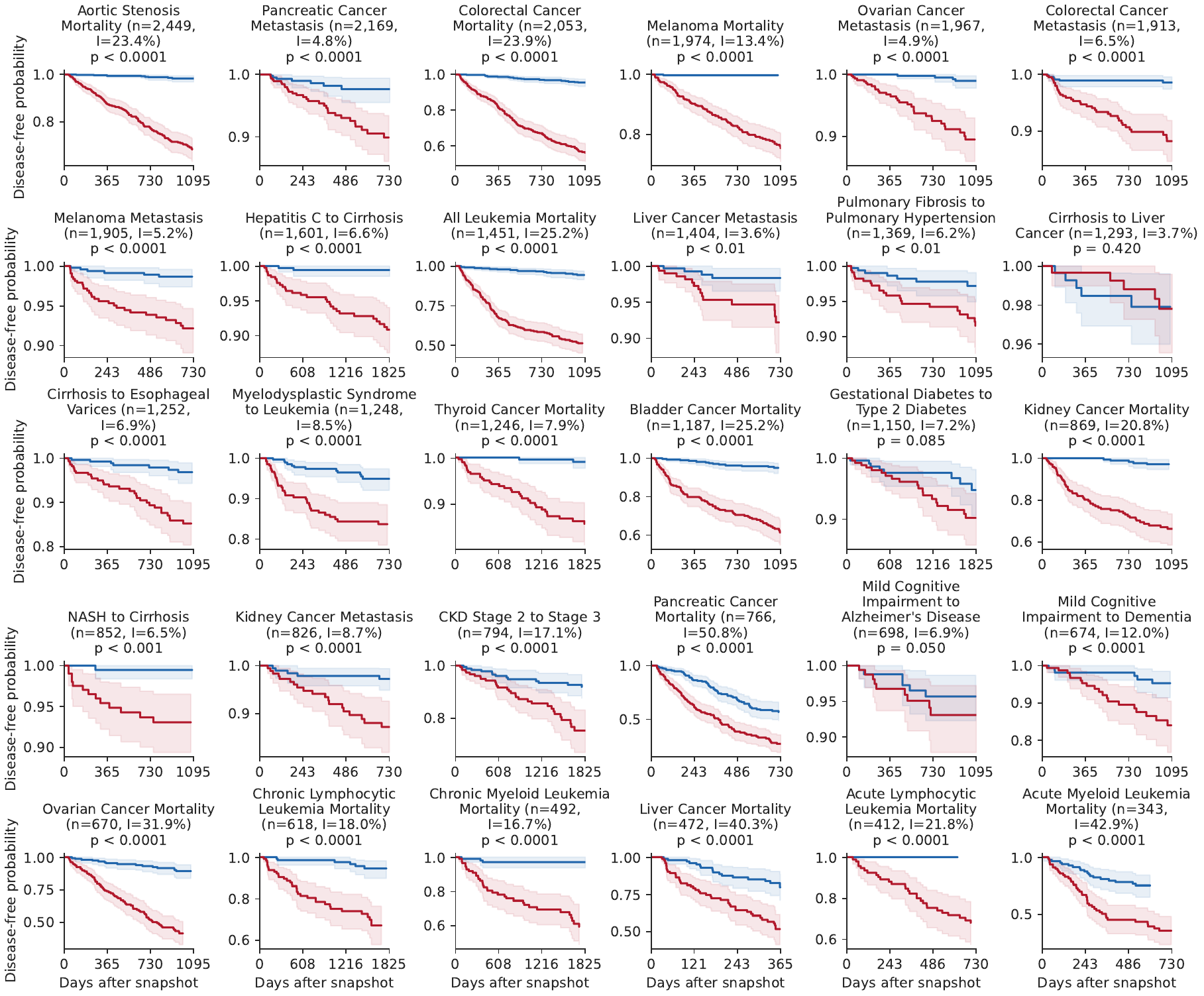}
    \caption{\textbf{Kaplan-Meier curves for disease progression tasks.} Continued.}
    \label{fig:edf_km_dp_1}
\end{figure}

\begin{figure}[ht!]
    \centering
    \includegraphics[width=0.96\linewidth]{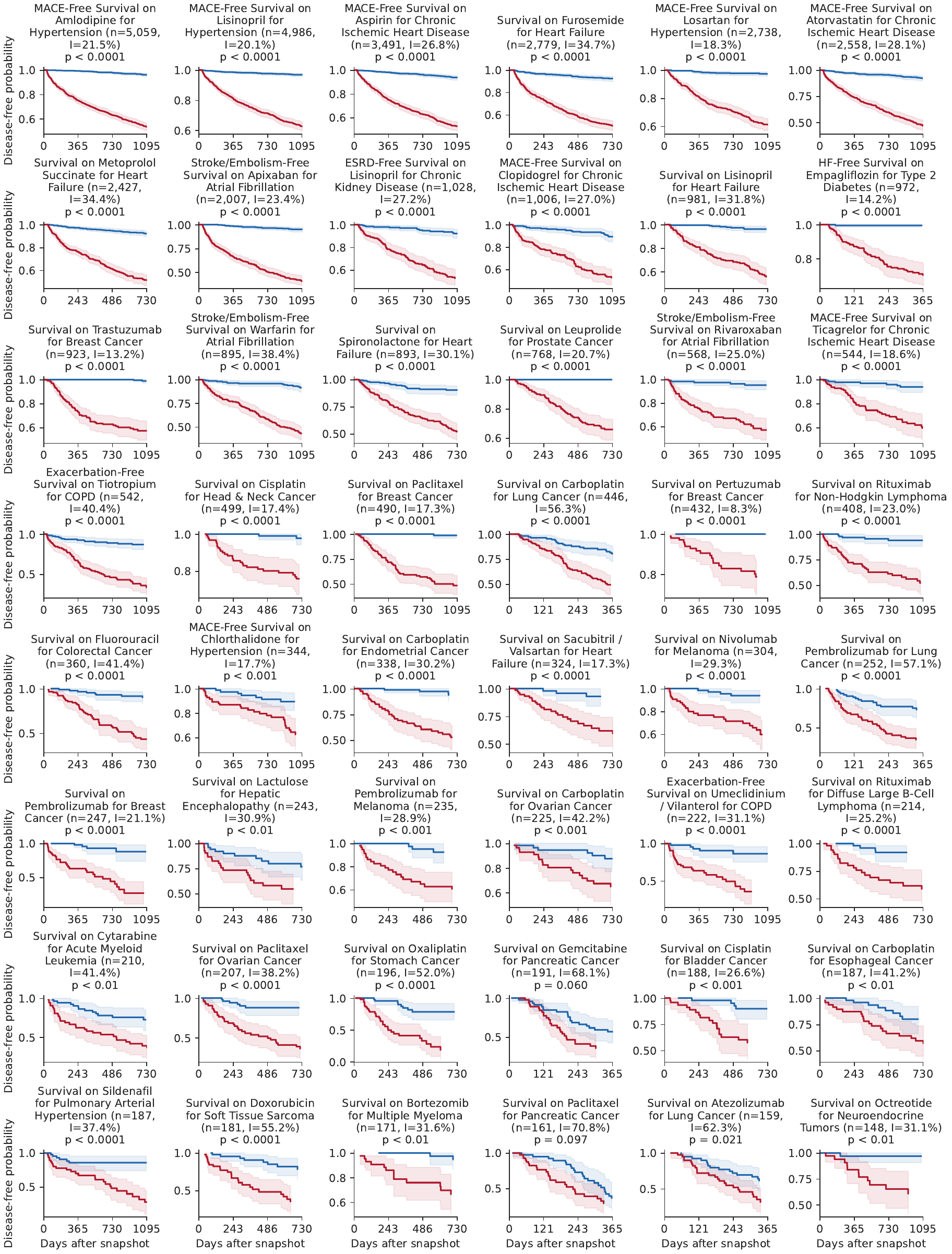}
    \caption{\textbf{Kaplan-Meier curves for treatment response tasks} using the top 25$\%$ (red) and bottom 25$\%$ (blue) predicted risk scores. Tasks are sorted by number of events (n) in the test set per task. Statistical significance is given by the p-value for every task; n: number of events; I: label imbalance; $\tau$: task duration.}
    \label{fig:edf_km_te_0}
\end{figure}

\begin{figure}[ht!]
    \centering
    \includegraphics[width=\linewidth]{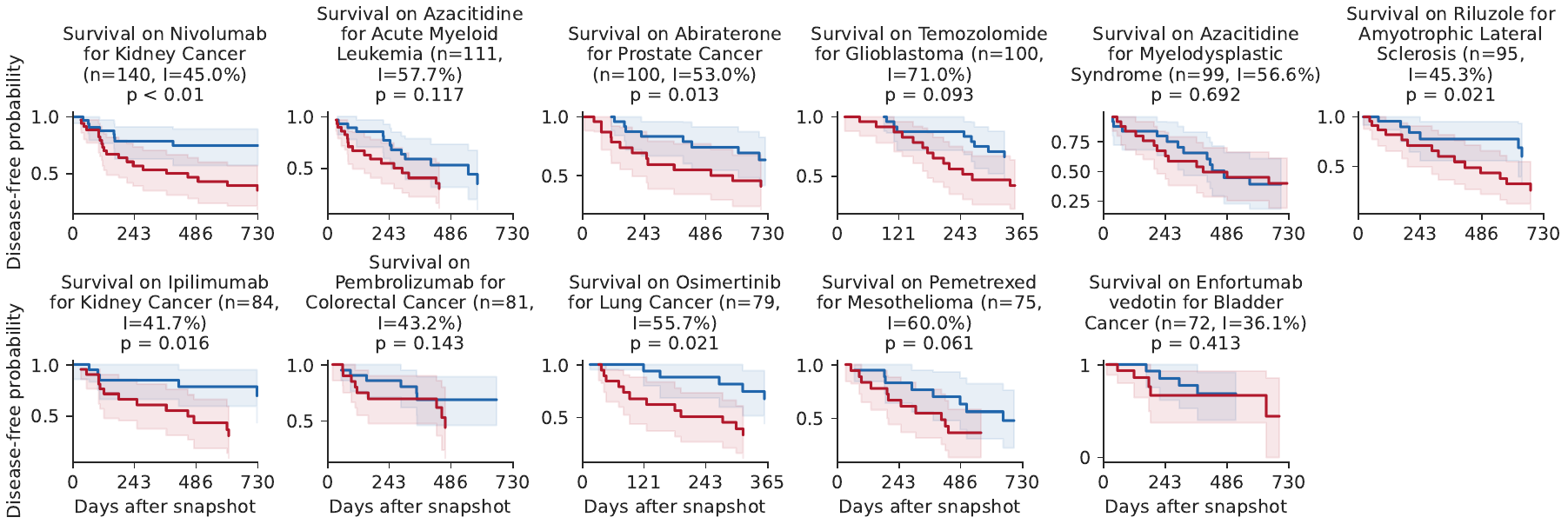}
    \caption{\textbf{Kaplan-Meier curves for treatment response tasks.} Continued.}
    \label{fig:edf_km_te_1}
\end{figure}

\begin{figure}[ht!]
    \centering
    \includegraphics[width=\linewidth]{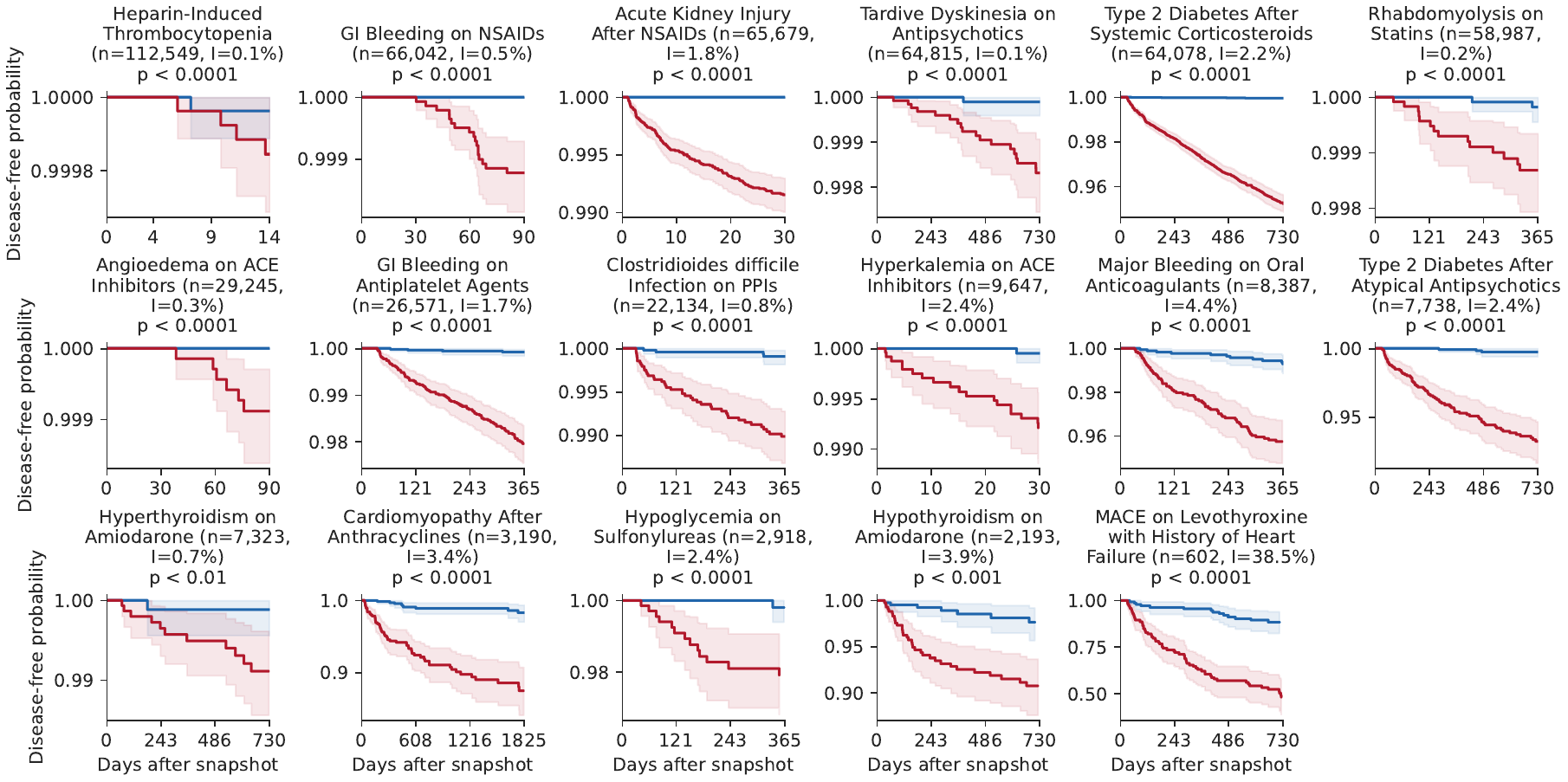}
    \caption{\textbf{Kaplan-Meier curves for adverse events tasks} using the top 25$\%$ (red) and bottom 25$\%$ (blue) predicted risk scores. Tasks are sorted by number of events (n) in the test set per task. Statistical significance is given by the p-value for every task; n: number of events; I: label imbalance; $\tau$: task duration.}
    \label{fig:edf_km_ae_0}
\end{figure}

\begin{figure}[ht!]
    \centering
    \includegraphics[width=\linewidth]{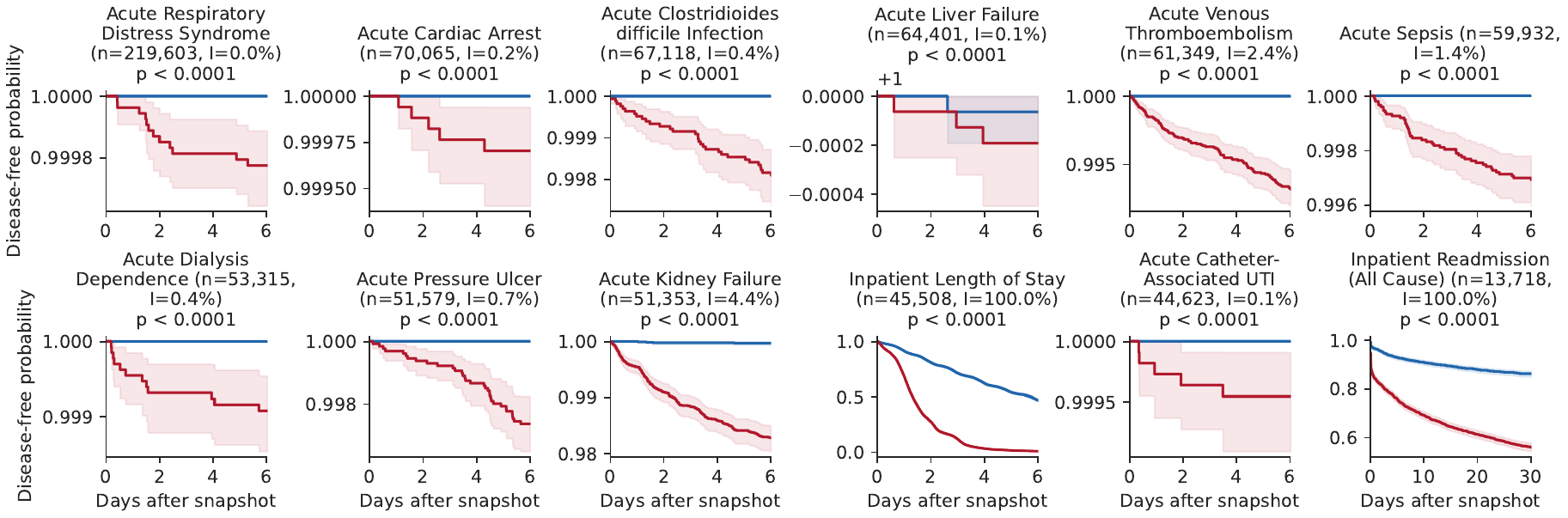}
    \caption{\textbf{Kaplan-Meier curves for clinical operation management tasks} using the top 25$\%$ (red) and bottom 25$\%$ (blue) predicted risk scores. Tasks are sorted by number of events (n) in the test set per task. Statistical significance is given by the p-value for every task; n: number of events; I: label imbalance; $\tau$: task duration.}
    \label{fig:edf_km_op_0}
\end{figure}

\begin{figure}[ht!]
    \centering
    \includegraphics[width=0.96\linewidth]{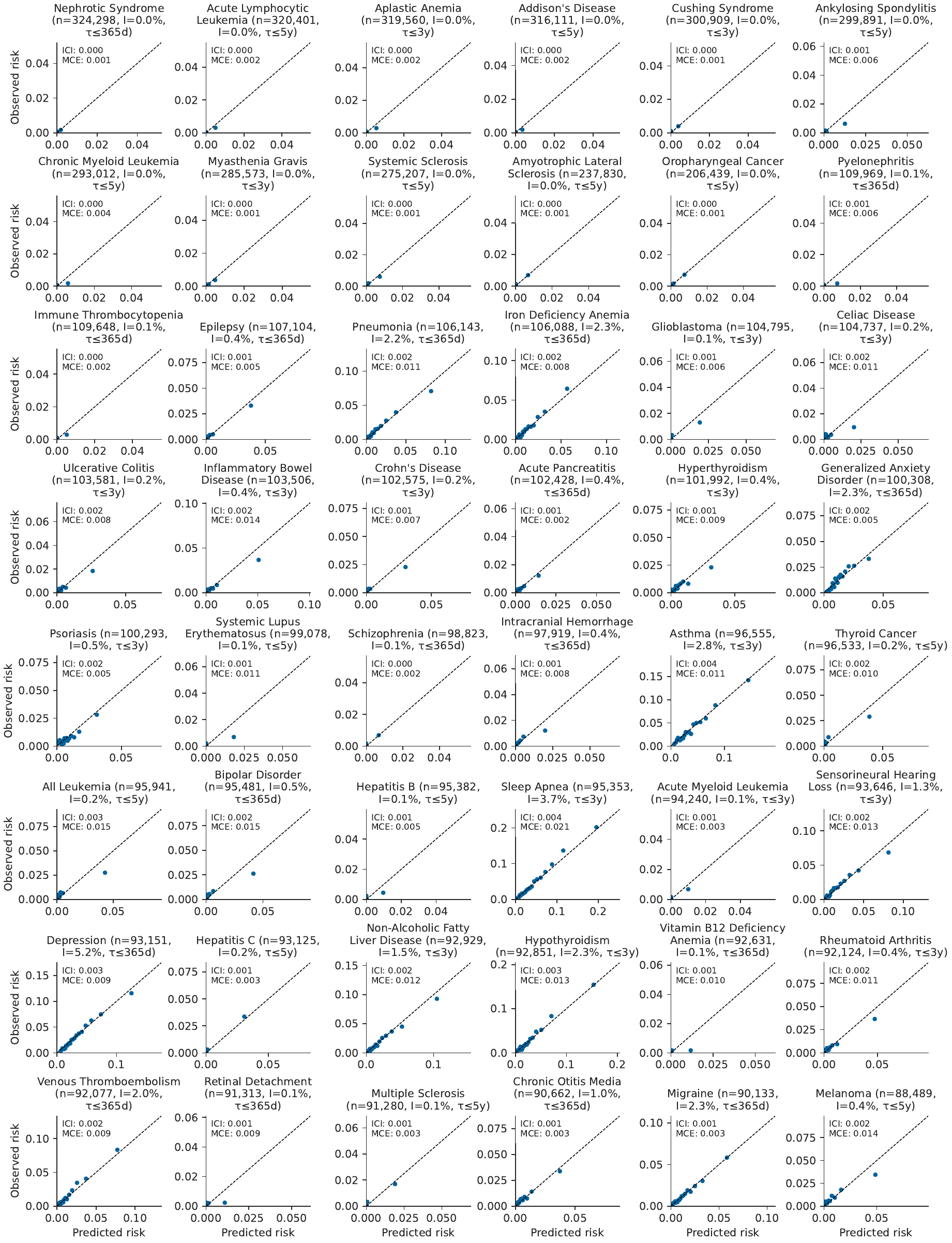}
    \caption{\textbf{Calibration curves for new disease onset tasks.} Predicted and observed risks are plotted in 20 bins, tasks are sorted by number of events (n) in the test set per task. To quantify the calibration performance, integrated calibration index (ICI) and mean calibration error (MCE) are given for every task; n: number of events; I: label imbalance; $\tau$: task duration.}
    \label{fig:edf_calib_new_0}
\end{figure}

\begin{figure}
    \centering
    \includegraphics[width=\linewidth]{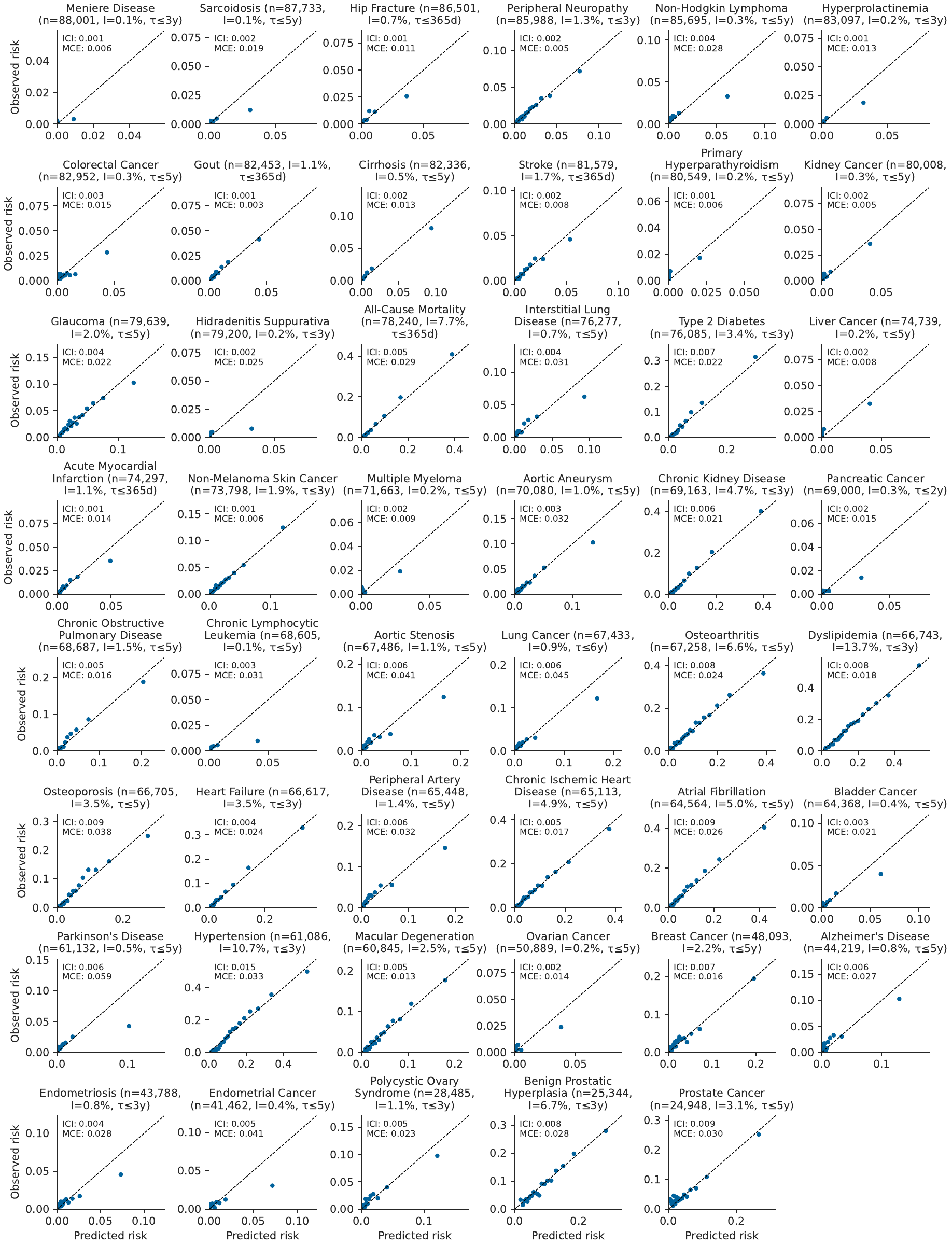}
    \caption{\textbf{Calibration curves for new disease onset tasks.} Continued.}
    \label{fig:edf_calib_new_1}
\end{figure}

\begin{figure}
    \centering
    \includegraphics[width=0.94\linewidth]{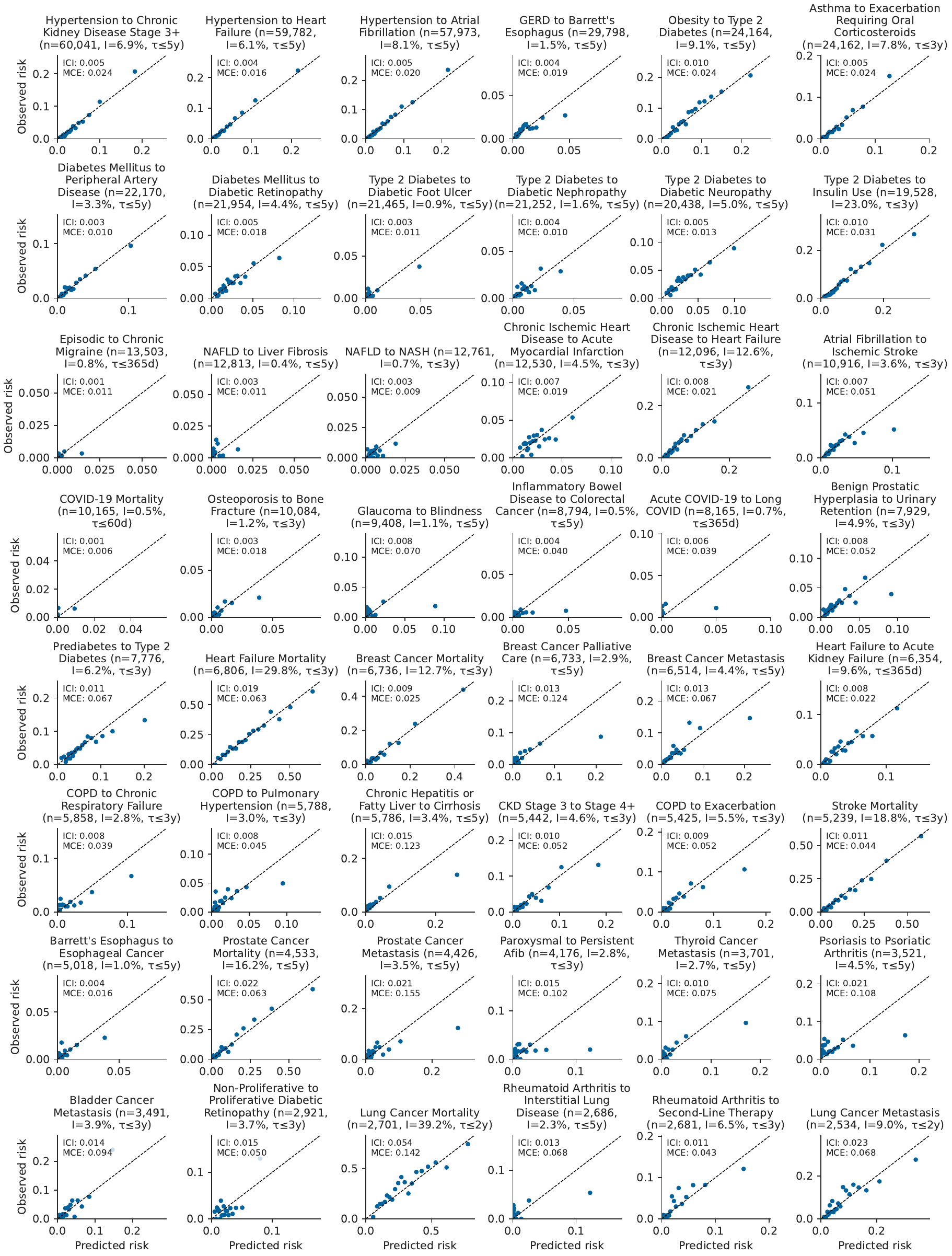}
    \caption{\textbf{Calibration curves for disease progression tasks.} Predicted and observed risks are plotted in 20 bins, tasks are sorted by number of events (n) in the test set per task. To quantify the calibration performance, integrated calibration index (ICI) and mean calibration error (MCE) are given for every task; n: number of events; I: label imbalance; $\tau$: task duration.}
    \label{fig:edf_calib_dp_0}
\end{figure}

\begin{figure}
    \centering
    \includegraphics[width=\linewidth]{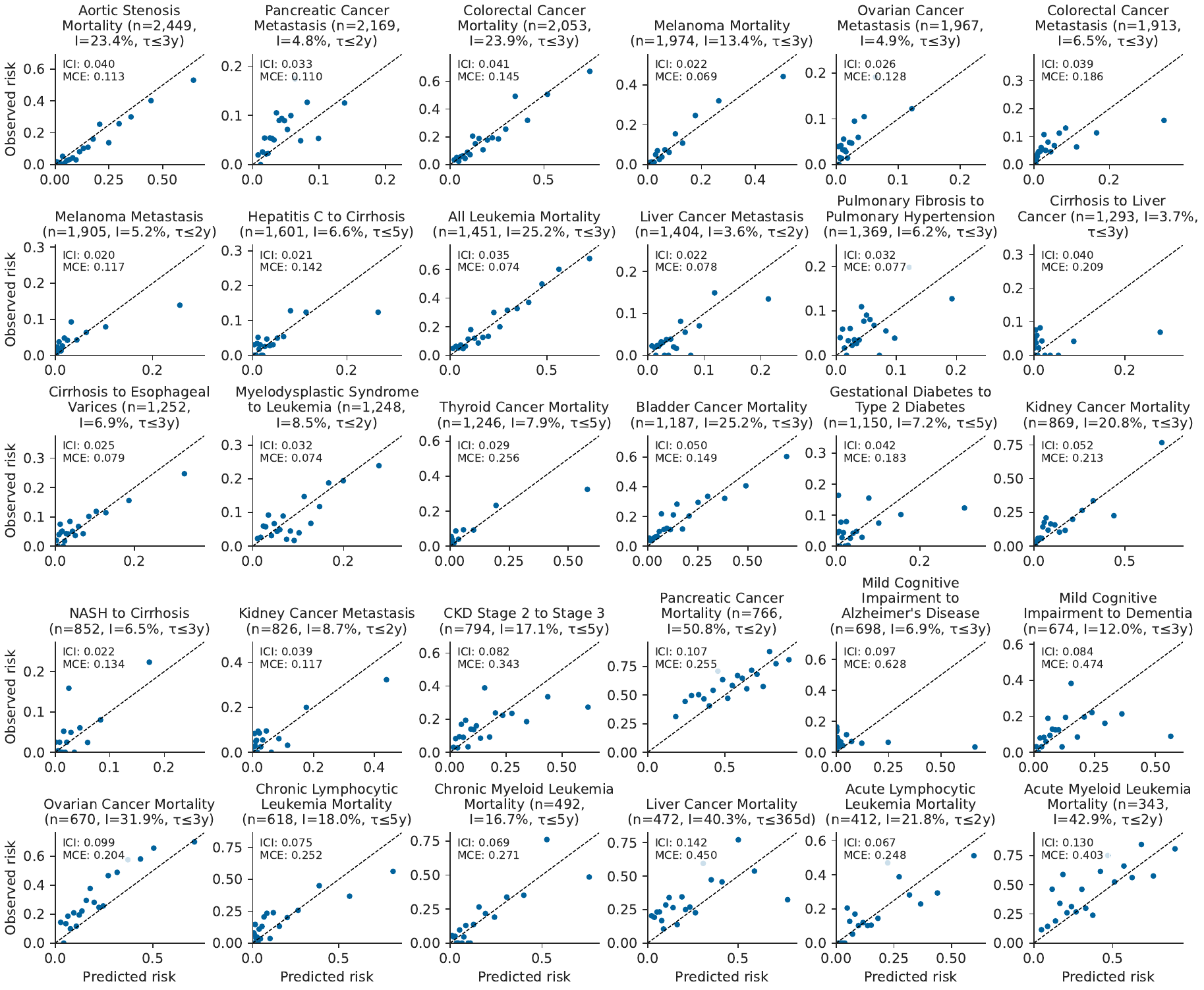}
    \caption{\textbf{Calibration curves for disease progression tasks.} Continued.}
    \label{fig:edf_calib_dp_1}
\end{figure}

\begin{figure}
    \centering
    \includegraphics[width=0.94\linewidth]{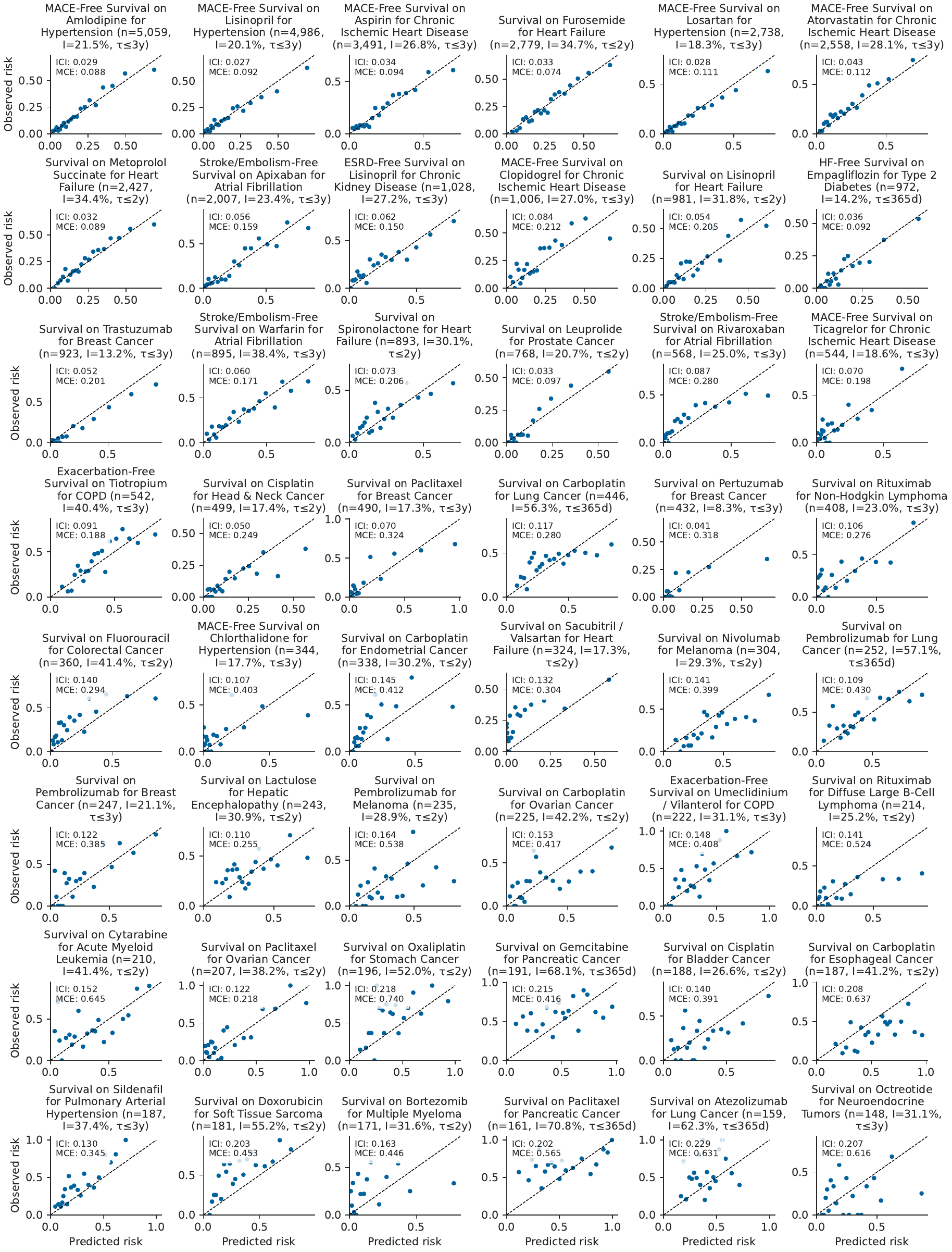}
    \caption{\textbf{Calibration curves for treatment response tasks.} Predicted and observed risks are plotted in 20 bins, tasks are sorted by number of events (n) in the test set per task. To quantify the calibration performance, integrated calibration index (ICI) and mean calibration error (MCE) are given for every task; n: number of events; I: label imbalance; $\tau$: task duration.}
    \label{fig:edf_calib_te_0}
\end{figure}

\begin{figure}
    \centering
    \includegraphics[width=\linewidth]{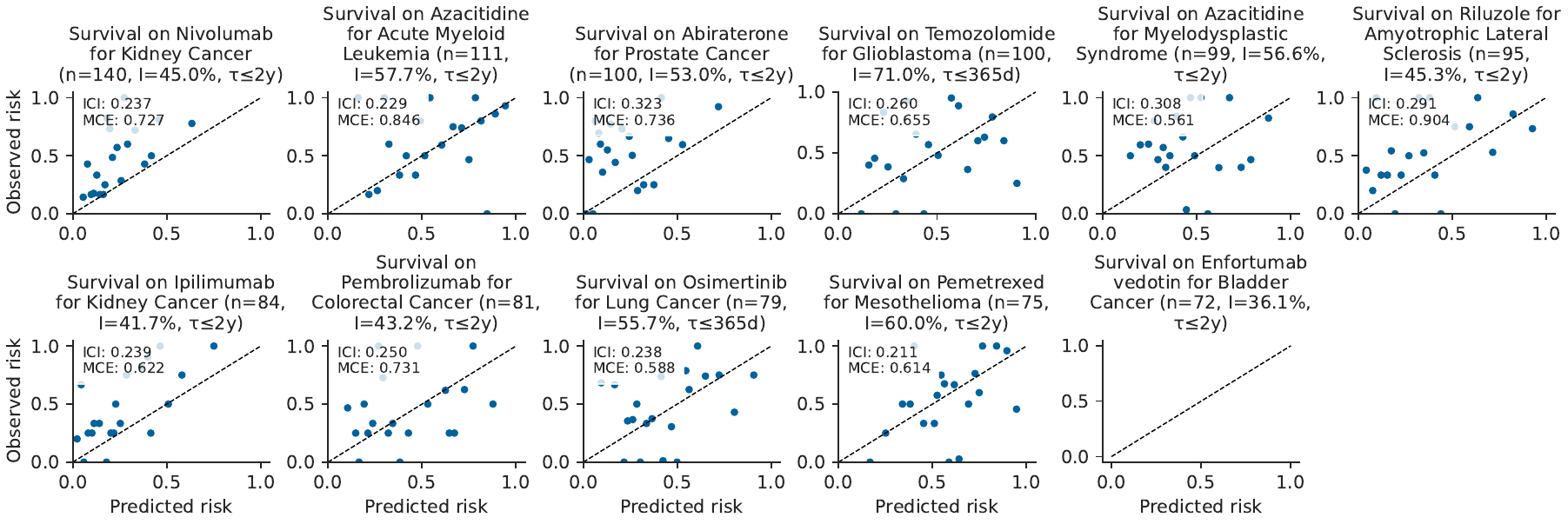}
    \caption{\textbf{Calibration curves for treatment response tasks.} Continued.}
    \label{fig:edf_calib_te_1}
\end{figure}

\begin{figure}
    \centering
    \includegraphics[width=\linewidth]{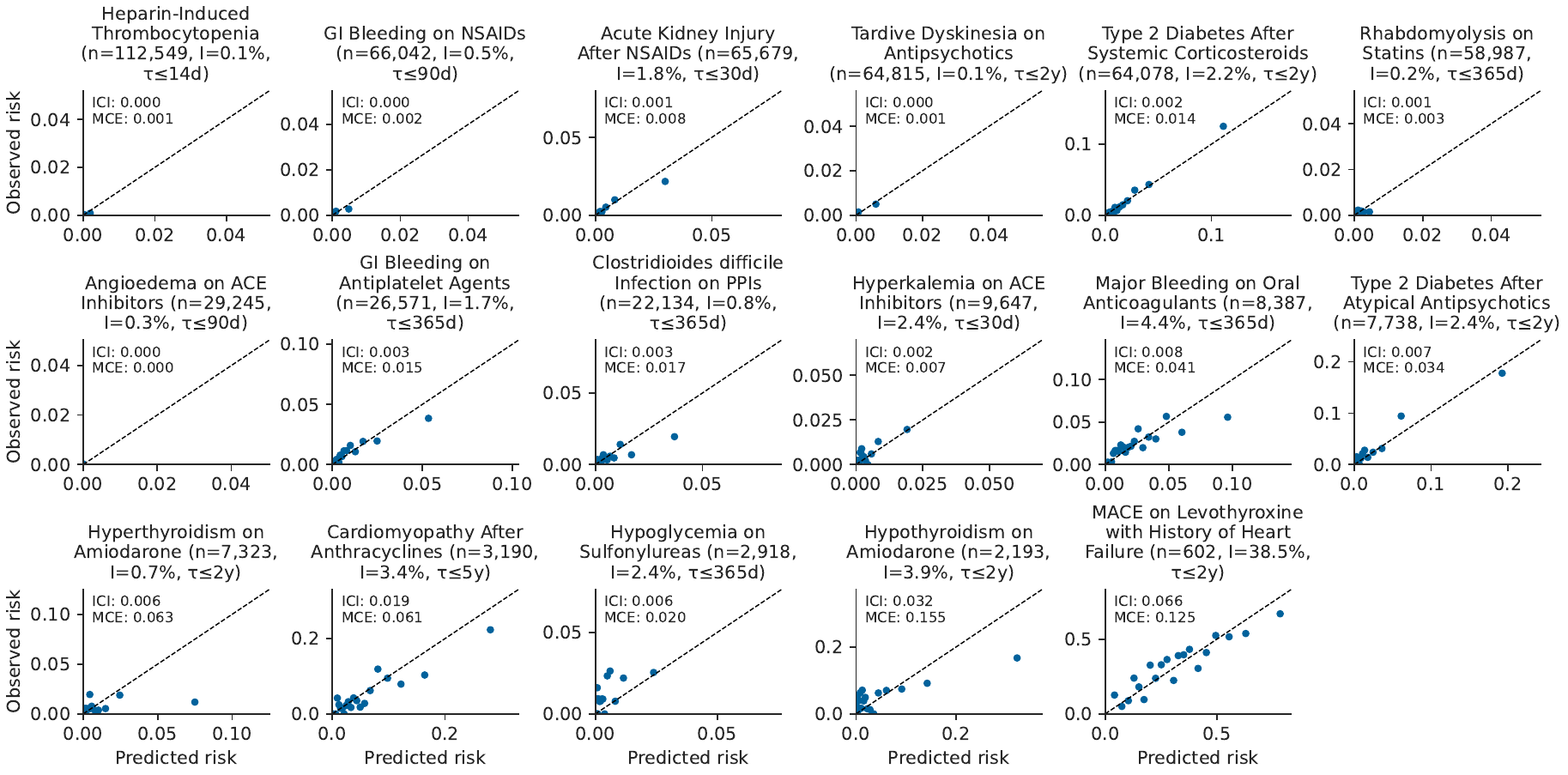}
    \caption{\textbf{Calibration curves for adverse events tasks.} Predicted and observed risks are plotted in 20 bins, tasks are sorted by number of events (n) in the test set per task. To quantify the calibration performance, integrated calibration index (ICI) and mean calibration error (MCE) are given for every task; n: number of events; I: label imbalance; $\tau$: task duration.}
    \label{fig:edf_calib_ae_0}
\end{figure}

\begin{figure}
    \centering
    \includegraphics[width=\linewidth]{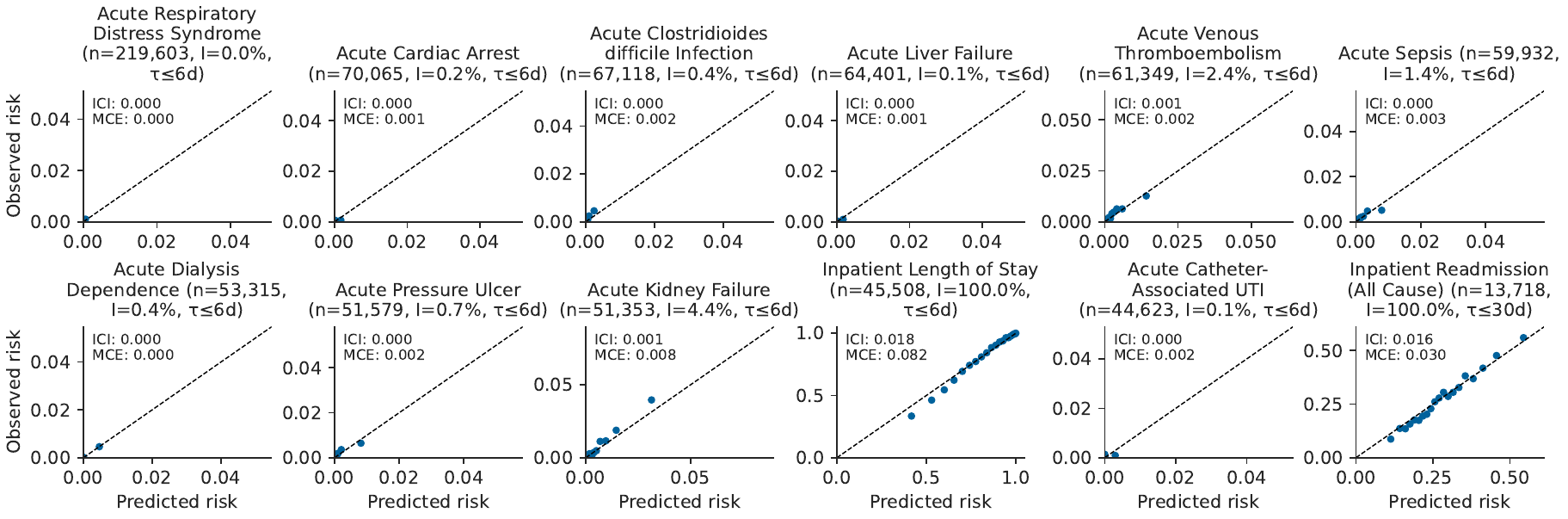}
    \caption{\textbf{Calibration curves for clinical operation management tasks.} Predicted and observed risks are plotted in 20 bins, tasks are sorted by number of events (n) in the test set per task. To quantify the calibration performance, integrated calibration index (ICI) and mean calibration error (MCE) are given for every task; n: number of events; I: label imbalance; $\tau$: task duration.}
    \label{fig:edf_calib_op_0}
\end{figure}

\begin{figure}[h]
    \centering
    \includegraphics[width=\linewidth]{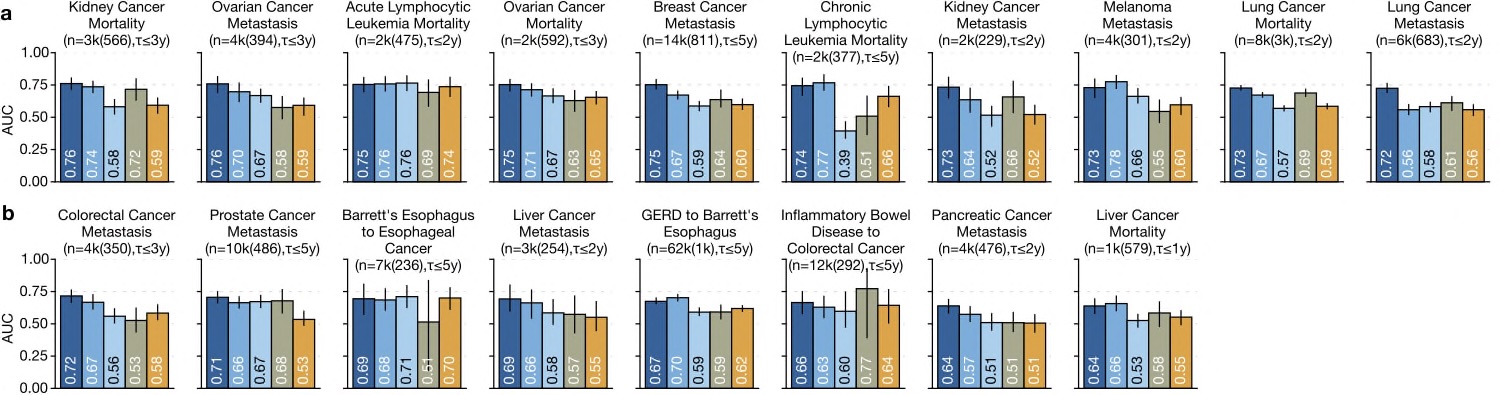}
    \caption{\textbf{Ablation study on architectural design choices of \ours,} evaluated on disease progression of neoplastic diseases. Top-10 tasks are shown in \textbf{\Cref{fig:results}i}, additional tasks shown here. Tasks are sorted by \ours’s performance.
$n$, number of patients (number of uncensored patients in parentheses); $\tau$, time-to-event.
}
    \label{fig:edf_ablation}
\end{figure}

\begin{figure}[h]
    \centering
    \includegraphics[width=\linewidth]{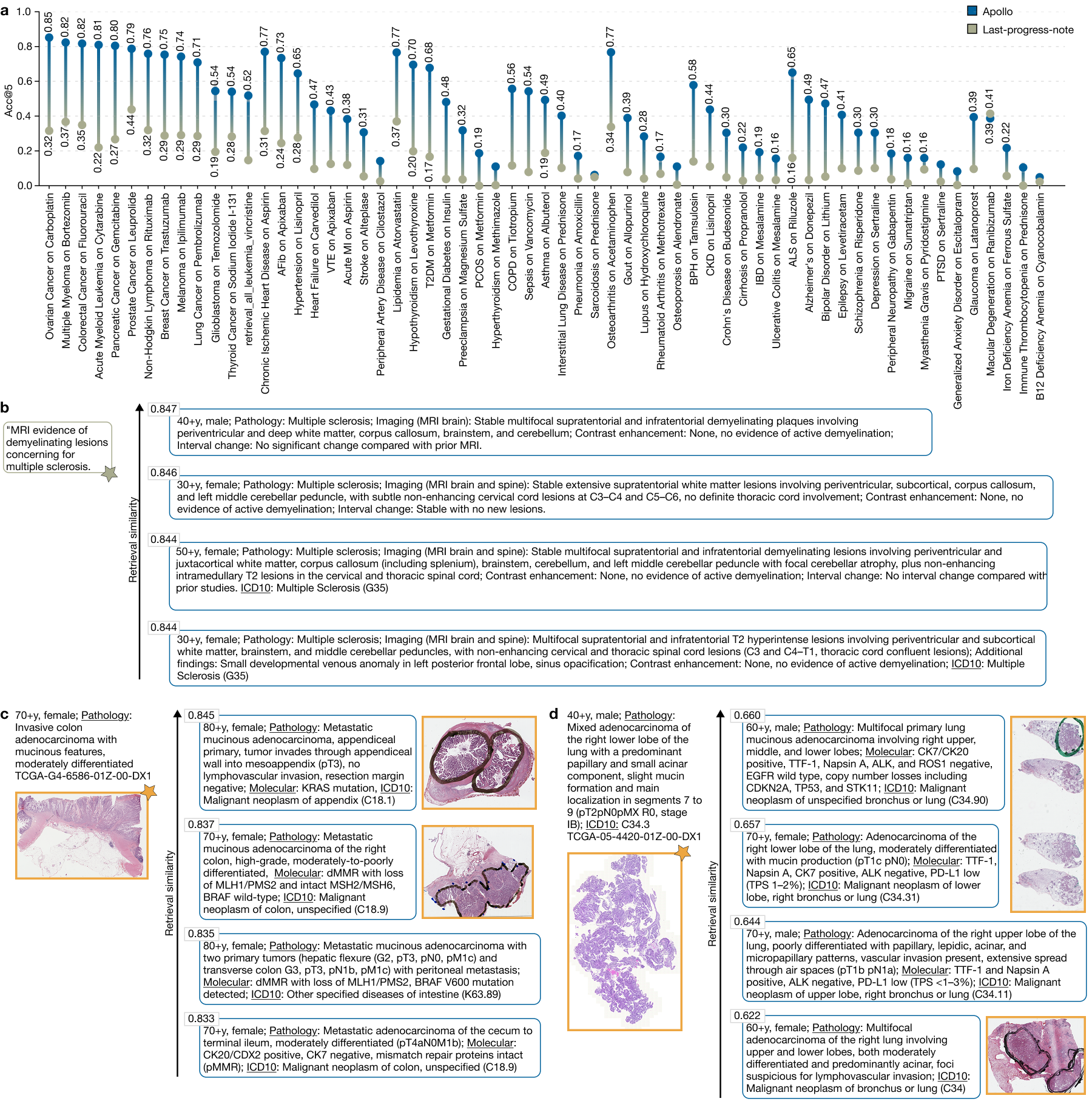}
    \caption{\textbf{Quantitative and qualitative retrieval performance}. (a) 61 patient retrieval tasks curated from combinations of ICD10 diagnosis codes and medications, assessed with accuracy among the five closest (Acc$@5$) embedded patients compared to retrieval using the last progress note embedding. (b)-(c) Top 4 retrieved patients using external TCGA slides for invasive colon adenocarcinoma (b) and lung adenocarcinoma (c). To preserve patient privacy, ages shown in the figure reported as ranges instead of exact numbers.}
    \label{fig:edf_retrieval}
\end{figure}

\newpage

\begin{figure}[h!]
    \centering
    \includegraphics[width=\linewidth]{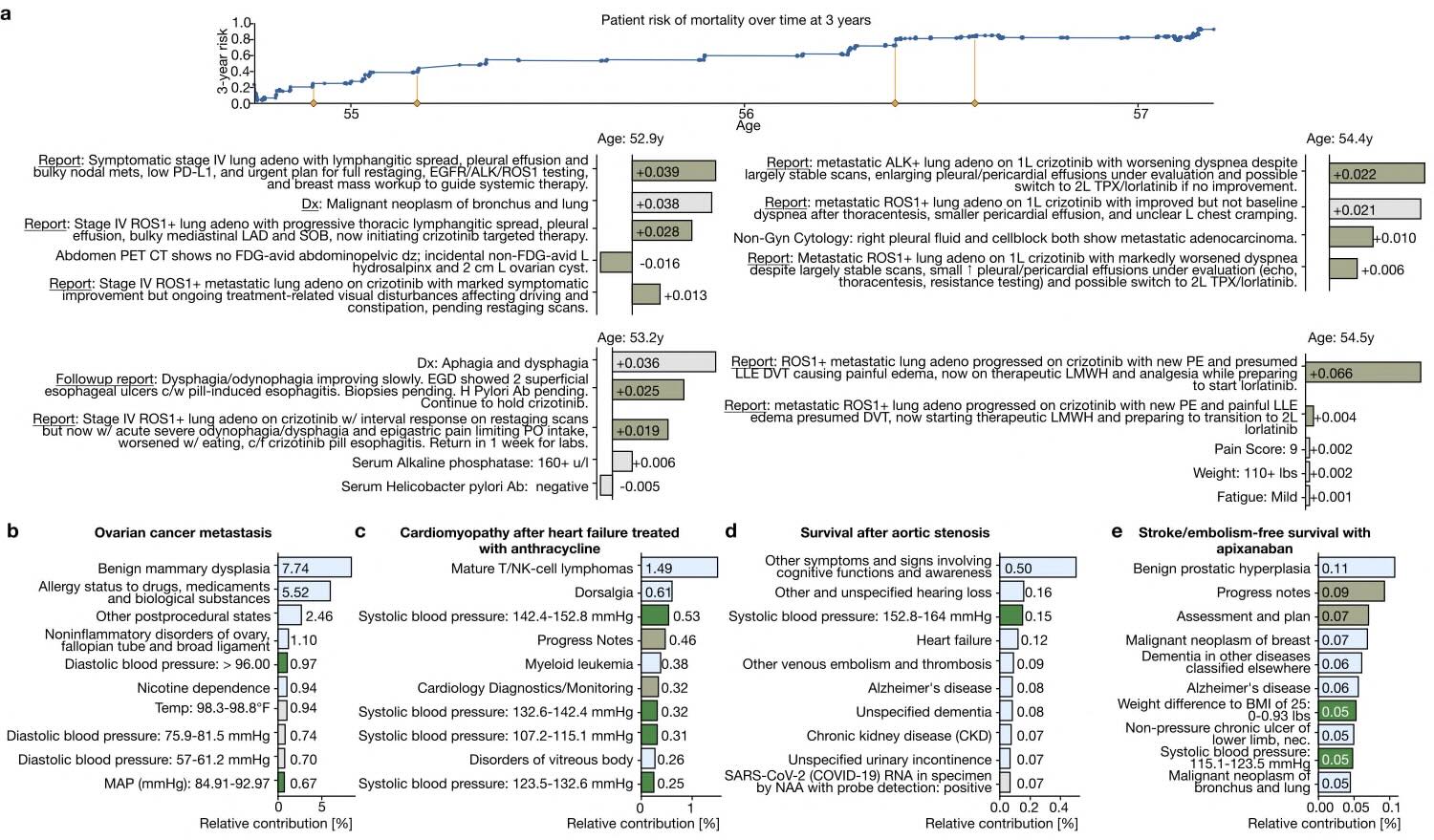}
    \caption{\footnotesize \textbf{Additional local and global interpretability examples.}
\textbf{(a)} Local analysis. Three-year mortality risk trajectory for a patient with metastatic ROS1/ALK-altered lung adenocarcinoma receiving targeted therapy. The largest contributors in each interval correspond to staging and restaging reports, thoracic imaging, treatment-response assessments, new thromboembolic events (pulmonary embolism and deep-vein thrombosis), and functional descriptors such as pain score, weight and fatigue. 
\textbf{(b–e)} Global analysis. Population-level feature importance for four additional downstream tasks—ovarian cancer metastasis, cardiomyopathy after heart failure treated with anthracyclines, survival after aortic stenosis, and stroke/embolism-free survival with apixaban—computed via Integrated Gradients (IG) as in Fig.~\ref{fig:interpretability}. For ovarian cancer metastasis (\textbf{b}), postprocedural states are consistent with the role of surgical and staging pathways in advanced ovarian cancer~~\cite{chase2024impact}. Breast-related terms such as benign mammary dysplasia plausibly act as a proxy for hereditary breast–ovarian cancer susceptibility and the downstream surveillance/diagnostic cascade, and BRCA1/2-associated HBOC confers markedly elevated ovarian and breast cancer risk~~\cite{PetrucelliDalyPal2025BRCA}.
Nicotine dependence is additionally linked to increased risk of mucinous ovarian cancer and has been associated with worse survival after ovarian cancer diagnosis~~\cite{praestegaard2017cigarette}.
For cardiomyopathy after heart failure treated with anthracyclines (\textbf{c}), the emergence of lymphoma and leukemia terms is clinically coherent because these diagnoses track anthracycline exposure and treatment intensity in real-world care. Anthracycline-related LV dysfunction is strongly influenced by cumulative dose and is amplified by patient-level risk factors such as older age, pre-existing cardiovascular disease, and cardiometabolic comorbidities~~\cite{neuendorff2020anthracycline}.
The specific appearance of myeloid leukemia aligns with recent cardio-oncology literature emphasizing that older AML populations treated with anthracycline-containing regimens have elevated cardiotoxicity risk~~\cite{neuendorff2020anthracycline}. 
For survival after aortic stenosis (\textbf{d}), top attributors concentrate on multimorbidities including cognitive impairment and dementia, sensory loss, urinary incontinence, heart failure, venous thromboembolism, CKD, and intercurrent infection~~\cite{wang2022impact}. 
CKD is also a well-established modifier of prognosis in severe aortic stenosis and after TAVR, with higher mortality and complications as kidney disease advances. For apixaban-treated stroke/embolism-free survival (\textbf{e}), several high-importance features including blood-pressure strata, obesity/weight patterns, dementia reflect baseline thromboembolic risk and frailty. Benign prostatic hyperplasia (BPH) also appears as the top risk factor, which is found to be associated with higher rates of hematuria-related complications in patients exposed to antithrombotics~~\cite{wallis2017association}. Clinically, hematuria complications can drive reduced adherence or discontinuation of anticoagulation, where both bleeding-related discontinuation and non-persistence have been linked to higher risks of stroke/systemic embolism and death in AF populations~~\cite{wallis2017association}. To preserve patient privacy, ages shown in the figure are randomly shifted by -2 to 2 years from their actual age, and
lab test and vitals values are presented as ranges instead of exact numbers.}
    \label{fig:edf_interpretability}
\end{figure}

\clearpage
\newpage
\newpage
\heading{Data Distributions}
\renewcommand{\arraystretch}{0.9}
\begin{table}[h]
\centering
\caption{\textbf{Cohort demographics.}}
\label{tab:demo_basic}
\begin{tabular}{llr}
\toprule
\textbf{Demographic} & \textbf{Category} & \textbf{Count} \\
\midrule
\multirow{3}{*}{Sex}
  & Male    & 3,915,625 \\
  & Female  & 3,236,747 \\
  & Unknown & 2,672\\
\midrule
\multirow{8}{*}{Age (years)}
  & 0--17    & 704,175 \\
  & 18--34   & 1,328,286 \\
  & 35--49   & 1,356,313 \\
  & 50--64   & 1,401,588 \\
  & 65--79   & 1,395,482 \\
  & 80--89   & 529,131 \\
  & 90--99   & 285,796 \\
  & $\geq$100 & 154,273 \\
\midrule
\textbf{Total} & & 7,155,044 \\
\bottomrule
\end{tabular}
\end{table}

\newpage

\newcolumntype{L}[1]{>{\raggedright\arraybackslash}p{#1}}

\begin{longtable}{L{0.12\textwidth} L{0.12\textwidth} L{0.42\textwidth} r}
\caption{\textbf{Modality distribution of \dataset.}}
\label{tab:modality_distribution}\\
\toprule
\textbf{Type} & \textbf{Modality} & \textbf{Detailed breakdown} & \textbf{Count} \\
\midrule

\multirow{5}{*}{Structured}
  & Flowsheet    & --- & 22,936,351,260 \\
  & Lab test     & --- & 1,442,631,333 \\
  & Vital signs  & --- & 505,193,723 \\
  & Medications  & --- & 112,189,178 \\
  & Diagnosis    & --- & 39,144,980 \\
\midrule

\multirow{27}{*}{Unstructured}
  & \multirow{8}{*}{Clinical notes}
      & Progress notes                                  & 158,683,290 \\
\end{longtable}

{
\centering

\begin{longtable}{lr}
\caption{\textbf{Data distribution of medications in \dataset.} Count of medication records per ATC category in the \dataset, sorted by number of records per category.} \label{tab:meds-counts} \\
\toprule
\textbf{ATC Category} & \textbf{Count} \\
\midrule
\endfirsthead
\midrule
\multicolumn{2}{r}{Continued on next page} \\
\midrule
\endfoot
\bottomrule
\endlastfoot
Systemic hormonal preparations, excluding sex hormones and insulins & 565,857 \\
Antiparasitic products, insecticides and repellents & 328,169 \\
Genito urinary system and sex hormones & 242,019 \\
Alimentary tract and metabolism & 235,168 \\
Nervous system & 181,774 \\
Cardiovascular system & 167,356 \\
Dermatologicals & 57,285 \\
Various & 30,181 \\
Respiratory system & 17,107 \\
Antineoplastic and immunomodulating agents & 9,770 \\
Blood and blood forming organs & 6,762 \\
Antiinfective for systemic use & 5,561 \\
Sensory organs & 5,285 \\
Musculo-skeletal system & 64 \\
Other & 110,336,820 \\
\end{longtable}

}

\newpage
{
\centering

\begin{table}[ht]
    \centering
    \caption{\textbf{Data distribution of diagnoses in \dataset.} Count of records per ICD10 chapter.}
    \label{tab:icd-counts}
    \begin{tabular}{lp{8cm}r}
    \toprule
    \textbf{\#} & \textbf{Chapter Name} & \textbf{Count} \\
    \midrule
    1 & Certain infectious and parasitic diseases & 698,584 \\
    2 & Neoplasms & 1,489,284 \\
    3 & Diseases of the blood and blood-forming organs & 668,249 \\
    4 & Endocrine, nutritional and metabolic diseases & 3,513,738 \\
    5 & Mental and behavioural disorders & 2,372,097 \\
    6 & Diseases of the nervous system & 1,595,885 \\
    7 & Diseases of the eye and adnexa & 1,482,031 \\
    8 & Diseases of the ear and mastoid process & 572,722 \\
    9 & Diseases of the circulatory system & 2,943,805 \\
    10 & Diseases of the respiratory system & 1,821,748 \\
    11 & Diseases of the digestive system & 2,081,121 \\
    12 & Diseases of the skin and subcutaneous tissue & 1,441,923 \\
    13 & Diseases of the musculoskeletal system and connective tissue & 3,671,049 \\
    14 & Diseases of the genitourinary system & 1,892,284 \\
    15 & Pregnancy, childbirth and the puerperium & 439,951 \\
    16 & Certain conditions originating in the perinatal period & 246,620 \\
    17 & Congenital malformations, deformations and chromosomal abnormalities & 243,414 \\
    18 & Symptoms, signs and abnormal clinical and laboratory findings & 5,375,320 \\
    19 & Injury, poisoning and certain other consequences of external causes & 883,940 \\
    20 & External causes of morbidity and mortality & 127,703 \\
    21 & Factors influencing health status and contact with health services & 5,451,213 \\
    22 & Codes for special purposes & 132,299 \\
    \bottomrule
    \end{tabular}
\end{table}

} 

\newpage
\heading{Training hyperparameters}

\begin{table}[h]
  \centering
  \begin{tabular}{p{7.5cm}|p{3cm}}
    \toprule
    \textbf{Hyperparameter} & \textbf{Value} \\
    \midrule
    Layers & 12 \\
    Heads & 12 \\
    Head activation & GELU \\
    Embedding dimension & 768 \\
    \midrule
    Mask ratio & 0.3 \\
    \midrule
    AdamW $\beta$ & (0.9, 0.999) \\
    Effective Batch size & 1024 \\
    Total training iterations & 30,000 \\
    Warmup iterations & 3,000 \\
    Learning rate schedule & Cosine \\
    Learning rate (start) & 0 \\
    Base encoder \& embedding layers learning rate (post warmup) & 2e-4 \\
    Projection layers \& prediction heads learning rate (post warmup) & 6e-4 \\
    Learning rate (final) & 0 \\
    Weight decay & 1e-5 \\
    Automatic mixed precision & fp16 \\
    \bottomrule
  \end{tabular}
  \caption{\textbf{Hyperparameters used in pretraining \ours.} 8 $\times$ 80GB NVIDIA A100 GPUs were used for training.}
  \label{tab:hparams_pretrain}
\end{table}

\newpage
\heading{Downstream evaluation}

{
\centering
\scriptsize
\begin{longtable}{p{4.5cm}p{4cm}ccccp{4cm}}
\caption{\textbf{Definition of new disease onset prediction tasks.} We evaluated TTE performance on 95 disease onset tasks. $\tau$ is the threshold duration. The snapshot is taken at time of random hospital discharge. For certain sex-specific diseases, we filter by patient sex as indicated in the Sex column. Incidence refers to the percentage of patients in the test set who reach the endpoint by the specified $\tau$. The Endpoint Event column specifies all ICD-10 codes used to identify the disease indicated in the Task column.} \label{tab:task_meta-new_onset} \\
\toprule
Category & Task & Sex & $\tau$ & N & Incid. & Endpoint Event \\
\midrule
\endfirsthead
\midrule
\multicolumn{7}{r}{Continued on next page} \\
\midrule
\endfoot
\bottomrule
\endlastfoot
Cardiovascular \& Circulatory & Acute Myocardial Infarction & - & 1y & 153,244 & 0.72\% & I21 \\
Cardiovascular \& Circulatory & Aortic Aneurysm & - & 5y & 144,435 & 5.64\% & I71 \\
Cardiovascular \& Circulatory & Aortic Stenosis & - & 5y & 139,345 & 6.05\% & I06.0, I06.2, I35.0, I35.2, Q23.0, Q25.3 \\
Cardiovascular \& Circulatory & Atrial Fibrillation & - & 5y & 145,377 & 23.53\% & I48 \\
Cardiovascular \& Circulatory & Chronic Ischemic Heart Disease & - & 5y & 146,743 & 22.98\% & I25 \\
Cardiovascular \& Circulatory & Heart Failure & - & 3y & 145,090 & 7.73\% & I50 \\
Cardiovascular \& Circulatory & Hypertension & - & 3y & 155,882 & 23.86\% & I10 \\
Cardiovascular \& Circulatory & Intracranial Hemorrhage & - & 1y & 198,214 & 0.24\% & I60, I61, I62 \\
\end{longtable}

} \newpage
{
\centering
\scriptsize
\begin{longtable}{p{4cm}cccccccc}
\caption{\textbf{Evaluation results of new disease onset prediction tasks.} We report time-dependent performance metrics computed at $\tau$ on the test set for 95 disease onset tasks. $\tau$ is the threshold duration. Arrows indicate whether higher ($\uparrow$) or lower ($\downarrow$) values are better. ICI, Integrated Calibration Index; MCE, Maximum Calibration Error. $10^{-4}$ is the default value of the penalizer ($\lambda$) for Cox proportional hazards; if convergence cannot be achieved then $\lambda$ is increased to 1e-3 ($^*$) or 1e-2 ($^{**}$). Mean and 95\% CI reported over 100 bootstraps of the test set.} \label{tab:perf-new_onset} \\
\toprule
Task & $\tau$ & Model & AUC ($\uparrow$) & Bal. Acc. ($\uparrow$) & C-Index ($\uparrow$) & Brier Score ($\downarrow$) & ICI ($\downarrow$) & MCE ($\downarrow$) \\
\midrule
\endfirsthead
\midrule
\multicolumn{9}{r}{Continued on next page} \\
\midrule
\endfoot
\bottomrule
\endlastfoot
Acute Myocardial Infarction & 1y & Age-Sex & \makecell{0.675\\[-2pt]\tiny (0.645-0.703)} & \makecell{0.641\\[-2pt]\tiny (0.612-0.663)} & \makecell{0.671\\[-2pt]\tiny (0.642-0.700)} & \makecell{\textbf{0.006}\\[-2pt]\tiny (0.005-0.006)} & \makecell{\textbf{0.001}\\[-2pt]\tiny (0.001-0.002)} & \makecell{\textbf{0.005}\\[-2pt]\tiny (0.003-0.008)} \\
 &  & Last Progress & \makecell{0.715\\[-2pt]\tiny (0.685-0.743)} & \makecell{0.653\\[-2pt]\tiny (0.624-0.682)} & \makecell{0.713\\[-2pt]\tiny (0.684-0.741)} & \makecell{\textbf{0.006}\\[-2pt]\tiny (0.005-0.006)} & \makecell{\textbf{0.001}\\[-2pt]\tiny (0.001-0.002)} & \makecell{0.006\\[-2pt]\tiny (0.003-0.009)} \\
 &  & \ours & \makecell{\textbf{0.817}\\[-2pt]\tiny (0.793-0.838)} & \makecell{\textbf{0.740}\\[-2pt]\tiny (0.715-0.766)} & \makecell{\textbf{0.818}\\[-2pt]\tiny (0.794-0.839)} & \makecell{\textbf{0.006}\\[-2pt]\tiny (0.005-0.007)} & \makecell{\textbf{0.001}\\[-2pt]\tiny (0.001-0.002)} & \makecell{0.007\\[-2pt]\tiny (0.004-0.011)} \\
\midrule
Aortic Aneurysm & 5y & Age-Sex & \makecell{0.727\\[-2pt]\tiny (0.705-0.747)} & \makecell{0.670\\[-2pt]\tiny (0.649-0.696)} & \makecell{0.710\\[-2pt]\tiny (0.688-0.729)} & \makecell{0.018\\[-2pt]\tiny (0.017-0.020)} & \makecell{0.004\\[-2pt]\tiny (0.003-0.006)} & \makecell{\textbf{0.013}\\[-2pt]\tiny (0.006-0.019)} \\
 &  & Last Progress & \makecell{0.656\\[-2pt]\tiny (0.634-0.680)} & \makecell{0.624\\[-2pt]\tiny (0.605-0.644)} & \makecell{0.654\\[-2pt]\tiny (0.633-0.677)} & \makecell{\textbf{0.017}\\[-2pt]\tiny (0.016-0.019)} & \makecell{0.004\\[-2pt]\tiny (0.003-0.005)} & \makecell{0.015\\[-2pt]\tiny (0.009-0.022)} \\
 &  & \ours & \makecell{\textbf{0.769}\\[-2pt]\tiny (0.753-0.785)} & \makecell{\textbf{0.707}\\[-2pt]\tiny (0.691-0.723)} & \makecell{\textbf{0.789}\\[-2pt]\tiny (0.771-0.803)} & \makecell{0.018\\[-2pt]\tiny (0.017-0.020)} & \makecell{\textbf{0.003}\\[-2pt]\tiny (0.002-0.004)} & \makecell{0.014\\[-2pt]\tiny (0.005-0.022)} \\
\end{longtable}
} \newpage
{
\centering
\scriptsize
\begin{longtable}{p{4cm}p{3cm}cccp{3cm}p{3cm}}
\caption{\textbf{Definition of disease progression prediction tasks.} We evaluated TTE performance on 78 disease progression tasks. $\tau$ is the threshold duration. Incidence refers to the percentage of patients in the test set who reach the endpoint by the specified $\tau$. The Snapshot Event and Endpoint Event columns specify all ICD-10 codes used to identify the disease indicated in the Task column. For some tasks, where the Endpoint Event is the administration of a medication, we provide the list of medication names used to identify the endpoint.} \label{tab:task_meta-dp} \\
\toprule
Category & Task & $\tau$ & N & Incid. & Snapshot Event & Endpoint Event \\
\midrule
\endfirsthead
\midrule
\multicolumn{7}{r}{Continued on next page} \\
\midrule
\endfoot
\bottomrule
\endlastfoot
Cardiovascular \& Circulatory & Aortic Stenosis Mortality & 3y & 7,352 & 14.87\% & I06.0, I06.2, I35.0, I35.2, Q23.0, Q25.3 & death \\
Cardiovascular \& Circulatory & Atrial Fibrillation to Ischemic Stroke & 3y & 23,886 & 2.87\% & I48 & I63 \\
Cardiovascular \& Circulatory & Chronic Ischemic Heart Disease to Acute Myocardial Infarction & 3y & 27,979 & 2.66\% & I25 & I21 \\
Cardiovascular \& Circulatory & Chronic Ischemic Heart Disease to Heart Failure & 3y & 31,812 & 7.30\% & I25 & I50 \\
Cardiovascular \& Circulatory & Heart Failure Mortality & 3y & 20,436 & 25.79\% & I50 & death \\
Cardiovascular \& Circulatory & Heart Failure to Acute Kidney Failure & 1y & 15,872 & 3.58\% & I50 & N17 \\
Cardiovascular \& Circulatory & Hypertension to Atrial Fibrillation & 5y & 139,256 & 5.72\% & I10 & I48 \\
Cardiovascular \& Circulatory & Hypertension to Chronic Kidney Disease Stage 3+ & 5y & 141,287 & 3.96\% & I10 & N18.3, N18.4, N18.5, N18.6 \\
\end{longtable}

} \newpage
{
\centering
\scriptsize
\begin{longtable}{p{4cm}cccccccc}
\caption{\textbf{Evaluation results of disease progression prediction tasks.} We report time-dependent performance metrics computed at $\tau$ on the test set for 78 disease progression tasks. $\tau$ is the threshold duration. Arrows indicate whether higher ($\uparrow$) or lower ($\downarrow$) values are better. ICI, Integrated Calibration Index; MCE, Maximum Calibration Error. $10^{-4}$ is the default value of the penalizer ($\lambda$) for Cox proportional hazards; if convergence cannot be achieved then $\lambda$ is increased to 1e-3 ($^*$) or 1e-2 ($^{**}$). Mean and 95\% CI reported over 100 bootstraps of the test set.} \label{tab:perf-dp} \\
\toprule
Task & $\tau$ & Model & AUC ($\uparrow$) & Bal. Acc. ($\uparrow$) & C-Index ($\uparrow$) & Brier Score ($\downarrow$) & ICI ($\downarrow$) & MCE ($\downarrow$) \\
\midrule
\endfirsthead
\midrule
\multicolumn{9}{r}{Continued on next page} \\
\midrule
\endfoot
\bottomrule
\endlastfoot
Aortic Stenosis Mortality & 3y & Age-Sex & \makecell{0.722\\[-2pt]\tiny (0.684-0.755)} & \makecell{0.653\\[-2pt]\tiny (0.618-0.687)} & \makecell{0.699\\[-2pt]\tiny (0.665-0.731)} & \makecell{0.099\\[-2pt]\tiny (0.091-0.108)} & \makecell{0.041\\[-2pt]\tiny (0.029-0.053)} & \makecell{\textbf{0.092}\\[-2pt]\tiny (0.061-0.135)} \\
 &  & Last Progress & \makecell{0.648\\[-2pt]\tiny (0.598-0.702)} & \makecell{0.618\\[-2pt]\tiny (0.575-0.670)} & \makecell{0.643\\[-2pt]\tiny (0.597-0.691)} & \makecell{0.135\\[-2pt]\tiny (0.119-0.147)} & \makecell{0.052\\[-2pt]\tiny (0.035-0.070)} & \makecell{0.172\\[-2pt]\tiny (0.103-0.262)} \\
 &  & \ours & \makecell{\textbf{0.825}\\[-2pt]\tiny (0.797-0.855)} & \makecell{\textbf{0.738}\\[-2pt]\tiny (0.701-0.772)} & \makecell{\textbf{0.801}\\[-2pt]\tiny (0.775-0.827)} & \makecell{\textbf{0.088}\\[-2pt]\tiny (0.078-0.096)} & \makecell{\textbf{0.038}\\[-2pt]\tiny (0.027-0.051)} & \makecell{0.103\\[-2pt]\tiny (0.058-0.169)} \\
\midrule
Atrial Fibrillation to Ischemic Stroke & 3y & Age-Sex & \makecell{0.615\\[-2pt]\tiny (0.572-0.653)} & \makecell{0.561\\[-2pt]\tiny (0.527-0.591)} & \makecell{0.594\\[-2pt]\tiny (0.552-0.632)} & \makecell{\textbf{0.022}\\[-2pt]\tiny (0.019-0.025)} & \makecell{\textbf{0.006}\\[-2pt]\tiny (0.004-0.008)} & \makecell{\textbf{0.017}\\[-2pt]\tiny (0.011-0.024)} \\
 &  & Last Progress & \makecell{0.562\\[-2pt]\tiny (0.500-0.618)} & \makecell{0.549\\[-2pt]\tiny (0.497-0.595)} & \makecell{0.561\\[-2pt]\tiny (0.499-0.616)} & \makecell{0.024\\[-2pt]\tiny (0.020-0.028)} & \makecell{0.015\\[-2pt]\tiny (0.010-0.019)} & \makecell{0.048\\[-2pt]\tiny (0.028-0.068)} \\
 &  & \ours & \makecell{\textbf{0.677}\\[-2pt]\tiny (0.644-0.712)} & \makecell{\textbf{0.630}\\[-2pt]\tiny (0.598-0.668)} & \makecell{\textbf{0.658}\\[-2pt]\tiny (0.628-0.693)} & \makecell{\textbf{0.022}\\[-2pt]\tiny (0.019-0.025)} & \makecell{0.008\\[-2pt]\tiny (0.006-0.011)} & \makecell{0.034\\[-2pt]\tiny (0.019-0.049)} \\
\end{longtable}
} \newpage
{
\centering
\scriptsize
\begin{longtable}{p{4cm}p{3.4cm}cccp{3cm}p{3cm}}
\caption{\textbf{Definition of treatment response prediction tasks.} We evaluated TTE performance on 59 treatment response tasks. $\tau$ is the threshold duration. Incidence refers to the percentage of patients in the test set who reach the endpoint by the specified $\tau$. The Snapshot Event column specifies all medications used to identify the treatment indicated in the Task column. Integers refer to ingredient RxCUIs, and the $\land$ symbol indicates combination therapies where both ingredients must be present. The Endpoint Event column specifies all ICD-10 codes used to identify the disease indicated in the Task column.} \label{tab:task_meta-te} \\
\toprule
Category & Task & $\tau$ & N & Incid. & Snapshot Event & Endpoint Event \\
\midrule
\endfirsthead
\midrule
\multicolumn{7}{r}{Continued on next page} \\
\midrule
\endfoot
\bottomrule
\endlastfoot
Cardiovascular \& Circulatory & MACE-Free Survival on Amlodipine for Hypertension & 3y & 15,192 & 26.96\% & 17767 & I21, I22, I60, I61, I62, I63, death \\
Cardiovascular \& Circulatory & MACE-Free Survival on Aspirin for Chronic Ischemic Heart Disease & 3y & 10,482 & 27.64\% & 1191 & I21, I22, I60, I61, I62, I63, death \\
Cardiovascular \& Circulatory & MACE-Free Survival on Atorvastatin for Chronic Ischemic Heart Disease & 3y & 7,681 & 31.51\% & 83367 & I21, I22, I60, I61, I62, I63, death \\
Cardiovascular \& Circulatory & MACE-Free Survival on Chlorthalidone for Hypertension & 3y & 973 & 22.45\% & 2409 & I21, I22, I60, I61, I62, I63, death \\
Cardiovascular \& Circulatory & MACE-Free Survival on Clopidogrel for Chronic Ischemic Heart Disease & 3y & 3,019 & 30.49\% & 32968 & I21, I22, I60, I61, I62, I63, death \\
Cardiovascular \& Circulatory & MACE-Free Survival on Lisinopril for Hypertension & 3y & 14,971 & 21.36\% & 29046 & I21, I22, I60, I61, I62, I63, death \\
Cardiovascular \& Circulatory & MACE-Free Survival on Losartan for Hypertension & 3y & 8,096 & 22.37\% & 52175 & I21, I22, I60, I61, I62, I63, death \\
Cardiovascular \& Circulatory & MACE-Free Survival on Ticagrelor for Chronic Ischemic Heart Disease & 3y & 2,702 & 21.71\% & 1116632 & I21, I22, I60, I61, I62, I63, death \\
\end{longtable}

} \newpage
{
\centering
\scriptsize
\begin{longtable}{p{4cm}cccccccc}
\caption{\textbf{Evaluation results of treatment response prediction tasks.} We report time-dependent performance metrics computed at $\tau$ on the test set for 59 treatment response tasks. $\tau$ is the threshold duration. Arrows indicate whether higher ($\uparrow$) or lower ($\downarrow$) values are better. ICI, Integrated Calibration Index; MCE, Maximum Calibration Error. $10^{-4}$ is the default value of the penalizer ($\lambda$) for Cox proportional hazards; if convergence cannot be achieved then $\lambda$ is increased to 1e-3 ($^*$) or 1e-2 ($^{**}$). Mean and 95\% CI reported over 100 bootstraps of the test set.} \label{tab:perf-te} \\
\toprule
Task & $\tau$ & Model & AUC ($\uparrow$) & Bal. Acc. ($\uparrow$) & C-Index ($\uparrow$) & Brier Score ($\downarrow$) & ICI ($\downarrow$) & MCE ($\downarrow$) \\
\midrule
\endfirsthead
\midrule
\multicolumn{9}{r}{Continued on next page} \\
\midrule
\endfoot
\bottomrule
\endlastfoot
MACE-Free Survival on Amlodipine for Hypertension & 3y & Age-Sex & \makecell{0.634\\[-2pt]\tiny (0.615-0.655)} & \makecell{0.585\\[-2pt]\tiny (0.568-0.605)} & \makecell{0.608\\[-2pt]\tiny (0.591-0.624)} & \makecell{0.156\\[-2pt]\tiny (0.150-0.163)} & \makecell{0.026\\[-2pt]\tiny (0.015-0.036)} & \makecell{\textbf{0.055}\\[-2pt]\tiny (0.030-0.079)} \\
 &  & Last Progress & \makecell{0.674\\[-2pt]\tiny (0.653-0.695)} & \makecell{0.624\\[-2pt]\tiny (0.603-0.641)} & \makecell{0.657\\[-2pt]\tiny (0.638-0.676)} & \makecell{0.152\\[-2pt]\tiny (0.145-0.159)} & \makecell{0.026\\[-2pt]\tiny (0.016-0.035)} & \makecell{0.065\\[-2pt]\tiny (0.038-0.100)} \\
 &  & \ours & \makecell{\textbf{0.792}\\[-2pt]\tiny (0.773-0.810)} & \makecell{\textbf{0.732}\\[-2pt]\tiny (0.713-0.749)} & \makecell{\textbf{0.765}\\[-2pt]\tiny (0.749-0.782)} & \makecell{\textbf{0.132}\\[-2pt]\tiny (0.124-0.138)} & \makecell{\textbf{0.022}\\[-2pt]\tiny (0.012-0.029)} & \makecell{0.069\\[-2pt]\tiny (0.029-0.122)} \\
\midrule
MACE-Free Survival on Aspirin for Chronic Ischemic Heart Disease & 3y & Age-Sex & \makecell{0.636\\[-2pt]\tiny (0.614-0.662)} & \makecell{0.595\\[-2pt]\tiny (0.575-0.614)} & \makecell{0.618\\[-2pt]\tiny (0.597-0.642)} & \makecell{0.166\\[-2pt]\tiny (0.156-0.173)} & \makecell{0.032\\[-2pt]\tiny (0.020-0.044)} & \makecell{0.076\\[-2pt]\tiny (0.045-0.116)} \\
 &  & Last Progress & \makecell{0.667\\[-2pt]\tiny (0.642-0.689)} & \makecell{0.626\\[-2pt]\tiny (0.604-0.645)} & \makecell{0.650\\[-2pt]\tiny (0.631-0.669)} & \makecell{0.167\\[-2pt]\tiny (0.157-0.178)} & \makecell{0.034\\[-2pt]\tiny (0.023-0.044)} & \makecell{0.084\\[-2pt]\tiny (0.051-0.128)} \\
 &  & \ours & \makecell{\textbf{0.768}\\[-2pt]\tiny (0.741-0.790)} & \makecell{\textbf{0.713}\\[-2pt]\tiny (0.691-0.734)} & \makecell{\textbf{0.747}\\[-2pt]\tiny (0.725-0.765)} & \makecell{\textbf{0.146}\\[-2pt]\tiny (0.137-0.155)} & \makecell{\textbf{0.028}\\[-2pt]\tiny (0.015-0.041)} & \makecell{\textbf{0.071}\\[-2pt]\tiny (0.037-0.125)} \\
\end{longtable}
} \newpage
{
\centering
\scriptsize
\begin{longtable}{p{4cm}p{3.4cm}cccp{3cm}p{3cm}}
\caption{\textbf{Definition of adverse event prediction tasks.} We evaluated TTE performance on 17 adverse event tasks. $\tau$ is the threshold duration. Incidence refers to the percentage of patients in the test set who reach the endpoint by the specified $\tau$. The Snapshot Event column specifies all medications used to identify the treatment indicated in the Task column. Integers refer to ingredient RxCUIs, and the $\land$ symbol indicates combination therapies where both ingredients must be present. The Endpoint Event column specifies all ICD-10 codes used to identify the disease indicated in the Task column.} \label{tab:task_meta-ae} \\
\toprule
Category & Task & $\tau$ & N & Incid. & Snapshot Event & Endpoint Event \\
\midrule
\endfirsthead
\midrule
\multicolumn{7}{r}{Continued on next page} \\
\midrule
\endfoot
\bottomrule
\endlastfoot
Cardiovascular \& Circulatory & Angioedema on ACE Inhibitors & 90d & 37,960 & 0.05\% & Lisinopril, Zestril, Prinivil, Enalapril, Vasotec, Ramipril, Altace, Benazepril, Lotensin, Captopril, Capoten, Fosinopril, Monopril, Quinapril, Accupril, Perindopril, Aceon, Trandolapril, Mavik, Moexipril, Univasc & T78.3 \\
Cardiovascular \& Circulatory & Cardiomyopathy After Anthracyclines & 5y & 6,254 & 9.28\% & doxorubicin, Adriamycin, Doxil, Caelyx, Myocet, Lipodox, Rubex, Celdoxome Pegylated Liposomal, Zolsketil Pegylated Liposomal, epirubicin, Ellence, Pharmorubicin, Farmorubicin, daunorubicin, Cerubidine, DaunoXome, Vyxeos, idarubicin, Idamycin, Zavedos, valrubicin, Valstar, aclarubicin, Aclacinon, Aclacin, pirarubicin, Pinorubin, Therarubicin, amrubicin, Calsed & I42, I50 \\
Cardiovascular \& Circulatory & MACE on Levothyroxine with History of Heart Failure & 2y & 1,805 & 36.45\% & 10582 & I21, I22, I60, I61, I62, I63, death \\
Cardiovascular \& Circulatory & Major Bleeding on Oral Anticoagulants & 1y & 18,723 & 2.72\% & Warfarin, Coumadin, Jantoven, Apixaban, Eliquis, Rivaroxaban, Xarelto, Edoxaban, Savaysa, Dabigatran etexilate, Pradaxa & D62, I60, I61, I62, I85.01, I97.4, I97.6, K25.0, K26.0, K29.01, K92.2, N02, R04.2, R04.89, R58 \\
Endocrine, Metabolic \& Reproductive & Hyperthyroidism on Amiodarone & 2y & 9,870 & 0.79\% & Amiodarone, Cordarone, Pacerone, Nexterone & E05 \\
Endocrine, Metabolic \& Reproductive & Hypoglycemia on Sulfonylureas & 1y & 4,620 & 1.04\% & Glipizide, Glucotrol, Glyburide, Diabeta, Micronase, Glibenclamide, Glimepiride, Amaryl, Chlorpropamide, Tolazamide, Tolbutamide & E16.0, E16.1, E16.2, E16.A \\
Endocrine, Metabolic \& Reproductive & Hypothyroidism on Amiodarone & 2y & 4,805 & 6.26\% & Amiodarone, Cordarone, Pacerone, Nexterone & E02, E03 \\
\end{longtable}

} \newpage
{
\centering
\scriptsize
\begin{longtable}{p{4cm}cccccccc}
\caption{\textbf{Evaluation results of adverse event prediction tasks.} We report time-dependent performance metrics computed at $\tau$ on the test set for 17 adverse event tasks. $\tau$ is the threshold duration. Arrows indicate whether higher ($\uparrow$) or lower ($\downarrow$) values are better. ICI, Integrated Calibration Index; MCE, Maximum Calibration Error. $10^{-4}$ is the default value of the penalizer ($\lambda$) for Cox proportional hazards; if convergence cannot be achieved then $\lambda$ is increased to 1e-3 ($^*$) or 1e-2 ($^{**}$). Mean and 95\% CI reported over 100 bootstraps of the test set.} \label{tab:perf-ae} \\
\toprule
Task & $\tau$ & Model & AUC ($\uparrow$) & Bal. Acc. ($\uparrow$) & C-Index ($\uparrow$) & Brier Score ($\downarrow$) & ICI ($\downarrow$) & MCE ($\downarrow$) \\
\midrule
\endfirsthead
\midrule
\multicolumn{9}{r}{Continued on next page} \\
\midrule
\endfoot
\bottomrule
\endlastfoot
Angioedema on ACE Inhibitors & 90d & Age-Sex & \makecell{0.600\\[-2pt]\tiny (0.524-0.719)} & \makecell{\textbf{0.592}\\[-2pt]\tiny (0.499-0.672)} & \makecell{0.603\\[-2pt]\tiny (0.527-0.720)} & \makecell{0.001\\[-2pt]\tiny (0.000-0.001)} & \makecell{\textbf{0.000}\\[-2pt]\tiny (0.000-0.001)} & \makecell{\textbf{0.001}\\[-2pt]\tiny (0.001-0.003)} \\
 &  & Last Progress & \makecell{0.556\\[-2pt]\tiny (0.420-0.683)} & \makecell{0.500\\[-2pt]\tiny (0.500-0.501)} & \makecell{0.560\\[-2pt]\tiny (0.426-0.686)} & \makecell{\textbf{0.000}\\[-2pt]\tiny (0.000-0.001)} & \makecell{\textbf{0.000}\\[-2pt]\tiny (0.000-0.001)} & \makecell{\textbf{0.001}\\[-2pt]\tiny (0.001-0.002)} \\
 &  & \ours & \makecell{\textbf{0.821}\\[-2pt]\tiny (0.680-0.931)} & \makecell{0.500\\[-2pt]\tiny (0.500-0.500)} & \makecell{\textbf{0.824}\\[-2pt]\tiny (0.685-0.932)} & \makecell{\textbf{0.000}\\[-2pt]\tiny (0.000-0.000)} & nan & nan \\
\midrule
Cardiomyopathy After Anthracyclines & 5y & Age-Sex & \makecell{0.714\\[-2pt]\tiny (0.662-0.771)} & \makecell{0.682\\[-2pt]\tiny (0.630-0.733)} & \makecell{0.683\\[-2pt]\tiny (0.633-0.738)} & \makecell{0.041\\[-2pt]\tiny (0.035-0.049)} & \makecell{\textbf{0.021}\\[-2pt]\tiny (0.014-0.028)} & \makecell{\textbf{0.049}\\[-2pt]\tiny (0.032-0.081)} \\
 &  & Last Progress & \makecell{0.592\\[-2pt]\tiny (0.509-0.659)} & \makecell{0.599\\[-2pt]\tiny (0.551-0.649)} & \makecell{0.566\\[-2pt]\tiny (0.483-0.636)} & \makecell{0.044\\[-2pt]\tiny (0.036-0.053)} & \makecell{0.034\\[-2pt]\tiny (0.025-0.043)} & \makecell{0.099\\[-2pt]\tiny (0.056-0.132)} \\
 &  & \ours & \makecell{\textbf{0.771}\\[-2pt]\tiny (0.716-0.817)} & \makecell{\textbf{0.723}\\[-2pt]\tiny (0.677-0.767)} & \makecell{\textbf{0.741}\\[-2pt]\tiny (0.687-0.780)} & \makecell{\textbf{0.040}\\[-2pt]\tiny (0.032-0.050)} & \makecell{\textbf{0.021}\\[-2pt]\tiny (0.013-0.032)} & \makecell{0.066\\[-2pt]\tiny (0.031-0.126)} \\
\end{longtable}
} \newpage
{
\centering
\scriptsize
\begin{longtable}{p{4.5cm}p{4cm}cp{4.9cm}cc}
\caption{\textbf{Definition of hospital operations prediction tasks.} We evaluated TTE performance on 12 hospital operations tasks. The patient history is given to the model up to the snapshot event. The model predicts whether the endpoint event (hospital admission, discharge, or disease diagnosis) will occur within the duration indicated by $\tau$.                 Incidence refers to the percentage of patients in the test set who reach the endpoint by the specified $\tau$.                  For tasks where the trigger event is a hospital admission, we set the snapshot time to 24 hours after admission to accumulate predictive features.                  For example, in the Acute Sepsis task, the model is given all data up to 24 hours after ED arrival and predicts whether the patient will be diagnosed with sepsis within the next 6 days (1 week after admission).} \label{tab:task_meta-ops} \\
\toprule
Task & Snapshot Event & $\tau$ & Endpoint Event & N & Incid. \\
\midrule
\endfirsthead
\midrule
\multicolumn{6}{r}{Continued on next page} \\
\midrule
\endfoot
\bottomrule
\endlastfoot
Acute Cardiac Arrest & 24hr after ED Admission & 6d & I46 & 140,994 & 0.01\% \\
Acute Catheter-Associated UTI & 24hr after ED Admission & 6d & T83.511 & 89,598 & 0.01\% \\
Acute Clostridioides difficile Infection & 24hr after ED Admission & 6d & A04.7 & 135,902 & 0.05\% \\
Acute Dialysis Dependence & 24hr after ED Admission & 6d & Z99.2 & 107,664 & 0.02\% \\
Acute Kidney Failure & 24hr after ED Admission & 6d & N17 & 114,095 & 0.51\% \\
Acute Liver Failure & 24hr after ED Admission & 6d & K72.00, K72.01, K72.90, K72.91 & 129,304 & 0.01\% \\
Acute Pressure Ulcer & 24hr after ED Admission & 6d & L89 & 105,170 & 0.07\% \\
Acute Respiratory Distress Syndrome & 24hr after ED Admission & 6d & J80 & 276,203 & 0.01\% \\
\end{longtable}

} \newpage
{
\centering
\scriptsize
\begin{longtable}{p{4cm}cccccccc}
\caption{\textbf{Evaluation results of hospital operations prediction tasks.} We report time-dependent performance metrics computed at $\tau$ on the test set for 12 hospital operations tasks. $\tau$ is the threshold duration. Arrows indicate whether higher ($\uparrow$) or lower ($\downarrow$) values are better. ICI, Integrated Calibration Index; MCE, Maximum Calibration Error. $10^{-4}$ is the default value of the penalizer ($\lambda$) for Cox proportional hazards; if convergence cannot be achieved then $\lambda$ is increased to 1e-3 ($^*$) or 1e-2 ($^{**}$). Mean and 95\% CI reported over 100 bootstraps of the test set.} \label{tab:perf-ops} \\
\toprule
Task & $\tau$ & Model & AUC ($\uparrow$) & Bal. Acc. ($\uparrow$) & C-Index ($\uparrow$) & Brier Score ($\downarrow$) & ICI ($\downarrow$) & MCE ($\downarrow$) \\
\midrule
\endfirsthead
\midrule
\multicolumn{9}{r}{Continued on next page} \\
\midrule
\endfoot
\bottomrule
\endlastfoot
Acute Cardiac Arrest & 6d & Age-Sex & \makecell{0.689\\[-2pt]\tiny (0.566-0.848)} & \makecell{0.577\\[-2pt]\tiny (0.404-0.779)} & \makecell{0.688\\[-2pt]\tiny (0.565-0.848)} & \makecell{\textbf{0.000}\\[-2pt]\tiny (0.000-0.000)} & \makecell{\textbf{0.000}\\[-2pt]\tiny (0.000-0.000)} & \makecell{\textbf{0.000}\\[-2pt]\tiny (0.000-0.001)} \\
 &  & Last Progress & \makecell{0.791\\[-2pt]\tiny (0.444-0.969)} & \makecell{0.692\\[-2pt]\tiny (0.499-0.775)} & \makecell{0.790\\[-2pt]\tiny (0.442-0.968)} & \makecell{\textbf{0.000}\\[-2pt]\tiny (0.000-0.000)} & \makecell{\textbf{0.000}\\[-2pt]\tiny (0.000-0.000)} & \makecell{0.001\\[-2pt]\tiny (0.000-0.001)} \\
 &  & \ours & \makecell{\textbf{0.832}\\[-2pt]\tiny (0.633-0.969)} & \makecell{\textbf{0.833}\\[-2pt]\tiny (0.671-0.922)} & \makecell{\textbf{0.831}\\[-2pt]\tiny (0.632-0.968)} & \makecell{\textbf{0.000}\\[-2pt]\tiny (0.000-0.000)} & \makecell{\textbf{0.000}\\[-2pt]\tiny (0.000-0.000)} & \makecell{0.001\\[-2pt]\tiny (0.000-0.001)} \\
\midrule
Acute Catheter-Associated UTI & 6d & Age-Sex & \makecell{0.894\\[-2pt]\tiny (0.808-0.956)} & \makecell{0.815\\[-2pt]\tiny (0.589-0.925)} & \makecell{0.892\\[-2pt]\tiny (0.806-0.955)} & \makecell{\textbf{0.000}\\[-2pt]\tiny (0.000-0.000)} & \makecell{\textbf{0.000}\\[-2pt]\tiny (0.000-0.000)} & \makecell{\textbf{0.000}\\[-2pt]\tiny (0.000-0.001)} \\
 &  & Last Progress & \makecell{\textbf{0.970}\\[-2pt]\tiny (0.945-0.989)} & \makecell{0.784\\[-2pt]\tiny (0.781-0.787)} & \makecell{\textbf{0.969}\\[-2pt]\tiny (0.943-0.989)} & \makecell{\textbf{0.000}\\[-2pt]\tiny (0.000-0.000)} & \makecell{\textbf{0.000}\\[-2pt]\tiny (0.000-0.000)} & \makecell{0.001\\[-2pt]\tiny (0.000-0.002)} \\
\end{longtable}
} \newpage
{
\centering
\scriptsize
\begin{longtable}{p{4cm}p{3.5cm}ccp{3cm}p{3cm}}
\caption{\textbf{Definition of retrieval cohorts.} We evaluated retrieval performance on 61 cohorts. Each cohort is defined by patients diagnosed with the indicated disease (ICD-10 codes) and later prescribed the indicated therapy (ingredient RxCUIs or drug names). Cohort Size indicates the number of patients in each cohort. $P$ is the cohort size divided by the total size of the evaluation set (1.4M patients).} \label{tab:task_meta-ret} \\
\toprule
Category & Task & Cohort Size & P & Disease & Therapy \\
\midrule
\endfirsthead
\midrule
\multicolumn{6}{r}{Continued on next page} \\
\midrule
\endfoot
\bottomrule
\endlastfoot
Cardiovascular \& Circulatory & AFib on Apixaban & 7,039 & 0.49\% & I48 & 1364430 \\
Cardiovascular \& Circulatory & Acute MI on Aspirin & 2,840 & 0.2\% & I21 & 1191 \\
Cardiovascular \& Circulatory & Chronic Ischemic Heart Disease on Aspirin & 11,036 & 0.77\% & I25 & 1191 \\
Cardiovascular \& Circulatory & Heart Failure on Carvedilol & 1,594 & 0.11\% & I50 & 20352 \\
Cardiovascular \& Circulatory & Hypertension on Lisinopril & 16,562 & 1.2\% & I10 & 29046 \\
Cardiovascular \& Circulatory & Peripheral Artery Disease on Cilostazol & 156 & 0.011\% & I70.2, I70.3, I70.4, I70.5, I70.6, I70.7, I73.9 & 21107 \\
Cardiovascular \& Circulatory & Stroke on Alteplase & 652 & 0.046\% & I60, I61, I62, I63 & 8410 \\
Cardiovascular \& Circulatory & VTE on Apixaban & 2,995 & 0.21\% & I26, I81, I82 & 1364430 \\
\end{longtable}

} \newpage
{
\centering
\scriptsize
\begin{longtable}{p{5cm}cccccc}
\caption{\textbf{Evaluation results of retrieval tasks.} We report retrieval accuracy (Acc.) and mean average precision (mAP) for each of the 61 cohorts defined in Table \ref{tab:task_meta-ret}. Higher ($\uparrow$) values are better. Mean $\pm$ SD reported over 5 folds.} \label{tab:perf-ret} \\
\toprule
Task & Model & Top-1 Acc. ($\uparrow$) & Top-5 Acc. ($\uparrow$) & Top-10 Acc. ($\uparrow$) & mAP@5 ($\uparrow$) & mAP@10 ($\uparrow$) \\
\midrule
\endfirsthead
\midrule
\multicolumn{7}{r}{Continued on next page} \\
\midrule
\endfoot
\bottomrule
\endlastfoot
AFib on Apixaban & Last Progress & 0.071 ± 0.001 & 0.245 ± 0.004 & 0.369 ± 0.002 & 0.038 ± 0.001 & 0.027 ± 0.001 \\
 & \ours & \textbf{0.302 ± 0.006} & \textbf{0.734 ± 0.005} & \textbf{0.870 ± 0.005} & \textbf{0.188 ± 0.002} & \textbf{0.151 ± 0.002} \\
\midrule
Acute MI on Aspirin & Last Progress & 0.035 ± 0.003 & 0.120 ± 0.004 & 0.199 ± 0.007 & 0.016 ± 0.001 & 0.011 ± 0.000 \\
 & \ours & \textbf{0.125 ± 0.006} & \textbf{0.383 ± 0.008} & \textbf{0.565 ± 0.013} & \textbf{0.070 ± 0.002} & \textbf{0.054 ± 0.002} \\
\midrule
Chronic Ischemic Heart Disease on Aspirin & Last Progress & 0.095 ± 0.001 & 0.314 ± 0.002 & 0.461 ± 0.003 & 0.051 ± 0.000 & 0.038 ± 0.000 \\
 & \ours & \textbf{0.339 ± 0.004} & \textbf{0.770 ± 0.003} & \textbf{0.889 ± 0.001} & \textbf{0.227 ± 0.003} & \textbf{0.190 ± 0.002} \\
\midrule
Heart Failure on Carvedilol & Last Progress & 0.024 ± 0.006 & 0.097 ± 0.009 & 0.162 ± 0.009 & 0.011 ± 0.001 & 0.007 ± 0.001 \\
 & \ours & \textbf{0.139 ± 0.009} & \textbf{0.467 ± 0.014} & \textbf{0.638 ± 0.018} & \textbf{0.076 ± 0.003} & \textbf{0.056 ± 0.002} \\
\end{longtable}
}

\end{nolinenumbers}

\end{document}